\documentclass[CGV,biber]{nowfnt} 

\usepackage[utf8]{inputenc}
\usepackage{gensymb}

\usepackage{comment}

\usepackage{amsmath}
\usepackage{amsfonts,amssymb}
\usepackage{longtable}
\usepackage{tabularx}
\usepackage{adjustbox}
\newcolumntype{L}[1]{>{\raggedright\let\newline\\\arraybackslash\hspace{0pt}}m{#1}}
\usepackage{fixltx2e}

\DeclareCiteCommand{\citeyear*}
    {}
    {\bibhyperref{\printdate\printfield{extrayear}}}
    {\multicitedelim}
    {}

\usepackage{array}
\usepackage{graphicx}
\newcommand*\rot{\rotatebox{90}}

\DeclareSourcemap{
\maps[datatype=bibtex]{
\map[overwrite]{
\step[fieldset=isbn, null]
\step[fieldset=issn, null]
\step[fieldsource=doi, final]
\step[fieldset=url, null]
}}}

\title{A Comprehensive Review of Modern Object Segmentation Approaches}

\maintitleauthorlist{
Yuanbo Wang \\
The Home Depot \\
mcdy143@gmail.com
\and
Unaiza Ahsan \\
The Home Depot \\
unaiza\_ahsan@homedepot.com
\and
Hanyan Li \\
Indeed, Inc. \\
f1annlee@gmail.com
\and
Matthew Hagen \\
Amazon.com, Inc. \\
mathage@amazon.com
}

\issuesetup
{%
 copyrightowner={Y. Wang \textit{et al.}},
 volume        = 13,
 issue         = 2-3,
 pubyear       = 2022,
 isbn          = 978-1-63828-070-5,
 eisbn         = 978-1-63828-071-2,
 doi           = 10.1561/0600000097,
 firstpage     = 111, 
 lastpage      = 283
 }

\usepackage{mwe}

\author[1]{Wang,Yuanbo}
\author[2]{Ahsan,Unaiza}
\author[3]{Li,Hanyan}
\author[4]{Hagen,Matthew}

\affil[1]{The Home Depot, USA; mcdy143@gmail.com}
\affil[2]{The Home Depot, USA; unaiza\_ahsan@homedepot.com}
\affil[3]{Indeed Inc., USA; f1annlee@gmail.com}
\affil[4]{Amazon.com, Inc., USA; mathage@amazon.com}

\articledatabox{\nowfntstandardcitation}

\begin{document}

\makeabstracttitle

\begin{abstract}
Image segmentation is the task of associating pixels in an image with their respective object class labels. It has a wide range of applications in many industries including healthcare, transportation, robotics, fashion, home improvement, and tourism. Many deep learning-based approaches have been developed for image-level object recognition and pixel-level scene understanding — with the latter requiring a much denser annotation of scenes with a large set of objects. Extensions of image segmentation tasks include 3D and video segmentation, where units of voxels, point clouds, and video frames are classified into different objects. We use ``Object Segmentation'' to refer to the union of these segmentation tasks. In this monograph, we investigate both traditional and modern object segmentation approaches, comparing their strengths, weaknesses, and utilities. We examine in detail the wide range of deep learning-based segmentation techniques developed in recent years, provide a review of the widely used datasets and evaluation metrics, and discuss potential future research directions.

\end{abstract}

\chapter{Introduction}
\label{c-intro} 

\section{Overview}
Automated visual recognition tasks such as image classification, image captioning, object detection and image segmentation are essential for image and video processing. Before the advent of deep neural networks, traditional techniques leveraged hand-crafted heuristics to extract visual features and manually tuned parameters to combine these features for inferences and decisions. These techniques are simple yet effective for many cases. However, they are often not generalizable and difficult to configure. For example, Canny Edge Detection, an algorithm that uses Gaussian filters and thresholding to identify edges in images have two key adjustable parameters-size of filters and edge strength thresholds that need to be tuned. If not selected carefully, these parameters can greatly impact the effectiveness of the algorithm, resulting in either missing or false positive edges.
\section{Convolutional Neural Networks}
With the advent of the deep learning era, more powerful and accurate automated visual recognition techniques were enabled by deep neural networks that can handle high-dimensional data. In particular, convolutional neural networks (CNN) \citep{lecun1999object} excel at image recognition and pattern identification due to its ability to capture space invariance of shapes in images. CNNs achieve this through two major components: the convolutional layer, pooling layer, and non-linear layers. The convolutional layers apply sliding kernels to extract features from the input image, starting from low-level features such as edges in the earlier layers, to high-level features such as eyes and nose in the later layers. The pooling layer then reduces the resolution of the output feature maps from the convolutional layers, allowing the network to achieve translational and deformation invariance. LeNet-5, developed by \citet{lecun1998gradient}, is the first convolutional neural network applied to the famous handwritten digit recognition task MNIST.\footnote{\url{http://yann.lecun.com/exdb/mnist/}} Although only consisting of a pair of consecutive convolutional and pooling layers followed by three fully connected layers, LeNet-5 is able to achieve a 98.49\% accuracy on the MNIST dataset. However, when applied to The ImageNet Large Scale Visual Recognition Challenge (ILSVRC) \citep{deng2009imagenet} task, the test accuracy drops to only 66\%. Compared with the human level accuracy of 94\%, LeNet-5 is not sufficient for more challenging visual recognition tasks.

In 2012, a breakthrough was made when Hinton's team won the ILSVRC challenge with deep learning \citep{krizhevsky2012imagenet}. The challenge included the largest image dataset of the time, totaling over one million images spanning 1,000 object categories. For the first three years since the challenge commenced, most visual recognition systems did not make any breakthrough at this image classification task. The deep learning model proposed by \citet{krizhevsky2012imagenet} was named AlexNet, which is often considered the first deep convolutional network. It is similar in structure with LeNet-5, but consists of 3 more convolutional layers and has a total of 62 million trainable variables. It significantly boosted the performance of previous state-of-art image classification techniques by achieving considerably better classification accuracy in the ILSVRC challenge than the second place, reducing the top-5 error rate by over 10\% \citep{krizhevsky2012imagenet}. It was not until then that deep learning started to take over as the go-to approach for challenging computer vision tasks.

Despite the significant improvement AlexNet made towards image classification, it uses too many parameters and is therefore intractable for large-scale training. VGGNet \citep{simonyan2014very} solves this problem by replacing the large kernel-sized convolutional layers with multiple layers containing 3 x 3 kernels. The network contains a total of 13 convolutional layers and 3 fully connected layers. The authors demonstrate that many small-sized filters achieve the same functions as fewer filters with large kernel sizes, but with much fewer parameters.

As CNNs continue to develop and grow deeper, a problem that often arises during the training stage is vanishing gradient – weights disappearing during back-propagation, which hinders the performance of models. ResNet \citep{he2016deep}, developed one year after VGGNet \citep{simonyan2014very}, leverages two types of skip connections – identity and projection, to tackle this issue. In addition, it also uses batch normalization \citep{ioffe2015batch} to stabilize and increase training speed.

Developed by Google, The Inception \citep{szegedy2015going} network is aimed to capture sparse salient features of varying sizes. It achieves this through stacking multiple sizes (5 x 5, 3 x 3, and 1 x 1) of kernels at the same layer and combining outputs to feed to the next layer. 1 x 1 convolutions are used to reduce the number of channels, thereby reducing the number of parameters of the network.

\section{Object Detection}
A precursor to the image segmentation problem, object detection not only classifies, but also localizes each object in an image. One of the earlier approaches to solving object detection is the sliding window approach \citep{bosch2007representing,chum2007exemplar,dalal2005histograms,ferrari2007groups,rowley1995human}. By taking different crops from an input image and applying CNNs to classify cropped regions (a cropped region is classified as “background” if no target object is identified), different objects are simultaneously localized and categorized from the input image. 

However, because objects appear in various sizes and in many locations in the image, it is very computationally expensive to apply CNNs to a potentially infinite set of cropped input regions. Instead, a more scalable approach called region proposals \citep{alexe2012measuring,uijlings2013selective,zitnick2014edge} was developed to find candidate regions in images that are likely to contain objects quickly based on traditional image and signal processing techniques. One example of such a technique is Selective Search, which uses sub-segmentation and greedy search to generate and recursively combine regions into a final list of 2,000 candidate regions \citep{uijlings2013selective}. With region proposals approaches, CNNs can then be applied to a much smaller set of cropped regions to produce final detection outputs, thus making this task much more computationally tractable.

The idea of combining region proposals and CNNs for object detection first materialized in \citet{girshick2014rich}. After the features are extracted from the proposed regions using CNNs, an SVM is used to 1) determine whether an object is present in the proposed region and classifies it if present, and 2) perform bounding box regression to compute offsets to the proposed region and refine its boundary. Although R-CNN is able to resolve the challenges with the sliding window approach to object detection, it is still very slow since CNNs need to be applied to 2,000 regions, and selective search could generate bad proposals since it is a fixed algorithm.

The same authors of R-CNN developed Fast R-CNN \citep{girshick2015fast} to address the slow selective search stage. Instead of applying CNNs to proposed regions, the input image is fed into the CNNs directly to generate a feature map. From the feature map, candidate regions are produced and an RoI pooling layer is then applied to reshape these regions into fixed sizes before they are fed into a fully connected region to produce a feature vector. Finally, a softmax layer is applied to classify this feature vector into objects, and a bounding box regressor is used to compute the offset values to refine boundaries. This is a much faster approach than R-CNN because the CNNs need only be run one time on the input image instead of on 2,000 regions. Although Fast-R-CNN is significantly faster than R-CNN during both training and testing, the region proposal stage remains to be the bottleneck. This is because it still uses selective search on the CNN feature map to propose candidate regions.

Faster R-CNN \citep{ren2015faster} is a similar approach to both algorithms above, but leverages a separate network on the CNN feature map to propose candidate regions instead of using selective search. As a result, Faster R-CNN beats its predecessors in inference speed and can generate detection results in real-time.

Whereas the object detection models from the R-CNN family use regions to identify objects in an image, YOLO \citep{redmon2016you} is an approach that looks at not just the regions of interest, but the entire image. The image is divided into grids. A set of base boxes centered at each grid cell is fed as input to a single CNN to predict both object class scores and bounding boxes. This approach is extremely fast as detection is reframed as a regression problem. YOLO also sees the entire image when making predictions and makes fewer mistakes compared to Fast R-CNN when identifying background patches \citep{redmon2016you}. However, it is not good at predicting small objects due to the constraints placed on the base boxes.

Similar to YOLO, SSD \citep{liu2016ssd} is a single-shot (achieves object localization and detection simultaneously) object detector with fast inference speed to enable real-time detection. It uses a modified version of MultiBox \citep{erhan2014scalable}, a bounding box regression technique, to generate candidate bounding boxes on each grid cell of the feature map. Given a set of default bounding boxes with different aspect ratios, SSD selects those with an intersection over union (IOU) score larger than 0.5 with the ground truth bounding box as candidates and performs bounding box regression to refine the boundaries. It also predicts, for each object class c and each candidate bounding box b, the probabilities of box b containing object c.

\section{Object Segmentation}
While object detection aims only to localize objects with bounding boxes and classifying localized objects, object segmentation adds the task of delineating precise object boundaries by classifying every unit in an input to an object class or the background. The general formulation for this problem can be defined as follows: given an input space $\mathcal{X}$, and a label space $\mathcal{L}$, object segmentation finds a mapping $f$: $\mathcal{X}\mapsto\mathcal{L}$. It has a wide range of applications in many industries including healthcare (e.g. cancer detection), transportation (self-driving), robotics (guidance), fashion (virtual try-ons), home improvement (decor visualizers), and tourism (destination recognition). Segmentation of an image breaks it down into simplified representations that can aid downstream interpretation tasks. Thus, segmentation differs from detection as it does not have to be tied to a specific task, but rather an intermediate step in machine perception. Nevertheless, it is more challenging and usually more time-consuming than object detection, thereby requiring more advanced techniques and more high-quality annotated training data. Object segmentation is fundamentally tied to human image perception as recognition of objects from an input image can be conceptualized into a process of dissecting regions of deep concavity into simple volumetric components such as blocks, wedges, and cones \citep{biederman1985human}.

Recognition-by-components (RBC) theory states that robust object perception is possible because detection of edge properties such as curvature, symmetry, and parallelism is usually invariant over image quality and viewing angle \citep{biederman1985human}. However, because machines process images through various means unlike the process of human interpretation, it is far from enough to only focus on these components when performing object segmentation. In this monograph, we review various techniques developed over the years for the task of object segmentation, with a focus on the deep learning-based techniques for the two most widely solved segmentation tasks: Semantic Segmentation and Instance Segmentation. We also survey methods developed for Panoptic, Video and 3D data segmentation. We categorize and compare various techniques developed for each task. The readers will find that various themes emerge from these techniques that push machines to their limits and often times deviate from human perception principles. In addition, we provide an overview of the widely used benchmark datasets for each of these techniques, along with the respective evaluation metrics to measure the models' performances. Finally, we discuss potential directions for future research in these areas.

\chapter{Traditional Methods in Image Segmentation}
Many traditional methods have been proposed for image segmentation. Otsu’s method uses the threshold between classes with the aim to minimize the intra-class variance \citep{otsu1979threshold}.  Seeded region growing, a region-based approach, uses initial points to expand or grow the regions and then makes up the segmented objects \citep{adams1994seeded,qin2010scene,ramli2020homogeneous}. The Edge detection method identifies pixels representing object edges in the images, which are then used to determine the labels of the remaining pixels \citep{rong2014improved,canny1986computational}. Level set techniques based on Hamilton-Jacobi formulations \citep{osher1988fronts} are also widely used \citep{malladi1995shape,luo2019convex,yan2020convexity,yang2017level}.   Graph-based algorithms use a data structure consisting of vertices and weighted edges, and some typical methods are Normalized cuts \citep{shi2000normalized}, Graph cuts \citep{moore2008superpixel}, and Entropy Rate Superpixel Segmentation \citep{liu2011entropy} algorithms. The mean shift method is based on density estimations in a feature space and efficiently finds peaks in a high-dimensional data distribution without explicitly computing the complete function \citep{d2002robust,szeliski2010computer}.  Mumford-Shah functional finds the edge set and a piecewise smooth approximation given the image intensity function \citep{mumford1989optimal}.

Active Contour Models (ACMs) is another traditional gradient-based method first introduced by \citet{kass1988snakes}. It can be applied to various tasks such as edge detection, motion tracking, and segmentation. It formulates segmentation as an energy minimization problem and uses partial differential equations (PDEs) – based methods for optimizing active contours. Several variations of ACMs have also been developed including active contour without edge (ACWE) \citep{chan2001active} and fast global minimization-based active contour model (FGM-aCM) \citep{chen2019learning}. Recent works have combined ACMs and deep neural networks for image segmentation.

\citet{chen2019learning} propose an active contour – based loss function that combines region and contour length information of segmentation masks for CVD diagnosis. The base architecture used for segmentation is Dense block based U-Net (Dense-Net) \citep{huang2017densely}. The proposed loss function has superior performance compared with the common Cross-entropy loss when using either U-Net \citep{ronneberger2015u} or Dense-Net \citep{huang2017densely}. 

\citet{gur2019unsupervised} develop another unsupervised loss function based on morphological ACWE \citep{chan2001active}. A novel pooling layer is developed that mimics curvature morphological operators, examples of which include dilation and erosion operations. The main network is an encoder-decoder based architecture with successive morphological pooling layers for smoothing. It achieves the state-of-the-art performance on the DeepVess \citep{haft2019deep} and VesselNN \citep{teikari2016deep} datasets .

Besides incorporating ACMs into loss functions, ACMs have also been used as a means for post-processing outputs from FCNs \citep{le2018reformulating,rupprecht2016deep}. Trainable Deep Active Contours (TDACs) \citep{hatamizadeh2020end} is one of the most recent works that fully integrates ACM and CNNs into an end-to-end trainable architecture. The backbone CNN directly predicts per-pixel 2D parameter maps to initiate and control a Eulerian energy functional that the ACM utilizes. Each term of the energy functional is automatically differentiable with a TensorFlow implementation framework, allowing for optimization through stochastic gradient descent. The Eulerian ACM energy model improves the boundaries of the object segmentation by adjusting the contours. This work achieves state-of-the-art performance on the Vaihingen buildings dataset \citep{gerke2014use} and the Bing Huts dataset \citep{marcos2018learning}. Other works integrating ACM into deep networks for image segmentation include \citet{hu2017deep}, \citeauthor{hatamizadeh2019deep} (\citeyear*{hatamizadeh2019deep}, \citeyear*{hatamizadeh2019end}), \citet{cheng2019darnet}, \citet{wang2019object}, \citet{hoogi2016adaptive}, \citet{akbari2021deep}, and \citet{peng2020deep}. 

Traditional approaches often focus on extracting the lower level features such as edges, shapes, and gradients from the scene. These methods have low demands for computational resources, but are limited in performance. They are often challenged by complex scenes or obscure objects, and sometimes require manual parameter tuning. Nevertheless, traditional methods provide the foundation for many advanced deep learning models for object segmentation, and many modern approaches have continued to leverage certain components from these methods to boost their performances.

\chapter{Deep Models for Semantic Segmentation}
Semantic segmentation is a form of dense prediction where each pixel from an image is associated with a class label. If multiple instances of the same object class exist in an image, there is no need to distinguish each individual instance with a different label. For clarity, this section only considers Semantic image segmentation, which takes as input either a colored RGB image with three channels ($height \times width \times 3$) or a grayscale image ($height \times weight \times 1$) and produces a label map ($height \times weight \times 1$), with each pixel (index) assigned an integer that represents an object (class) label corresponding to the predicted object type at the corresponding pixel location in the input image. Mathematically, semantic image segmentation can be formulated as the following: given an input image $\mathcal{P}={p\textsubscript{1},p\textsubscript{2}, ..., p\textsubscript{m}}$, and a label space $\mathcal{L}={l\textsubscript{1},l\textsubscript{2}, ..., l\textsubscript{n}}$, it assigns an element from $\mathcal{L}$ to each element of $\mathcal{P}$.

Segmentation by itself can be defined without a specific task. Prior to the deep learning era, due to the limited capacity of traditional computer vision methods, segmentation was limited to ``processing'' of images, where the output included regions and structures such as edges (lines and curves) and shapes, which can be obtained with low-level image information such as gradients and colors \citep{arbelaez2010contour}. As deep learning technologies matured, semantic segmentation has evolved from ``processing'' to ``recognizing and understanding'' the images, particularly on a dense, pixel-level. Semantic segmentation plays an important role in the medical domain, autonomous driving, and modern robot vision and understanding. This section reviews various methods that focus on solving the problem of semantic segmentation.

\label{prep}
\section{Fully Convolutional Networks}
\begin{figure}[h]
\centering
\includegraphics[width=0.9\textwidth]{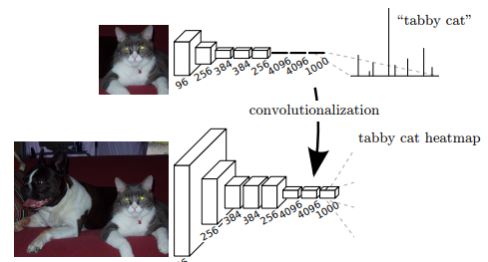}
\caption{Architecture of the Fully Convolutional Networks. From \citet{long2015fully}.}
\label{fig:fcn}
\end{figure}

One of the earliest influential methods towards semantic segmentation is the Fully Convolutional Networks (FCN) \citep{long2015fully}. By removing the last fully connected layer from traditional CNN networks, FCN generates a heatmap that preserves the spatial information from the input image. This feature map is then up-sampled using fixed interpolation techniques such as bilinear interpolation or using learned deconvolution layers. In addition, skip-connections are added between final up-sampling layers and the earlier feature maps so that visual information at different granularities can be combined to produce the final segmentation output. The entire architecture is shown in Figure \ref{fig:fcn}. FCN achieves 62.7\% mean IOU on the PASCAL VOC 2011 \citep{everingham2010pascal} dataset, approximately 10\% better than the previous state-of-art using SDS \citep{hariharan2014simultaneous}. It also reduces inference time by 286x overall. However, FCN remains to be limited by its high computational requirements due to the number of parameters and its slow real-time inference speed. More importantly, by using only convolution layers, FCN only captures local information when making pixel inferences. Additionally, it suffers from rough segmentation boundaries due to the loss of information during the up-sampling stages.

The portion of the FCN responsible for down-sampling and feature map generation is an encoder, which is usually a base classification model such as VGGNet or ResNet stripped of its fully connected layers. The deconvolution / up-sampling portion of the FCN is a decoder, which produces pixel-wise predictions based on the feature map output from the encoder. Many semantic segmentation approaches are based on this encoder-decoder structure, with variations mainly at the decoder portion of the network. 

\subsubsection{U-Net}
The U-Net \citep{ronneberger2015u} is based on FCN, resembling its encoder-decoder structure. U-Net addresses the issue of FCN’s information loss by adding shortcut connections between each pair of corresponding down-sampling and up-sampling layers so that the earlier layers that capture lower-level details of the input image can propagate such information to the up-sampling layers to help with more precise localization. U-Net was developed and widely used for biomedical imaging tasks and has been later extended into various modified architectures \citep{quan2016fusionnet, li2018h, shah2018stacked} for other tasks.

\section{Deconvolution-based Segmentation}
Rather than using direct upsampling in the decoder portion of the network to recover the original dimension of the input image for the final dense prediction map, deconvolution-based segmentation approaches use deconvolution layers to learn how to expand the output feature map from the encoder layers. This allows the model to learn non-linear upsampling with more model parameters, whereas bilinear upsampling used in previous approaches only allow linear upsampling of encoded features with fewer model parameters but a faster computation speed. Based on the FCN architecture, DeconvNet \citep{noh2015learning} is the first method that applies trainable filters in the decoder portion of the network to generate dense feature maps. 

\subsubsection{SegNet}
The SegNet \citep{badrinarayanan2017segnet} model also uses the encoder-decoder structure and is a lightweight version of DeconvNet \citep{noh2015learning}. Instead of transferring the entire feature maps from down-sampling to up-sampling layers like U-Net, it uses pooling indices computed from the max pooling layers during the down-sampling stage when performing deconvolution so that the up-sampling process is non-linear. This reduces the memory required and consequently model size. Bayesian SegNet \citep{kendall2015bayesian} is a modified version of SegNet that aims to add a probability measure to the segmentation output, and demonstrates that by modeling this uncertainty, state-of-art model architectures benefit from improved segmentation performance without additional parameters.

\subsubsection{SDN}
\begin{figure}[h]
\centering
\includegraphics[width=\textwidth]{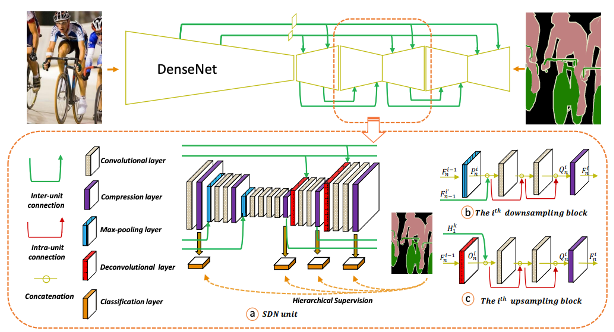}
\caption{Overall SDN architecture (upper) and SDN unit (lower). From \citet{fu2019stacked}.}
\label{fig:sdn}
\end{figure}

Stacked Deconvolutional Network (SDN) \citep{fu2019stacked} aims to improve the resolution of the segmentation results progressively through stacking of SDN units. Each SDN unit is a shallow deconvolutional network composed of two down-sampling blocks followed by two up-sampling blocks. Intra-unit connections links inputs between two consecutive convolutional layers within each SDN unit. Inter-unit connections are similar to the skip connections employed in FCN that links corresponding down-sampling and up-sampling blocks. These dense connections allow efficient reuse of multiscale information across SDN units but increase GPU memory usage. Compression units that reduce feature map channels are therefore employed to reduce the computational demands. Figure \ref{fig:sdn} shows the architecture of the SDN. SDN achieves a superior performance of 86.6\% MIoU on the PASCAL VOC 2012 \citep{everingham2010pascal} dataset, making it one of the best among the earlier semantic segmentation methods.

\subsubsection{LSD-Net}
The Locality-Sensitive Deconvolution Networks (LSD-Net) \citep{cheng2017locality} is developed to improve two aspects of the deconvolution layers: boundary refinement and RGB-D fusion. It achieves the former by incorporating visual cues from raw RGB-D data, and the latter through a gated fusion layer that learns to adjust the RGB and depth’s contributions at each pixel location. Through these enhancements, the model is able to achieve state-of-art performance on the NYU-Depth v2 \citep{silberman2012indoor} dataset.

\section{Context-based Segmentation}
\looseness=-1
Pixel classification can be a difficult process due to the confusion made when the network only relies on local information – a drawback of FCN-based architectures. These models address this issue by incorporating contextual knowledge when predicting the semantic label for a pixel. Various forms of image contexts can be encoded using different approaches, including multi-resolution feature maps, atrous (dilated) convolutions, recurrent networks, attention, and Conditional Random Fields. 
\subsection{Multi-resolution Features-based Segmentation}
By taking into account multi-resolution features, these models often achieve better performances at the expense of increased model parameters and slower computational speeds. Many approaches have successfully captured and fused multiscale features for dense pixel predictions and will be discussed in this section.
\subsubsection{FPN}
\begin{figure}[h]
\centering
\includegraphics[width=0.9\textwidth]{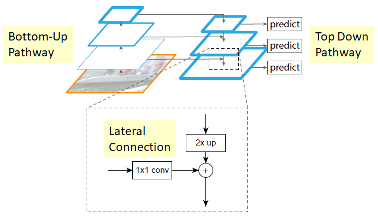}
\caption{Feature Pyramids. From \citet{lin2017feature}.}
\label{fig:fpn}
\end{figure}

Feature Pyramids (Figure \ref{fig:fpn}) is a common approach for encoding multiscale features. The Feature Pyramid Network (FPN) model \citep{lin2017feature} is originally developed for object detection task and consists of a top-down and a bottom-up pathway. The bottom-up pathway computes hierarchical feature maps in a feedforward fashion using a backbone network. The top-down pathway up-samples feature maps from higher pyramid levels using nearest neighbor, while lateral connections between the same spatial levels of the two pathways are merged with element-wise addition to produce enhanced features. After applying a 3x3 convolution to each of the final feature maps to reduce up-sampling’s aliasing effect, a prediction is generated from each level for object detection. For image-segmentation tasks, a multi-layer perceptron (MLP) is applied to each level of the feature pyramid to predict the segmentation masks and object scores \citep{lin2017feature}.

\subsubsection{PSPNet}
The Pyramid scene parsing network (PSPNet) \citep{zhao2017pyramid} features a CNN backbone, dilated convolutions, and a Pyramid Pooling Module which allows the network to take global information into account while making local predictions. The feature map output from the encoder layers are pooled at 4 different resolutions and 1x1 convolutions are applied to reduce the channel dimensions. These outputs are standardized through up-sampling and fused together, and finally merged with the original feature map. Final segmentation predictions are generated after a convolutional layer is applied to the final concatenated feature map. The Pyramid Pooling Module has laid the foundation for many later works including the Deeplab (\citeauthor{chen2017deeplab}, \citeyear*{chen2017deeplab}; \citeyear*{chen2017rethinking}) family of approaches.
\subsubsection{DMNet}
The Dynamic multiscale Network (DMNet) \citep{he2019dynamic} is another model that leverages feature maps at multiple resolutions to generate segmentation predictions. The Dynamic Convolutional Module uses context-aware filters whose kernel sizes are dynamically determined based on the input. DMNet applies multiple Dynamic Convolutional Modules (DCMs) to the feature map output from the encoder layers in parallel to generate multiscale feature maps, which are then fused with the original feature map and passed into a convolutional layer to produce final pixel-level predictions. 
\subsubsection{CCN}
\citet{ding2018context} developed a context contrasted local model that simulates how human processes contextual information when identifying an object. More specifically, it introduces a function that allows the network to generate discriminative context features for an object and is thus able to disregard more dominant but irrelevant features when classifying the target pixel. This framework generates multi-level context-aware local features and uses gated sum to aggregate features selectively before making final predictions.
\subsubsection{MSCI}
\citet{lin2018multi} introduce a novel method called Multiscale Context Intertwining (MSCI) for combining local and global contextual information to improve semantic segmentation. It allows pairs of high- and low-resolution feature maps to exchange mutual contextual information through bidirectional connections inspired by long short-term memory (LSTM) units \citep{hochreiter1997long}. Specifically, it allows context information propagation between different regions of the input image adaptively by computing super-pixels based on image structures and establishing connection pathways based on their relationships. LSTM units are trained to propagate information along these pathways to build more powerful feature maps that combine multi-feature scales.
\subsubsection{ParseNet}
ParseNet \citep{liu2015parsenet} uses a simple technique to incorporate global context into semantic label predictions. Specifically, a ParseNet module first applies global average pooling to a feature map to obtain global features. L2 normalization is then performed, and an unpooling layer simply replicates the values in the global feature vector so that the output is the same size and can be finally combined with the original feature map. The combined feature map is then used for final predictions.
\subsubsection{RefineNet}
The multi-path refinement network (RefineNet) \citep{lin2017refinenet} is designed to address the loss of image resolution during the downsampling process in the semantic segmentation networks. RefineNet employs long-range residual connections \citep{he2016deep} to enable low-level information from the downsampling process to transfer to and refine the high-level feature maps. Local short-range residual connections are incorporated in residual Convolution Units (RCUs) and residual pooling components. These residual connections allow direct back propagation of gradients so that the network can be trained from end-to-end. In addition, this paper also designs chained residual pooling to capture contextual information from a large image region. It uses an efficient structure to fuse multiscale pooled features. Specifically, multiple pooling blocks are connected in chains, allowing later blocks to reuse earlier ones’ outputs so that features from larger regions is accessible without a large pooling window. RefineNet achieved state-of-the-art performance on seven public datasets.
\subsubsection{FaPN}
In FPN \citep{lin2017feature}, feature maps from the top-down and bottom-up pathways are merged with element-wise addition. While it is effective in capturing long-range information and contextual information, its simplicity could result in the issue of feature misalignment. \citet{huang2021fapn} design a Feature-aligned Pyramid Network (FaPN) to address this issue. There are two key components of the FaPN framework: a Feature Selection Module (FSM) and a Feature Alignment Module (FAM). The FSM applies a feature importance modeling layer to the global information extracted from each input feature map with global average pooling and learns an importance vector that is used to scale the input feature maps \citep{lin2017feature}. A feature selection layer then selectively drops useless feature maps, leaving important ones behind. FSM replaces the conventional 1x1 convolution for channel reduction. The FAM uses deformable convolution \citep{dai2017deformable} to learn the transformation offset values that guide alignment of up-sampled higher-level features. When integrated in strong baselines, FaPN improved state-of-the-art performance on Cityscapes \citep{cordts2016cityscapes} and COCO-Stuff \citep{caesar2018coco} datasets.
\subsubsection{ExFuse}
ExFuse \citep{zhang2018exfuse} is a feature fusion framework aimed to improve semantic segmentation. The authors argue that simply merging low-level and high-level features is ineffective due to semantic and resolution gaps \citep{zhang2018exfuse}. Instead, they attach semantic information to low-level features through layer rearrangement, semantic supervision and semantic embedding branch, and embed spatial information into high-resolution features through explicit channel resolution embedding and densely adjacent prediction \citep{zhang2018exfuse}. Either approach improves segmentation performance significantly. Using ResNeXt \citep{xie2017aggregated} as the backbone CNN, ExFuse achieved state-of-the-art performance on the PASCAL VOC 2012 \citep{everingham2010pascal} dataset.
\subsubsection{APCNet}
The Adaptive Pyramid Context Network (APCNet) \citep{he2019adaptive} replaces the DCMs in the DMNet \citep{he2019dynamic} with Adaptive Context Modules (ACMs). Similar to DCM, each ACM also has two separate branches. One branch is responsible for estimating Global-guided Local Affinity (GLA) coefficients which assigns weights to different areas based on their contributions to determine the semantic label of local pixels. The weight assignment takes into account both local and global representations. The other branch is used to obtain subregion representations. Outputs from the two branches are multiplied to produce the adaptive context vectors. multiscale context vectors are combined with the feature map output from the encoder before being passed into the final convolutional layer to obtain semantic labels. APCNet achieves the integration of both multiscale and global contextual information.

\subsection{Atrous-convolution-based Segmentation}
Another line of approach for incorporating multiscale contextual information to feature maps is through dilated convolutions (a.k.a. atrous convolutions) \citep{chen2017deeplab}. This type of convolution inserts gaps into kernels, expanding its field-of-view to incorporate context from a larger area without needing additional parameters. By using atrous convolution kernels of varying gap sizes, models can effectively learn the trade-offs between assimilating more contexts with larger field-of-views and localizing details with smaller field-of-views.

Atrous convolution allows us to enlarge the field of view of filters to incorporate larger context. It thus offers an efficient mechanism to control the field-of-view and finds the best trade-off between accurate localization (small field-of-view) and context assimilation (large field-of-view).

\subsubsection{DilatedNet}
\citet{yu2015multi} is one of the first works that proposes a semantic segmentation approach based on aggregation of contextual information through atrous convolutions with multiple sampling rates. Specifically, DilatedNet uses five different rates of dilation, which allows the extraction of multiscale features without losing feature resolution.

\subsubsection{DeepLab}
\begin{figure}[h]
\centering
\includegraphics[width=0.9\textwidth]{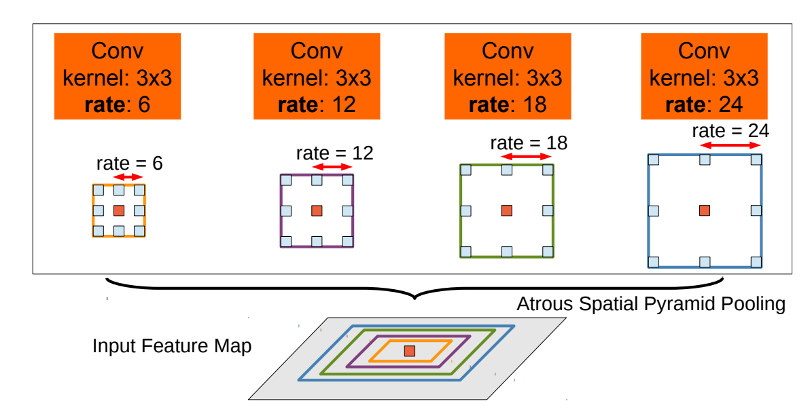}
\caption{Atrous Spatial Pyramid Pooling. From \citet{chen2017deeplab}.}
\label{fig:aspp}
\end{figure}

DeepLab \citep{chen2017deeplab} is a family of models developed at Google that proposed several key techniques to improve quality of segmentation output – atrous convolution, atrous spatial pyramid pooling (ASPP), and the incorporation of conditional random fields (CRFs). By varying the gap size, or sampling rate, atrous convolution kernels extract varying scales of information from the input feature layer. ASPP uses multiple atrous convolution kernels with different sampling rates to capture the object along with multiple scales of contextual information (Figure \ref{fig:aspp}). 

DeepLab also introduces a Graphical CRFs to address the coarse boundaries of segmentation output resulting from the usage of pooling layers. When classifying a pixel, the fully connected CRF considers other pixel labels. This refines and sharpens boundaries of segmentation results.

DeepLabV3+ \citep{chen2018encoder} adds a decoder module to further refine segmentation output, particularly on boundary refinement. Instead of using a naïve decoder such as the one used in \citet{chen2017rethinking} to simply up-sample features using bilinear interpolation, it first up-samples encoder features by a factor of 4, and then concatenates the partially up-sampled features with corresponding features generated from the backbone CNN with the same spatial resolution. These features are refined through a few 3x3 convolutions before being up-sampled by another factor of 4 to recover the original resolution and generate the final segmentation output.

\subsubsection{ResNet-DUC-HDC}
\citet{wang2018understanding} identify the “gridding” issue from dilated convolutions and propose Hybrid Dilated Convolutions (HDC) as the solution. In a sequence of dilated convolutions, the sampling rate is gradually increased to improve coverage of the receptive field’s areas, resulting in more accurate pixel-label predictions. In contrast to DeepLab’s ASPP, HDCs are applied sequentially as opposed to in parallel. In addition, the authors also design Dense Upsampling Convolution (DUC) to achieve improved pixel-level decoding. Specifically, instead of using bilinear up-sampling or deconvolution to transform the output feature map from the ResNet encoder before making final predictions, DUC uses a simple yet more effective approach. First output feature map’s dimension of $h \times w \times c$ is used to obtain $h\times w\times ({d}^2\times L)$, where $L$ is the number of categories in the segmentation task. Then, the feature map is transformed into $h\times w\times L$ using softmax before element-wise argmax is used to obtain the prediction labels. The key idea is DUC is the division of the whole label map into $d^2$ parts with the same height and width as the input feature map. DUC is also by learnable by nature and can be trained to capture fine-grained details for semantic label predictions. 

\subsubsection{DenseASPP}
Densely connected Atrous Spatial Pyramid Pooling (DenseASPP) \citep{yang2018denseaspp} was developed to address the issue of sparsity in ASPP, particularly when dealing with autonomous driving scenario, where objects exhibit large variations of scales. DenseASPP organizes atrous convolutions in a densely connected, cascade fashion, so that sampling rate continues to increase through each layer and earlier layers’ outputs are concatenated with each later layers’ outputs before the next layer of atrous convolution is applied. Without increasing the number of atrous layers, this structure can produce a much denser multiscale feature map.

\subsubsection{GALD}
The use of global aggregation techniques such as average pooling and spatial-wise feature propagation can help FCNs model long-range dependencies \citep{li2019global}. However, large patterns often dominate Global Aggregation (GA), which leads to oversmoothed smaller patterns such as boundaries and small objects. Global Aggregation and then Local Distribution (GALD) is a framework for solving this issue. Specifically, from the output feature map of the backbone FCN, the GA module first exploits the long-range contextual information by calculating global statistics for each group of spatial and channel positions and multiplying back to features in the same group \citep{li2019global}. The Local Distribution (LD) module then adaptively distributes GA features to each spatial and channel position of the output feature \citep{li2019global}. When used in semantic segmentation, GALD achieves competitive performance on Cityscapes dataset \citep{cordts2016cityscapes}.

\subsubsection{MRFM}
One of the main challenges associated with semantic segmentation is the diversity of the size of the objects in an image. The Multi-Receptive Field Module (MRFM) \citep{yuan2020multi} is designed to explicitly incorporate multiscale features to address this challenge. Unlike previous works that integrate multiscale information by modifying network structures attached to the backbone CNNs, MRFM is a redesign of the backbone network itself. Any basic module (e.g. a bottleneck in ResNet) can be replaced with MRFM, which has two separate paths – one is exactly the same as basic module, and the other follows the same structure as the basic module, but replaces basic convolution with atrous convolution. The output of each path is weighted to adaptively select the more suitable receptive field before fusion. The authors of MRFM also develops an edge-aware loss function \citep{yuan2020multi} to tackle boundary pixel classification. This loss function penalizes misclassified pixels near or on the edge more than other pixels. The Multi Receptive Field Network that combines these two components achieved state-of-the-art results on Cityscapes \citep{cordts2016cityscapes} and Pascal VOC2012 datasets \citep{everingham2010pascal}.

\subsection{Attention-based Models}
The self-attention mechanism computes representations of current input location by taking into account its relationship with all other positions of the input. In the case of semantic segmentation, self-attention effectively captures contextual information when trying to predict a target object. Attention blocks are often applied in feature concatenation, allowing the network to incorporate features that are more relevant for the prediction task by assigning more weights to them. By nature, attention mechanisms are designed for sequence-to-sequence problems, which requires reformulating the segmentation task, as was done in SEgmentation TRansformer (\textbf{SETR}) \citep{Zheng_2021_CVPR}. In addition, the versatility of the attention mechanism is shown by the models that leverage it to achieve a diverse range of sub-tasks such as capturing intra- and inter-class similarity and improving feature representation for more precise predictions. 
\subsubsection{DANet}
\begin{figure}[h]
\centering
\includegraphics[width=\textwidth]{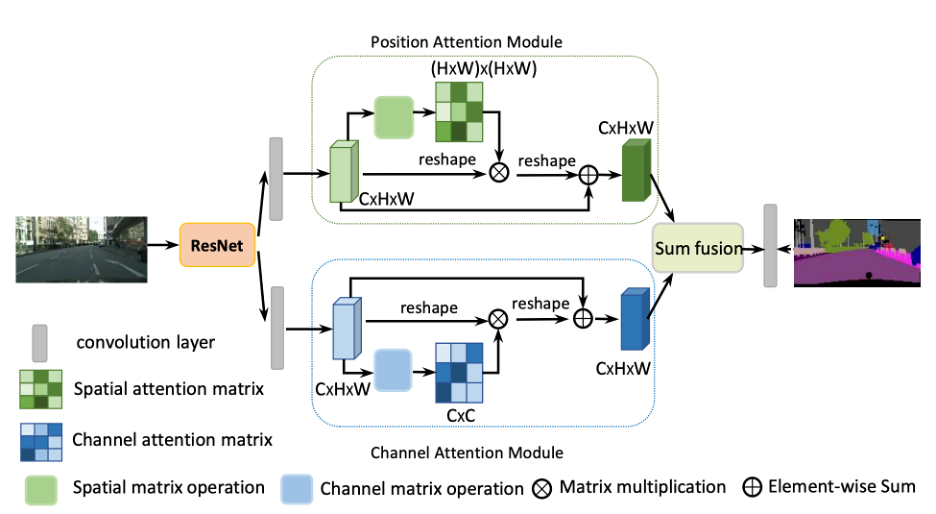}
\caption{Architecture of DANet. From \citet{fu2019dual}.}
\label{fig:danet}
\end{figure}

The Dual Attention Network (DANet) \citep{fu2019dual} utilizes attention modules to adaptively incorporate contextual dependencies into local features to improve pixel-level predictions. It applies two parallel attention modules to the feature map output from a dilated residual network backbone. The position attention module captures semantic similarities between features at all locations using self-attention mechanism. Specifically, similar features at different positions contribute to mutual improvement at each position regardless of their spatial distances. The channel attention module captures interdependent channel maps in a similar fashion. The outputs from the two parallel attention modules are summed to create improved feature representations that lead to more precise segmentation predictions \citep{fu2019dual}. The architecture of DANet is shown in Figure \ref{fig:danet}.

\subsubsection{PSANet}
The Point-wise Spatial Attention Network (PSANet) \citep{zhao2018psanet} is another model that utilizes attention module to involve contextual information to enhance feature maps. The PSA module consists of two parallel branches that generates point-wise spatial attention maps \citep{zhao2018psanet}. One branch generates distribute attention (each position distributes information to others), whereas the other branch generates collect attention (each position collects information from others). The final feature representation is composed of this bidirectional information combined with local features and is used for final segmentation predictions.

\subsubsection{GT-OCR}
The Object-Contextual Representation (OCR) \citep{yuan2020object} is an attention-like model that calculates a set of weights for aggregating object region representations based on the relationships between object regions and contextual pixels. Ground-truth OCR (GT-OCR) \citep{yuan2021segmentation} exploits ground-truth labels to improve the estimates of OCR. It is based on the motivation that “the class label assigned to one pixel is the category of the object that the pixel belongs to” \citep{yuan2021segmentation}. First, supervised by ground-truth segmentation, a backbone CNN computes a coarse soft segmentation that divides the contextual pixels into object regions. Then, each object region’s representation is estimated based on an aggregation of the representations of corresponding pixels. Finally, OCR is used to augment each pixel’s representation. GT-OCR differs from ASPP \citep{chen2017deeplab} in that it differentiates between same-object-class and different-object-class contextual pixels. GT-OCR achieves competitive performance on multiple benchmark datasets including Cityscapes \citep{cordts2016cityscapes}, ADE20K \citep{zhou2017scene}, LIP \citep{gong2017look}, PASCAL-Context \citep{mottaghi2014role}, and COCO-Stuff \citep{caesar2018coco}.

\subsubsection{SegFormer}
SegFormer \citep{xie2021segformer} is a simple and lightweight transformer-based network for semantic segmentation. It has two main components: a hierarchical Transformer encoder and an all-MLP decoder. The encoder module is a series of transformers based on ViT \citep{dosovitskiy2020image} with the same architecture but different sizes \citep{xie2021segformer}. These transformers produce multiscale features including high-resolution coarse features and low-resolution fine features \citep{xie2021segformer}, and patch merging is used to generate the final hierarchical feature map. The large effective receptive field (ERF) realized through this hierarchical transformer encoder allows a lightweight MLP to serve as the decoder, which fuses the multiscale features and predicts a final segmentation mask.

\subsubsection{EMA-Net}
Resources required to compute self-attention can be demanding. The authors of EMA-Net \citep{li2019expectation} addresses this issue by designing the Expectation-Maximization Attention (EMA) module that estimates a smaller set of bases over image pixels using the EM algorithm. The expectation (AE) step computes the attention map’s expectation, and the maximization step (AM) subsequently updates the parameters (bases).

\subsubsection{DFN}
Authors of the Discriminative Feature Network \citep{yu2018learning} attempt to improve intra-class consistency and inter-class distinction – two important challenges in semantic segmentation. They design a Channel Attention Block (CAB) based on the attention mechanism to enhance consistency. It introduces a set of weight parameters that is applied to different channels. This effectively selects discriminative features from each stage of the network to improve the consistency of intra-class prediction.

\subsubsection{SANet}
The squeeze-and-attention network (SANet) \citep{zhong2020squeeze} incorporates a novel squeeze-and-attention (SA) module to address the implicit task of pixel grouping in semantic segmentation. The authors argue that previous multiscale feature aggregation strategies employed in semantic segmentation models under-exploit the global perspective due to the spatial restrictions enforced by the grid structures of the kernels. The SA module is designed to bypass these restrictions by including an additional path similar to the squeeze-and-excitation (SE) module \citep{hu2018squeeze} for re-calibrating output feature maps’ channels. Different from the SE module, SA module does not fully squeeze the feature maps, but instead uses average pooling to downsample them. This mechanism allows attention of pixel groups belonging to the same class at different spatial scales to be emphasized, thus alleviating the local constraints of convolution kernels \citep{zhong2020squeeze}. SANet integrates outputs from multiple SA heads at different resolutions to generate class-wise masks that consider pixel grouping. It achieves state-of-the-art performance on the PASCAL Context dataset \citep{mottaghi2014role}.

\subsubsection{Swin Transformer}
To address the differences when adapting Transformer from language to vision, \citet{liu2021swin} develop a general-purpose Transformer backbone that computes hierarchical feature map representations with Shifted windows (Swin). Specifically, starting from small-sized patches, Swin Transformer gradually merges neighboring patches going into deep layers. These hierarchical feature maps allow the Swin Transformer to conveniently incorporate multiscale architectures such as FPN \citep{lin2017feature} or U-Net \citep{ronneberger2015u}. It achieves linear computational complexity through computing self-attention within local non-overlapping windows that contains fixed number of patches and evenly partition an image. The Swin approach alternates partitioning configurations in consecutive layers, establishing cross-window connections that enhance modeling power. The Swin Transformer can be used as backbones for many vision tasks including image classification and object detection. It also achieves state-of-the-art performance on COCO \citep{lin2014microsoft} and ADE20K \citep{zhou2017scene} datasets in the semantic segmentation task.

\enlargethispage{\baselineskip}
\subsubsection{CCNet}
\citet{huang2019ccnet} propose a different way to reduce the computational complexity of using attention mechanism. Instead of creating dense connections to capture pixel-wise contextual information, they elect to connect each position in the feature map to only the other positions that are in the same row and column, or the “criss-cross” path. They argue that this approach resembles using multiple sparsely-connected graphs to represent a common single densely-connected graph \citep{huang2019ccnet}. This strategy effectively reduces both space and time computational complexities from $O({N}^2)$ to $O(\sqrt{N})$. This model is able to achieve superior performance on various public semantic segmentation datasets than DeepLab \citep{yu2018learning, chen2018encoder} models and similar attention-based models including PSANet \citep{zhao2018psanet} and DFN \citep{yu2018learning}.

\subsubsection{Hierarchical Multiscale Attention for Semantic Segmentation}
\citet{chen2016attention} develop an attention mechanism that utilizes multiscale feature maps as input to train attention masks simultaneously at each scale. In contrast, \citet{tao2020hierarchical} develops a hierarchical attention-based approach that learns relative attention mask between adjacent scales, which allows for faster training with pairs of images. Using the HRNet-OCR \citep{yuan2020object} backbone network, it also achieves state-of-the-art performance on Cityscapes \citep{cordts2016cityscapes} dataset.

\enlargethispage{\baselineskip}
\subsubsection{MaskFormer}
\citet{cheng2021per} propose the MaskFormer model based on the observation that semantic- and instance-level segmentation can be solved in a unified manner with mask classification using the same model, loss, and training procedure. MaskFormer performs simple mask classification with three modules. The pixel-level module generates per-pixel embeddings using the feature map output from a backbone CNN. The transformer module computes pairs of class predictions and mask embedding vectors, each corresponding to a semantic class. Finally, the segmentation module computes the dot product between the mask embedding vectors and the per-pixel embeddings followed by sigmoid activation to produce binary mask predictions. These binary mask predictions are combined through matrix multiplication to produce final semantic segmentation labels. MaskFormer also naturally solves panoptic segmentation and achieves superior performance for both tasks on ADE20K \citep{zhou2017scene} and COCO \citep{lin2014microsoft} benchmark datasets.

\subsubsection{Mask2Former}
Mask2Former \citep{cheng2022masked} builds on MaskFormer and makes three improvements. First, standard cross-attention in the Transformer decoder is replaced with masked attention which limits the attention to localized features centered around predicted segments instead of the full feature map \citep{cheng2022masked}. This improves both the efficiency and the performance of the model. Second, high-resolution features are utilized efficiently via a feature pyramid that feeds one scale of the multi-scale feature to one Transformer decoder layer at a time \citep{cheng2022masked}. Finally, reduced computations are achieved by calculating mask loss on K randomly sampled points instead of the entire mask \citep{cheng2022masked}.

\subsubsection{Additional Works}
Attention-based models have become increasingly popular due to their ability to capture contextual information in a more efficient manner. Other attention-based semantic segmentation models include \citet{strudel2021segmenter, fu2020scene, yin2020disentangled, li2018pyramid}, and \cite{yuan2018ocnet}.

\subsection{Segmentation with Graphical Models}
CRF can be used to refine coarse feature map output from FCN, as demonstrated in the DeepLab \citep{chen2017deeplab} model. It incorporates CRFs as a post-processing module that significantly boosts its prediction accuracy. However, this mechanism has three major drawbacks: 1) CRF uses some hard-coded parameters that can be difficult to adapt to different test sets; 2) these parameters cannot be trained jointly with FCN; and 3) Incorporation of CRFs significantly increases the computational complexity of the models during inference time. A few models discussed in this section resolve some of these issues through integrating CRFs into the CNN layers to achieve end-to-end training. This improves model efficiency as well as achieving better segmentation performance.
\subsubsection{CRF-RNN}
\begin{figure}[h]
\centering
\includegraphics[width=\textwidth]{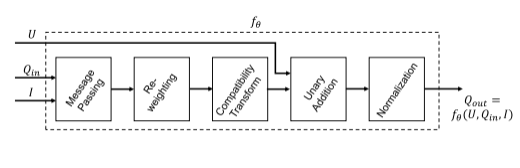}
\caption{A mean-field iteration as a CNN. From \citet{zheng2015conditional}.}
\label{fig:mfcnn}
\end{figure}

\citet{zheng2015conditional} propose a method that formulates CRF as Recurrent Neural Network (RNN), which is integrated with FCN so that the network can be trained from end-to-end. Specifically, the energy function described in \citet{krahenbuhl2011efficient} is used to optimize the CRF for semantic label prediction. This energy function consists of Unary energy – which is obtained from a CNN and predicts pixel labels independent of assignment consistency and smoothness, and Pairwise energy – which uses a smoothing term to improve label consistency in similar pixels. The mean-field algorithm is introduced in \citet{kendall2015bayesian} to approximate the CRF distribution. \citet{zheng2015conditional} first formulates a single iteration of the mean-field algorithm as a stack of CNNs (Figure \ref{fig:mfcnn}), and then further formulates repeated iterations as an RNN whose parameters can be learned through back-propagation through-time \citep{mozer1989focused}. This approach achieved superior performance on the Pascal VOC 2012 \citep{everingham2010pascal} challenge compared to state-of-art methods.

\subsubsection{Piecewise}
Unlike Deeplab \citep{chen2017deeplab} or CRF-RNN \citep{zheng2015conditional}, \citet{lin2016efficient} use CRFs to learn ‘patch-patch’ context in image regions. Instead of training CNNs and CRFs jointly, they develop a more efficient piecewise training approach. A node is created in the CRF for each rectangular region in the feature map, and pairwise connections are established between nodes within a spatial range box \citep{lin2016efficient}. Different types of range box are created to represent different spatial relations including “surrounding” and “above/below”, and a pairwise potential function is used to model each type of relation. Pairwise potential functions are defined based on the output of Pairwise-Net \citep{lin2016efficient}, which consists of two fully connected layers that are applied to the edge features that are concatenated feature vectors of two connected nodes. Unary-Net \citep{lin2016efficient}, which also consists of two fully connected layers, is applied to output from the CNN feature maps to produce the Unary potential functions. Piecewise training of CRFs is done with CNN potentials to allow efficient and parallel training.

\subsubsection{DPN}
The Deep Parsing Network (DPN) \citep{liu2015semantic} is developed to incorporate context information into CRF for semantic segmentation. DPN is based on an extension of the CNN architecture and models unary terms and various types of pairwise terms by approximating the mean field algorithm. DPN achieves high computational efficiency, and its operations can be parallelized. It also achieves competitive performance on the PASCAL VOC 2012 \citep{everingham2010pascal} dataset.

\subsubsection{GCRF}
The Gaussian Mean Field (GMF) network \citep{vemulapalli2016gaussian} is another approach that incorporates a CRF model and the mean field inference to improve semantic segmentation. This work is different from CRF-RNN \citep{zheng2015conditional} in two different ways: first, it uses simpler Gaussian CRFs with reachable optimality, instead of discrete CRFs that are more complex; second, a fixed number of steps of the mean field inference are unrolled as a deep network instead of formulated as RNNs. The proposed GMF network is combined with CNNs to form what the authors term the Gaussian CRF Network (GCRF) \citep{vemulapalli2016gaussian} that is trainable from end-to-end.

\subsubsection{Additional Works}
In addition to the work discussed in this section, models from the DeepLab \citep{chen2017deeplab} family also use CRF for improving prediction results. Other works include \citet{bell2015material} and \citet{fu2016retinal}, where a fully connected CRF is used to predict material segmentation in the wild and retina vessel segmentation, respectively.

\subsection{Models Incorporating Other Forms of Contextual Knowledge}
\subsubsection{Domain Transform}

\begin{figure}[h]
\centering
\includegraphics[width=\textwidth]{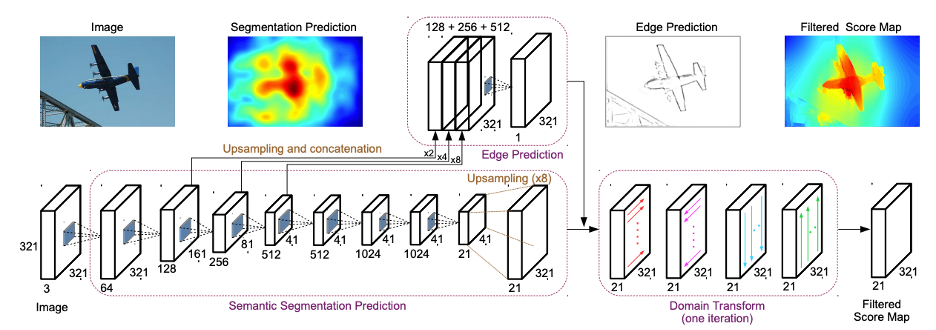}
\caption{Architecture of the proposed Domain Transform approach. From \citet{chen2016semantic}.}
\label{fig:domain}
\end{figure}

Despite the enhanced performance CRFs bring to semantic segmentation tasks, dense CRF inference can be demanding on computational power. \citet{chen2016semantic} propose an approach to replace dense CRFs with a domain transform (DT) edge-preserving filter based on a reference edge map that is directly produced from the encoder CNN that also outputs coarse semantic segmentation scores (Figure \ref{fig:domain}). Specifically, DeepLab \citep{chen2017deeplab} is used for semantic segmentation prediction, and EdgeNet \citep{chen2016semantic} is used for edge prediction. DT runs 1-D filters across both rows and columns of the coarse segmentation scores and the reference edge map to calculate the final segmentation scores. This architecture improves boundary pixel predictions while significantly reducing inference time when compared to CRF-based methods \citep{chen2017deeplab,bell2015material,fu2016retinal,vemulapalli2016gaussian,liu2015semantic}.

\subsubsection{GSCNN}
Gated Shape CNN (GSCNN) \citep{takikawa2019gated} is a two-stream CNN architecture developed to explicitly incorporate shape information from the image to improve boundary detection for semantic segmentation. A new type of gating mechanism is designed to connect the intermediate layers of the classical and the shape stream. Specifically, higher-level information from the classical stream is passed into the shape stream to denoise activations in early stages. This helps the shape stream to only focus on relevant information. Outputs from the two streams are fused with an ASPP module in multiscale fashion. A dual task regularizer is formulated to capture semantic boundary loss which penalize mismatched boundary pixels. GSCNN outperformed state-of-art approaches such as DeepLab-v3+ \citep{chen2018encoder} on the Cityscapes \citep{cordts2016cityscapes} benchmark dataset.

\subsubsection{EncNet}
Context Encoding Network (EncNet) \citep{zhang2018context} is a framework for semantic segmentation that captures global scene context through a novel Context Encoding Module and Semantic Encoding Loss (SE-loss). The architecture of EncNet is as follows: first, a pretrained ResNet backbone extracts the feature maps from the input image; the Encoding Layer, which is part of the Context Encoding Module, then generates encoded semantics capturing feature statistics and predicts scaling factors for selectively emphasizing class-dependent feature maps; the other part of the Context Encoding Module predicts each object category’s presence in the scene based on the encoded semantics and employs the SE-loss that equally considers small and big objects to regularize the training; finally, the representation of the Context Encoding Module is incorporated into the last convolutional layer before EncNet produces the per-pixel predictions. This framework only incurs marginal extra computational costs and achieved new state-of-the-art performance on the PASCAL-Context \citep{mottaghi2014role} dataset.

\subsubsection{Other Models}
Contextual Knowledge is a powerful addition to vanilla FCN features in semantic segmentation task, regardless of the form of the knowledge, as seen in the approaches discussed in this section. Other models that leverage contextual information include ACNet \citep{hu2019acnet}, E-Net \citep{paszke2016enet}, \citet{raj2015multi}, \citet{roy2016multi}, \citet{bian2016multiscale}, and \citet{pinheiro2014recurrent}.

\section{RNN-based Segmentation}
In addition to the previously discussed CRF-RNN \citep{zheng2015conditional} model, RNNs have been utilized in semantic segmentation tasks to capture the global contextual information through long-range dependencies. Objects in images usually exhibit certain relationships even when they are far away from each other. The use of RNNs addresses the limitation of CNN-based features where only local contexts are incorporated with smaller kernels. Features produced through RNN layers are usually combined with CNN-based features before being passed into the decoder for dense pixel predictions.

\subsubsection{ReSeg}
\citet{visin2016reseg} design an architecture called ReSeg based on an ReNet \citep{visin2015recurrent}, a previous model for image classification. ReNet is composed of RNN layers that capture contextual information through both horizontal and vertical sweeps of the image and hidden layer outputs. ReNet layers are stacked on the encoder layer outputs from the pretrained CNNs in ReSeg, resulting in an improved feature map that is fed into an upsampling layer before reaching final semantic predictions.

\subsubsection{DAG-RNNs}
\begin{figure}[h]
\centering
\includegraphics[width=0.7\textwidth]{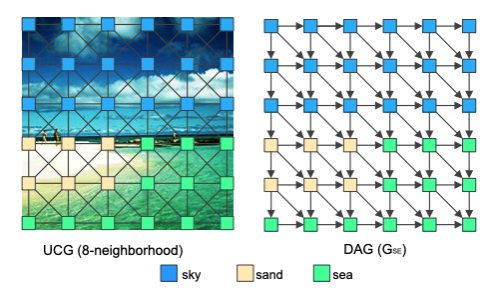}
\caption{An 8-neighborhood UCG and one of its induced DAG in the southeastern (SE) direction. From \citet{shuai2016dag}.}
\label{fig:ucgdag}
\end{figure}

In a separate work, \citet{shuai2016dag} incorporate directed acyclic graph RNNs (DAG-RNNs) to leverage long-range contextual information in dense pixel predictions. They argue that chain-structured RNNs do not capture dependencies among image units, and that undirected cyclic graphs (UCGs) can be used to better capture interactions among image units \citep{shuai2016dag}. Furthermore, they approximate UCG’s topology with multiple directed acyclic graphs (DAGs) since RNNs cannot be applied directly to UCG-structured images. Figure \ref{fig:ucgdag} shows an example of such approximation. DAG-RNNs are integrated after the encoder CNN layers to model dependencies among the elements in the output feature map. The refined feature map is then passed into the deconvolution layers and up-sampled to dense pixel predictions.

\subsubsection{LSTM-based Models}
Apart from traditional RNNs, Long Short Term Memory (LSTM) networks have also been applied to semantic segmentation tasks. In \citet{byeon2015scene}, the authors incorporate \textbf{2D LSTM} networks to extract in a single process both global and local dependencies from images for scene labeling. The input image is first divided into a nxn grid. The authors argue that window-based input units maintain global coherence of the image and local correlation of the pixels while reducing computational load. Four LSTM memory blocks which function as a 2D LSTM layer are applied to each window. The windows are connected to each surrounding directions and propagate contexts. A feedforward layer then sums up outputs from each LSTM block and applies a Hyperbolic tangent (tanh) activation. Outputs from the final LSTM blocks are summed before fed into the softmax layer to obtain probabilistic label output for each input window. This approach achieves competitive performance on Stanford Background \citep{gould2009decomposing} and SIFT Flow \citep{liu2011nonparametric} datasets with much computational complexity. However, it suffers from mistakes of mislabeling well-segmented regions due to some properties of the LSTM networks \citep{byeon2015scene}.

\citet{li2016lstm} integrate a novel LSTM structure named \textbf{LSTM-CF} into a CNN to capture contextual information from photometric and depth data. This allows end-to-end training of the network for semantic RGB-D label predictions. It also incorporates 2D contexts with both vertical and horizontal information. 

\begin{figure}[h]
\centering
\includegraphics[width=\textwidth]{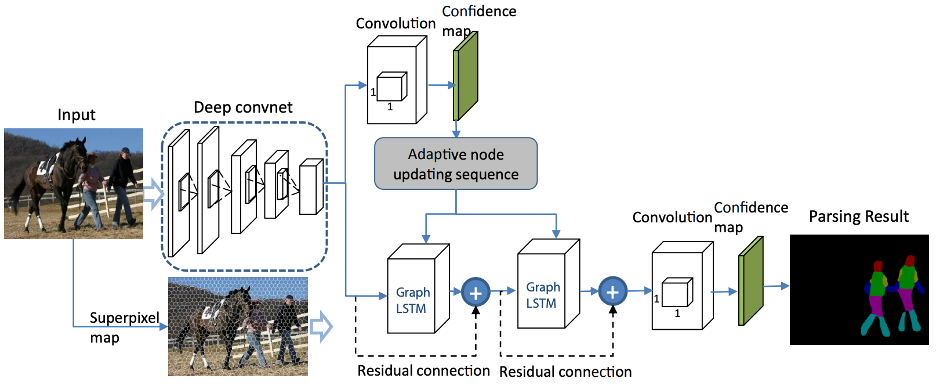}
\caption{The Graph LSTM structure. From \citet{liang2016semantic}.}
\label{fig:graphlstm}
\end{figure}

In \citet{liang2016semantic}, \textbf{Graph LSTM} Network is used to improve the modeling of the visual patterns such as object boundaries and appearance similarities by more effectively propagating information among pixels based on graph-structured representations. An undirected graph is adaptively constructed based on superpixels obtained using SLIC \citep{achanta2010slic}. Edges naturally represent spatial relations of the superpixels \citep{liang2016semantic}, and Confidence-Driven Search (CDS) is used to update the states of all nodes \citep{liang2016semantic}. Additionally, each graph LSTM unit’s forget gate learns to incorporate local contextual interactions among neighboring nodes. This Graph LSTM structure effectively exploits global context to boost local predictions, achieving superior performance on semantic object parsing challenge datasets including PASCAL-Person-Part dataset \citep{chen2014detect}, Horse-Cow parsing dataset \citep{wang2015semantic}, ATR dataset \citep{liang2015deep} and Fashionista dataset \citep{yamaguchi2012parsing}. Figure \ref{fig:graphlstm} shows the architecture of this Graph LSTM Network.

In \citet{hu2016segmentation}, an approach combining LSTMs and CNNs is used to address the novel task of segmenting images based on linguistic descriptions.  The LSTM encodes the natural language expression into a vector representation, and a FCN extracts a spatial feature map from the image \citep{hu2016segmentation}. These two outputs are passed into a fully convolutional classification network and an upsampling network to produce a final pixel-wise segmentation map. These networks are trained from end-to-end to jointly process image and text information.

\section{GAN-based Segmentation}
Generative Adversarial Networks (GANs) are often used to distinguish ground truth and synthetic images, and are widely used in image generation tasks. Recently, GANs have also shown promising results when used for semantic segmentation. They have been incorporated into segmentation models to serve two main functions: improving prediction quality with the discriminator and producing additional data with the generator in a semi-supervised training framework. 

\begin{figure}[h]
\centering
\includegraphics[width=\textwidth]{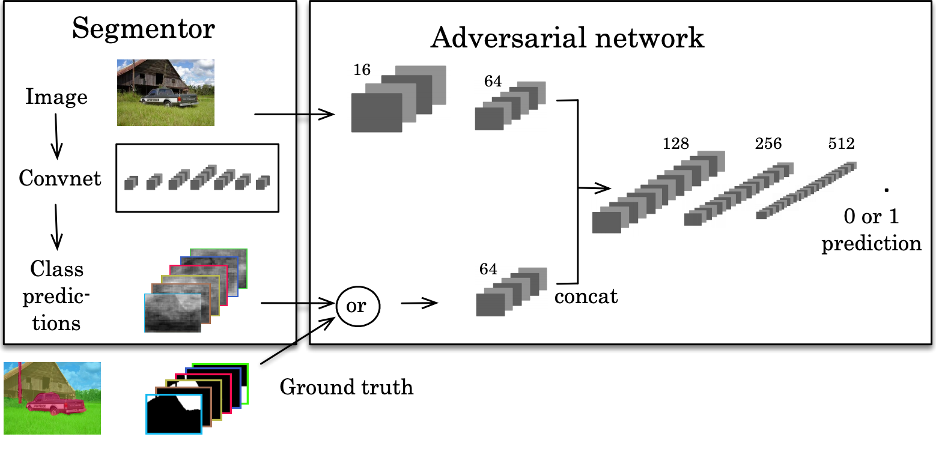}
\caption{Overview of the proposed approach by \citet{luc2016semantic}.}
\label{fig:gan1st}
\end{figure}

\citet{luc2016semantic} complement convolutional semantic segmentation network with an adversarial network, whose job is to discriminate between the ground truth segmentation maps and the output label maps from the semantic segmentation network. Their objective function includes a conventional multi-class cross-entropy loss and an adversarial term, allowing the two networks to be jointly trained (Figure \ref{fig:gan1st}).

\citet{souly2017semi} leverage GANs in a semi-supervised training framework for semantic segmentation. A generator network is designed to generate synthetic data from both noise and class label information from the input image. Given a limited amount of image data with pixel-level labels, a large amount of image data with only image-level class labels, and the fake image data from the generator network, a discriminator network is trained to produce pixel-level confidence maps for each semantic class and assign fake class labels to the generated samples. A novel loss function different from traditional GANs is designed to handle these three types of input data – it includes a term for pixels from labeled data to be classified as one of the available classes, a term for decreasing the probability of pixels from unlabeled data to be classified as a fake class, and a term for distinguishing real image data from fake samples produced by the generator \citep{souly2017semi}.

\citet{hung2018adversarial} use another semi-supervised learning approach for improving semantic segmentation. The segmentation network is trained by minimizing a multitask loss composed of a cross-entropy loss, an adversarial loss, and a semi-supervised loss. The standard cross-entropy loss is computed for the ground truth and the predicted segmentation map. The predicted segmentation map is passed through a discriminator to produce a confidence map, which is used to semi-supervise the segmentation network during training by providing signals on the quality of predictions. In addition, the segmentation network is supervised by the adversarial loss from the discriminator network, which allows the segmentation network to learn to produce from unlabeled data segmentation outputs with distributions close to that of the ground truth.

\textbf{SegAN} \citep{xue2018segan} is a framework designed for medical image segmentation. It includes a segmentor and a critic network trained in an alternating fashion. The segmentor S is a fully convolutional neural network with skip connections. The critic C is trained by maximizing a multiscale L1 loss function between predicted and ground truth segmentation. S aims to minimize the same objective function. During training, S is first fixed while C is trained for one step with gradients from the loss function, and C is fixed while S is trained with the gradients passed from C from the same loss function. These two steps alternate to train the S and C networks in an adversarial fashion. The multiscale feature loss enforces the networks to learn hierarchical features by capturing long- and short-range spatial relations between pixels \citep{xue2018segan}. This approach achieves competitive performance on the BRATS brain tumor segmentation \citep{menze2014multimodal} dataset, outperforming U-net \citep{ronneberger2015u}.

In \citet{chartsias2017adversarial}, cardiac MR images from CT images were generated with a conditional GAN to boost segmentation accuracy. Similar works leveraging GAN for synthetic image generation include \citet{zhang2018translating}, \citet{shin2018medical}, \citet{changhee2019learning}, \citet{yang2018mri}, \citet{yu20183d}, and \citet{abhishek2019mask2lesion}. These methods are widely applied to medical image segmentation, but have also been used for natural images such as remote sensing imagery \citep{mohajerani2019cloudmaskgan}.

\section{Meta-learning for Network Structures}
Design of network architectures has significant impact on its performance. Many works discussed in this review are based on human-invented architectures, but recently, meta-learning methods have shown promising results for image classification tasks. These methods produce network structures dynamically by conducting a graph search. Various graph structures and search algorithms have been designed to optimize the efficiency of the process and the structure of the resulting network. Given sufficient computational resources, these methods have the potential to generate novel and powerful network architectures that can surpass human designs and achieve superior predictive accuracy.

\subsubsection{DCNAS}
\begin{figure}[h]
\centering
\includegraphics[width=\textwidth]{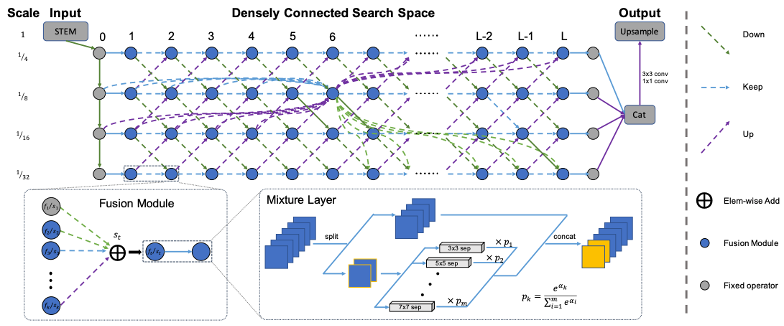}
\caption{The DCNAS framework. From \citet{zhang2021dcnas}.}
\label{fig:dcnas}
\end{figure}

The Densely Connected NAS (DCNAS) framework \citep{zhang2021dcnas} is a proxyless searching paradigm aimed to allow direct search of optimal network structure for semantic image segmentation within a densely connected space while meeting computational demands. The densely connected search space (DCSS) consists of mixture layers that sample efficiently from a collection of operators with a portion of features and fusion modules that aggregate semantic features successively from previous layers while sampling a portion of connections to save GPU memory (Figure \ref{fig:dcnas}). DCNAS can successfully discover optimal architectures for semantic segmentation that surpass human designs. It achieves competitive performances on Cityscapes \citep{cordts2016cityscapes} and PASCAL VOC 2012 \citep{everingham2010pascal} datasets.
\vspace{0.5\baselineskip}

\vspace{-\baselineskip}

\subsubsection{DPC}
\citet{chen2018searching} explore the potential of meta-learning for dense image prediction tasks including semantic segmentation and scene parsing. They demonstrate the effectiveness of meta-learning by constructing Dense Prediction Cell (DPC), a recursive search space represented by a DAG encoding multiscale context information, and using an efficient random search algorithm \citep{golovin2017google} to identify architectures that achieve superior performance than human-invented architectures. They achieve competitive performance on Cityscapes \citep{cordts2016cityscapes}, PASCAL-Person-Part \citep{chen2014detect}, and PASCAL VOC 2012 \citep{everingham2010pascal} benchmark datasets.

\section{Domain Adaptation}
Recently, as practical applications for semantic segmentation continue to grow, another area of research that has gained interest is domain adaptation. Algorithms developed for segmentation under normal circumstances could suffer from significant accuracy downgrade when applied to adverse conditions such as nighttime, accident scene, or foggy weather due to the significant domain shift in the image data. In order to allow the models to better generalize, image data are augmented, transformed, or synthesized to improve the robustness of segmentation models to unseen domains. Generative Adversarial Networks (GANs) are also particularly useful for generating artificial images that complement the original training data to help the models achieve better performance.

\citet{dai2018dark} propose the \textbf{DMAda} framework to progressively build the adaptability of the models trained on daytime scenes to nighttime with an intermediate twilight domain as a bridge. Specifially, models trained on daytime scenes are first used to generate semantic labels for twilight images, and these labeled images are used to further finetune the original daytime models to adapt to nighttime images. The authors show that this unsupervised process reduces the domain gap between daytime and nighttime and makes daytime-to-nighttime knowledge transfer feasible \citep{dai2018dark}. \citeauthor{sakaridis2019guided} (\citeyear*{sakaridis2019guided}, \citeyear*{sakaridis2020map}) follow a similar progressive approach, and utilize GANs to construct synthetic images in the process. Other approaches that leverage synthetic stylized images include \citet{nag2019s} and \citet{sun2019see}. \citet{romera2019bridging} explore both options of constructing synthetic nighttime images during training and converting images to daytime as a pre-processing step at inference time. They demonstrate that domain gaps can be considerably reduced with both approaches. In \citet{wu2021dannet}, a multi-target domain adaptation framework named \textbf{DANNet} is built to achieve one-stage adaptation from daytime to nighttime semantic segmentation. \textbf{DANNet} jointly trains an image relighting network that adjusts the intensity distributions and a semantic segmentation network that produces segmentation predictions of the input images from different domains. \citet{lengyel2021zero} propose Color Invariant Convolution \textbf{(CIConv)}, a trainable layer that generates a domain invariant representation of the input image which can be passed into subsequent layers for segmentation prediction, thus achieving zero-shot day-night domain adaptation.

\citet{zhang2021exploring} explore the role of event-based data in accident-scene segmentation. They propose an Event-driven Dynamic Context Network (\textbf{EDCNet}) which captures dynamic context from event-based data and uses them to bridge the domain gap between normal and accident imageries. Additionally, they provide DADA-seg, a pixel-wise annotated dataset containing a variety of critical scenes from traffic accidents \citep{zhang2021exploring}. \citet{luo2022towards} propose the Multi-source Meta-learning Unsupervised Domain Adaptation (\textbf{MMUDA}) framework to allow segmentation transformers to novel unseen domains such as accident scenes. \textbf{MMUDA} leverages 1) Multi-Domain Mixed Sampling (MDMS) to augment the training data with target data appearances (abnormal scenes); 2) meta-learning for domain generalization (MLDG) strategy; and 3) an enhanced HybridASPP replacing the vanilla MLP-based decoder of SegFormer \citep{xie2021segformer} to efficiently extract large regions of global context and long-range dependencies and improve the performance and efficiency of the framework.

In addition to nighttime and accident scene segmentation, other works have also tackled segmentation and scene understanding under foggy-whether conditions \citep{sakaridis2018model, sakaridis2018semantic, dai2020curriculum, hahner2019semantic}, where similar approaches for domain adaptation are explored.

\let\mysectionmark\sectionmark
\renewcommand\sectionmark[1]{}
\section[Semantic Segmentation in Large Field of View (FoV) \\ Images]{Semantic Segmentation in Large Field of View (FoV) Images}
\let\sectionmark\mysectionmark
\sectionmark{Semantic Segmentation in Large Field of View (FoV) Images}
With rapid advancement in visual understanding, parsing a scene is not limited to low FoV images captured with a pinhole camera. This has led to wide FoV image capturing systems such as fisheye cameras, surround view cameras etc. These systems provide a 360\degree~holistic understanding of the visual world and their applications range from self-driving cars, security cameras to atmospheric inspection from satellite images. However, segmenting these images requires expensive training data so various workarounds are proposed in the research community to alleviate this issue. 

\subsection{Fisheye Segmentation}
A fisheye camera captures scenes with an 180\degree~FoV or sometimes more. This results in more information captured compared to traditional pinhole cameras. \citet{deng2017cnn} are the first to propose a system for fisheye image segmentation where the authors begin with a pinhole segmentation dataset \citep{cordts2016cityscapes} and convert it to a synthetic one. They refer to this process as zoom augmentation. This is then used to train a deep CNN-based model. \citet{blott2018semantic} propose a fisheye segmentation framework which does not require an expensive training data generation step. Their core contribution is to use a projection model transformation to convert rectilinear images to fisheye images and design a CNN based model to train on this dataset. \citet{deng2019restricted} propose Restricted Deformable Convolution (RDC) to address the large distortion issue in fisheye images. Their proposed multi-task learning architecture trains an end-to-end segmentation model with a new loss weighting method for real images and the transformed images. They test this system using surround-view cameras (four fisheye cameras providing 360\degree~view of the scene). \citet{ye2020universal} propose a seven degree of freedom (DoF) augmentation framework for transforming rectilinear images to synthetic fisheye images. This is then applied to road scene segmentation on urban driving images. Similarly, \citet{dufour2020instance} propose an augmentation method for instance segmentation on fisheye images. 

\subsection{Panoramic Segmentation}
Another type of visual scene is captured by a panoramic annular lens which creates less distortion and is more compact than fisheye optical system.  \citet{yang2019can} propose the Panoramic Annular Semantic Segmentation (\textbf{PASS}) framework that segments images from a single panoramic annular lens camera with a distortion of less than 1\%. They train with pinhole camera images and pair it with a network adaptation strategy to use the similarity between pinhole features and the features from panoramic segments with a similar FoV. An improvement to PASS is proposed by \citet{yang2020ds} using attention between the low-level encoder layers and the high-level decoder layers. Their proposed model is more efficient than PASS and they apply their method to robotics as well as vehicles. \citet{yang2020omnisupervised} leverage omnisupervised learning in a multi-task framework which was extended by \citet{yang2021capturing}. Doing away with the issue of not taking global context into account and high latency of PASS/DS-PASS methods, \citet{yang2021capturing} propose Efficient Concurrent Attention Networks (\textbf{ECANets}) that capture omni-range dependencies across the 360\degree~FoV. The model also integrates unlabeled full-size panoramic images into the training strategy. They also contribute WildPASS dataset which comprises panoramic images collected from 6 continents and 65 cities. \citet{yang2021context} extend the context-aware omnisupervised segmentation model to work in a single pass. Their model jointly extracts vertical and horizontal contextual information in panoramas to integrate them in the multi-source omnisupervised training scheme. \citet{zhang2022bending} propose a distortion-aware transformer-based framework named Transformer for Panoramic Semantic Segmentation (\textbf{Trans4PASS}) integrating deformable patch embedding and deformable MLP modules to handle distortions and mix patches. They also employ Mutual Prototypical Adaptation, a domain adaptation framework that transfers knowledge from the label-rich pinhole to the label-scarce panoramic domain \citep{zhang2022bending}.

There are several works exploring panoramic segmentation from a domain adaptation perspective \citep{ma2021densepass,zhang2021transfer}. Other applications of panoramic segmentation include aerial imagery \citep{sun2021aerial,wang2022high} and panoptic segmentation \citep{jaus2021panoramic}. 

\section{Polarization Driven Semantic Segmentation}
For effective and robust segmentation, using RGB cameras alone is not sufficient. Low illumination, mirror-like reflections, poor visibility due to rain and fog can severely degrade segmentation performance \citep{li2022unconventional}. One of the emerging imaging technologies that are used for segmentation is a polarization cameras. Polarization is a property that refers to the orientation of light wave's oscillation. Natural light can be converted to polarized light via reflection from surfaces \citep{li2022unconventional}. Imaging sensors can consist of a polarizer that polarizes light in a particular orientation - which is then easy to analyze in vision systems. This has already shown improvement in car detection on roads \citep{blin2019road,blin2020new}. Other works have demonstrated empirical improvement in car and window segmentation based on polarization features \citep{blanchon2019outdoor,xiang2021polarization}. \citet{xiang2021polarization} propose a polarization driven architecture called EAFNet for semantic segmentation by fusing the polarization and conventional features. They also propose a new RGB-P dataset that comprise of 394 annotated pixel aligned RGB-polarization images. Other modalities (like thermal and depth information) are fused along with polarization in \citet{yan2021nlfnet} for multimodal semantic segmentation. 

\section{Real-time Semantic Segmentation}
Balancing the trade-off between inference accuracy and speed can be difficult to achieve. A lot of efforts have been invested into real-time segmentation, as many modern applications such as robot navigation and autonomous driving have high demand for both speed and accuracy. Most of these efforts have focused on the design of 1) lightweight architectures, and 2) multi-branch networks and fusion modules for efficient capturing of contextual information.

\textbf{E-Net} \citep{paszke2016enet} is one of the earliest semantic segmentation model that meets the demand for real-time inference by using a compact encoder-decoder architecture. \citet{nekrasov2018light} identify and modify computationally expensive blocks in the original RefineNet \citep{lin2017refinenet} to reduce the number of parameters and operations. \textbf{EDANet} \citep{lo2019efficient} improves both segmentation accuracy and speed by employing Point-wise convolution layer, assymetric convolution, and dilated convolution to reduce the number of parameters and the computational load. \textbf{LEDNet} employs an asymmetric encoder-decoder architecture, an attention pyramid network (APN), and introduces two lightweight operations - channel split and shuffle, in each residual block of the network to balance the computation cost and segmentation performance \citep{wang2019lednet}. \citet{li2020semantic} propose \textbf{SFNet} to more effectively and efficiently attain high resolution features with strong semantic representation through the Flow Alignment Module (FAM) that learns the Semantic Flow between feature maps of adjacent levels and efficiently fuses high-level and low-level features. Other models that achieve both fast inference while maintaining segmentation performance through lightweight architectures include \citet{wang2019esnet, wu2020cgnet, treml2016speeding, mehta2018espnet,romera2017erfnet, peng2022pp, nirkin2021hyperseg, li2019dfanet,lin2020graph}.

\textbf{BiSeNet} \citep{yu2018bisenet} is a two-branch model incorporating a Spatial Path for preserving spatial information, a Context Path for obtaining sufficient receptive field, and a Feature Fusion Module for combining features effectively. It aims to addresses the challenge of achieving fast inference while maintaining spatial resolution. The Image Cascade Netork (\textbf{ICNet}) leverages multi-resolution branches and a feature fusion unit with label guidance to efficiently combine semantic information from low resolution and details from high-resolution images \citep{zhao2018icnet}. \textbf{LDFNet} \citep{hung2019incorporating} employs a two-branch architecture that incorporates luminance, depth, and color information from RGB-D images in a fusion-based network. \citet{hong2021deep} propose a family of efficient backbone architectures with deep dual resolution branches and multiple bilateral fusions named \textbf{DDRNets} and a Deep Aggregation Pyramid Pooling Module (DAPPM) for extraction of rich contextual information to improve segmentation performance while maintaining high inference speed. \citet{xu2022pidnet} design a three-branch network named \textbf{PIDNet} to address the issue of ``overshoot'', where low-level features are easily overwhelmed by surrounding contextual information during feature fusion. The three branches are responsible for parsing detailed, context, and boundary information respectively, and detailed and context information are fused with the guidance of boundary attention \citep{xu2022pidnet}. Additional works that incorporate multi-branch structures or efficient feature fusion modules for real-time semantic segmentation include \citet{zhuang2019shelfnet, gao2021rethink, orsic2019defense,peng2022pp}.

\section{Discussion}
Deep learning-based approaches have demonstrated much potential in solving the semantic segmentation problem in the recent few years. Since the development of FCN \citep{long2015fully}, different approaches have focused on improving the output quality of semantic predictions and addressing various other challenges such as the need for faster computations, reduced resource consumption, and better ground-truth datasets. Four major areas of exploration have been observed during the development of these algorithms and are discussed below. 

\begin{enumerate}
\item \textbf{Effective integration of multi-scale and multi-range contextual features to improve prediction accuracy}. Contextual information plays an essential role in object localization and classification. Various combinations of convolutional kernels capturing different field-of-views make it possible for the models to gather multiscale features, whereas deep learning architectures such as LSTM and attention allow the model to learn multi-range features. Novel architectures and configurations have continued to emerge as the state-of-the-art, which signals that there are still many potentials for more optimizations in this area.

\item \textbf{Efficient utilization of training datasets and generation of high-quality artificial datasets to strengthen model robustness and reduce overfitting}. Due to the nature of the task, semantic segmentation requires high-quality annotated datasets that are time-consuming to build. Given the limited quantity of high-quality datasets, it is important that they are utilized to the full extent. Artificial datasets have also emerged as a popular alternative for training semantic segmentation models, especially given the power of GANs in generating realistic images and precise annotations. Moreover, unsupervised and semi-supervised approaches have started to shape up a part of the segmentation landscape.

\item \textbf{Various strategies for saving resources and pushing for real-time inference}. Deep-learning based semantic segmentation models are resource-hungry and time-consuming to train due to the dense nature of the prediction task. Efforts have been made at various sections of the network architectures, but balancing the speed and accuracy of these models remains a challenging topic. 

\item \textbf{Automatic exploration of optimized network architectures to eliminate the need for manual design and experimentation}. This is a promising area as the space of possible network architectures are infinite and large-scale experimentation are nearly impossible due to memory constraints. Intelligent and automatic exploration of the search space has potential to bypass these constraints and generate models that can achieve superior performance.
\end{enumerate}

Table \ref{tab:semantic} summarizes the main contribution, benchmark performances, and categorization of the popular semantic segmentation methods. As seen through the review of the approaches, there remain a lot of potentials in each of these areas of focus, but many key foundations have been laid out for future developments.

\begin{center}
\scriptsize
\begin{longtable}{|L{1.1cm}|L{3.8cm}|L{3.6cm}|L{1.4cm}|} 
\caption{Summary of Popular Semantic Segmentation Methods.} \label{tab:semantic} \\

\hline \textbf{Name} &                                                                                                                                                           \textbf{Main Contribution} &                                                                                                                                                         \textbf{Benchmarks Performances (default metric: mIoU (\%))} &                  \textbf{Categorization} \\
                               \hline
\endfirsthead

\caption* {\textbf{Table 3.1 Continued}}\\

\hline \textbf{Name} &                                                                                                                                                           \textbf{Main Contribution} &                                                                                                                                                         \textbf{Benchmarks and Performances (default metric: mIoU (\%))} &                  \textbf{Categorization} \\
                               \hline
\endhead

\hline \multicolumn{4}{|r|}{{Cont.}} \\ \hline
\endfoot

\hline \hline
\endlastfoot

                                                                FCN &                                                                                                                   End-to-end dense learning with fully convolutional layers &                                                                                                           PASCAL VOC (62.2), NYUDv2 (34.0), SIFT Flow (39.5), PASCAL-Context (53.5) &                       FCN-based \\ \hline
                              U-Net &                                                                                                                                                            Skip connections &                                                                                                                                                  PhC-U373 (92.03), DIC-HeLa (77.56) &                       FCN-based \\ \hline
                          DeconvNet &                                                                                                                         First to use deconvolution to generate feature maps &                                                                                                                                                                    PASCAL VOC (72.5) &                   Deconvolution \\ \hline
                             SegNet &                                                                                                                                                             Pooling indices &                                                                                                                                                                    SUN-RGBD (31.84) &                   Deconvolution \\ \hline
                                SDN &                                                                   SDN unit - shallow deconvolutional network with two downsampling blocks followed by two upsampling blocks &                                                                                                                                     PASCAL VOC (86.6), CamVid (71.8), GATECH (55.9) &                   Deconvolution \\ \hline
                            LSD-Net &                                                                                                                                        Boundary refinement and RGB-D fusion &                                                                                                                                                                 NYU-Depth v2 (45.9) &                   Deconvolution \\ \hline
                                FPN &                                                                                                                      Feature pyramids with top-down and bottom-up pathways  &                                                                                                                                                               MS COCO val (AR=48.1) & Multi-resolution Features-based \\ \hline
                             PSPNet &                                                                            Pyramid Pooling Module that takes global information into account while making local predictions &                                                                                                                                                PASCAL VOC (85.4), Cityscapes (80.2) & Multi-resolution Features-based \\ \hline
                              DMNet &                                                                                                           Dynamic Convolutional Module that generates context-aware filters &                                                                                                                            PASCAL VOC (84.4), ADE20K (45.50), PASCAL-Context (54.4) & Multi-resolution Features-based \\ \hline
                                CCN &                                                                                                        Selective aggregation of appropriate scale features with gated sum &                                                                                                                           Pascal Context (51.6), SUN-RGBD (47.1), COCO Stuff (35.7) & Multi-resolution Features-based \\ \hline
                               MSCI &                                                   Contextual information exchange through bidirectional connections between high-resolution and low-resolution feature maps &                                                                                                            PASCAL VOC (88.0), NYUDv2 (49.0), PASCAL-Context (50.3), SUN-RGBD (50.4) & Multi-resolution Features-based \\ \hline
                           ParseNet &                                                                                                                   Global pooling and simple global feature map incoporation &                                                                                                                          PASCAL VOC (69.8), SIFT Flow (40.4), PASCAL-Context (40.4) & Multi-resolution Features-based \\ \hline
                          RefineNet &                                                              Residual Convolution Units for end-to-end training and chained residual pooling for multi-scale context fusion &                                                                                                            PASCAL VOC (83.4), ADE20K (40.7), PASCAL-Context (47.3), SUN-RGBD (45.9) & Multi-resolution Features-based \\ \hline
                               FaPN &                                                        Feature alignment between top-down and bottom-up pathways with Feature Selection Module and Feature Alignment Module &                                                                                                                                 ADE20K(56.7), Cityscapes (80.0),  COCO-Stuff (40.6) & Multi-resolution Features-based \\ \hline
                             ExFuse &                                                                                                           More effective multi-scale feature fusion with semantic embedding &                                                                                                                                                                   PASCAL VOC (87.9) & Multi-resolution Features-based \\ \hline
                             APCNet &                                                   Adaptive Context Module that estimate Global-guided Local Affinity coefficients for dynamic weight assignments for pixels &                                                                                                                            PASCAL VOC (84.2), ADE20K (45.38), PASCAL-Context (54.7) & Multi-resolution Features-based \\ \hline
                         DilatedNet &                                                                             Aggretation of contextual information through dilated convolutions with multiple sampling rates &                                                                                                                                                                   PASCAL VOC (75.3) &        Atrous convolution-based \\ \hline
                          DeepLab3+ &                                                     Segmentation quality improvement with atrous convolution, atrous spatial pyramid pooling, and conditional random fields &                                                                                                                                                                   PASCAL VOC (89.0) &        Atrous convolution-based \\ \hline
                    ResNet-DUC-HDC  &                 Solve ``Gridding'' issue from dilated convolutions using hybrid dilated convolutions; achieve improved pixel-level decoding with Dense Upsampling Convolution &                                                                                                                                                                   PASCAL VOC (83.1) &        Atrous convolution-based \\ \hline
                          DenseASPP &                                                                                         Dense multi-scale feature map generation with densely connected atrous convolutions &                                                                                                                                                                   Cityscapes (80.6) &        Atrous convolution-based \\ \hline
                               GALD &                                                                            Improves small pattern segmentation with Global Aggregation Module and Local Distribution Module &                                                                                                                                           Cityscapes (83.3), MS-COCO (Mask AP=37.8) &        Atrous convolution-based \\ \hline
                               MRFM &                                                                                                          Adaptive selection of receptive field and edge-aware loss function &                                                                                                                                                PASCAL VOC (88.4), Cityscapes (83.0) &        Atrous convolution-based \\ \hline
                              DANet &                                                                                                              Dual-attention modules for incoporating contextual information &                                                                                                      Cityscapes (81.5), PASCAL VOC (82.6), PASCAL-Context (52.6), COCO Stuff (39.7) &                 Attention-based \\ \hline
                             PSANet &                                                                                                       Distribute and collect attention branches for feature representations &                                                                                                                                ADE20K (43.77), PASCAL VOC (85.7), Cityscapes (81.4) &                 Attention-based \\ \hline
                             GT-OCR &                                                         Aggregation of object region representations based on the relationship between object regions and contextual pixels &                                                                                             Cityscapes (84.5), LIP (56.65), PASCAL-Context (56.2), ADE20K(45.66), COCO Stuff (40.5) &                 Attention-based \\ \hline
                          SegFormer &                                                                                                                        Hierarchical Transformer encoder and all-MLP decoder &                                                                                                                                  Cityscapes (84.0), ADE20K(50.3), COCO-Stuff (46.7) &                 Attention-based \\ \hline
                            EMA-Net &                                                                          Formulation of attention into Expectation-Maximization for efficient computation of attention maps &                                                                                                                         PASCAL VOC (88.2), PASCAL-Context (53.1), COCO-Stuff (39.9) &                 Attention-based \\ \hline
                                DFN &                         Application of different weights to different channels using Channel Attention Block to improve intra-class consistency and inter-class distinction &                                                                                                                                                PASCAL VOC (86.2), Cityscapes (80.3) &                 Attention-based \\ \hline
                              SANet &                                                                                                                               Class-wise mask generation for pixel grouping &                                                                                                                                            PASCAL VOC (86.1), PASCAL-Context (54.4) &                 Attention-based \\ \hline
                               Swin &                                                                         Efficient computation of hierarchical feature maps with shifted windows and effective patch merging &                                                                                                                                               ADE20K (53.5), MS-COCO (Mask AP=51.1) &                 Attention-based \\ \hline
                              CCNet &                                                                                               Improves efficiency of attention mechanism with criss-cross path connections &                                                                                                                                                   ADE20K (45.22), Cityscapes (81.4) &                 Attention-based \\ \hline
\citet{chen2016attention}  &                                                                          Hierarchical attention-based approach that learns relative attention mask between adjacent scales  &                                                                                                                                            Cityscapes test (85.1), Mapillary (61.1) &                 Attention-based \\ \hline
                         MaskFormer &                                                                                                Solves semantic and instance segmentation in unison with mask classification &                                                                                                                                 ADE20K (55.6), Cityscapes (78.5), COCO-Stuff (37.1) &                 Attention-based \\ \hline
                         Mask2-Former &                                                                       Masked attention that restricts the attention to localized features centered around predicted segment &                                                                                                                                                    ADE20K (56.4), Cityscapes (83.3) &                 Attention-based \\ \hline
                            CRF-RNN &                                                                                   Integrates CRF modeling with CNN by formulating CRF as RNN using Mean Field Approximation &                                                                                                                                                                   PASCAL VOC (74.7) &                Graphical models \\ \hline
                          Piecewise & Formulates CNN-based general pairwise potential functions in CRFs to explicitly model patch-patch semantic relations; approximate training using piecewise training of CRFs &                                                                                                                                  PASCAL VOC (78.0), NYUDv2 (40.6), SIFT-flow (44.9) &                Graphical models \\ \hline
                                DPN &                           Solves Markov Random Fields with a single CNN while incoporating high-order relations and mixtures of label contexts with pairwise terms modeling &                                                                                                                                                                   PASCAL VOC (77.5) &                Graphical models \\ \hline
                               GCRF &                                               Unification of Mean Field algorithm with CNN using Gaussian CRFs and fixed number unfolding of Gaussian mean field iterations &                                                                                                                                                                   PASCAL VOC (73.2) &                Graphical models \\ \hline
                   Domain Transform &                                                                                              Boundary learning with domain transfer edge-preserving filter with unified CNN &                                                                                                                                                                   PASCAL VOC (71.8) &      Other contextual knowledge \\ \hline
                              GSCNN &                                                                                                                            Boundary detection improvement with shape stream &                                                                                                                                                              Cityscapes test (82.8) &      Other contextual knowledge \\ \hline
                             EncNet &                                                Context Encoding Module that captures the semantic context of scenes and selectively highlights class-dependent feature maps &                                                                                                                            PASCAL-Context (51.7), PASCAL VOC (85.9), ADE20K (44.65) &      Other contextual knowledge \\ \hline
                              ReSeg &                                  Incorporation of contextual information with RNN layers that performs horizontal and vertical sweeps of the image and hidden layer outputs &                                                                                                                        CamVid (58.8), Weizmann Horses (91.6), Oxford Flowers (93.7) &                       RNN-based \\ \hline
                            DAG-RNN &                                                                                                  Leveraging long-range contextual information with directed acyclic graphs  &                                                                                SIFT Flow (global pixel acc=85.3), CamVid (global pixel acc=91.6), Barcelona (global pixel acc=74.6) &                       RNN-based \\ \hline
                            2D LSTM &                                                                                                                       Capturing global and local dependencies with 2D LSTM  &                                                                                                                                              SIFT Flow (70.11), Stanford BG (78.56) &                      LSTM-based \\ \hline
                            LSTM-CF &                           Capturing and fusion of contextual information from multiple channels of photometric and depth data with Long Short-Term Memorized Context Fusion &                                                                                                                                                      SUN-RGBD (48.1), NYUDv2 (49.7) &                      LSTM-based \\ \hline
                         Graph LSTM &                                                                                                                       Modeling natural visual pattern flows with graph LSTM &                                                                                                                                                          PASCAL-Person-Part (60.35) &                      LSTM-based \\ \hline
                   \citet{luc2016semantic} &                                             Incorporation of GAN to discriminate between ground truth segmentation maps and the output label maps from segmentation network &                                                                                                                                               PASCAL VOC (73.3), Stanford BG (72.0) &                       GAN-based \\ \hline
                 \citet{souly2017semi} &                                   GAN-based CNN leveraging limited pixel-level annotations, large amount of image-level class labels, and fake data from generative network & all labeled + unlabeled PASCAL VOC (65.8), fully labeled + 2000 unlabeled SIFT Flow (35.1), all labeled + 10k unlabeled Stanford BG (63.3), unlabeled + 11k unlabeled CamVid (58.2) &                       GAN-based \\ \hline
                  \citet{hung2018adversarial} &                               Segmentation network trained by minimizing a multi-task loss composed of a cross-entropyloss, an adversarial loss, and a semi-supervised loss &                                                                                                                                                                   PASCAL VOC (74.9) &                       GAN-based \\ \hline
                              SegAN &                                                                                                            Segmentor and a critic network trained in an alternating fashion &                                                                                                                                                   BRATS 2015 test (Dice score=85.0) &                       GAN-based \\ \hline
                              DCNAS &                                                                 Direct search of optimal network structure for semantic image segmentation within a densely connected space &                                                                                                                           Cityscapes test (84.3), PASCAL VOC (86.9), ADE20K (47.12) &                   Meta-learning \\ \hline
                                DPC &                                                         Recursive search space for optimal network structure represented by a DAG encoding multi-scale context information  &                                                                                                               Cityscapes test (82.7), PASCAL VOC (87.9), PASCAL-Person-Part (71.34) &                   Meta-learning \\ \hline
                                DMAda &                                  Progressively build the adaptability of the models trained on daytime scenes to nighttime with an intermediate twilight domain as a bridge &                                                                                                                                                              Dark Zurich-test(36.1) &               Domain Adaptation \\ \hline
                             DANNet &                                                                                   Multi-target domain adaptation via joint training of relighting and segmentation networks &                                                                                                                                                              Dark Zurich-test(47.7) &               Domain Adaptation \\ \hline
                             CIConv &                                                                                        Trainable layer that generates a domain invariant representation of the input image  &                                                                                                                                                              Dark Zurich-test(41.6) &               Domain Adaptation \\ \hline
                             EDCNet &                                                               Captures dynamic context from event-based data to bridge the domain gap between normal and accident imageries &                                                                                                                                              Cityscapes val (69.4), DADA-seg (28.3) &               Domain Adaptation \\ \hline
                              MMUDA &                                                            Uses Multi-Domain Mixed Sampling (MDMS) to augment the training data and meta-learning for domain generalization &                                                                                                                                                                    DADA-seg (46.97) &               Domain Adaptation \\ \hline
                              E-Net &                                                                                                                                        Compact encoder-decoder architecture &                                                                                                                                                              Cityscapes test (58.3) &                       real-time \\ \hline
                             EDANet &                                                                                  Lightweight point-wise convolution layer, assymetric convolution, and dilated convolution  &                                                                                                                                                              Cityscapes test (67.3) &                       real-time \\ \hline
                             LEDNet &                                                                                                                         Lightweight asymmetric encoder-decoder architecture &                                                                                                                                                              Cityscapes test (70.6) &                       real-time \\ \hline
                              SFNet &                    Flow Alignment Module (FAM) that learns the Semantic Flow between feature maps of adjacent levels and efficiently fuses high-level and low-level feature &                                                                                                                                                              Cityscapes test (80.4) &                       real-time \\ \hline
                            BiSeNet &                                Two-branch model incorporating a Spatial Path for preserving spatial information and a Context Path for obtaining sufficient receptive field &                                                                                                                                                              Cityscapes test (74.7) &                       real-time \\ \hline
                              ICNet &                                                                                   Combines semantic information from low resolution and details from high-resolution images &                                                                                                                                                              Cityscapes test (70.6) &                       real-time \\ \hline
                             LDFNet &                                                                         Two-branch architecture that incorporates luminance, depth, and color information from RGB-D images &                                                                                                                                                              Cityscapes test (71.3) &                       real-time \\ \hline
                             DDRNet &                                                             A  family of efficient backbone architectures with deep dual resolution branches and multiple bilateral fusions &                                                                                                                                                              Cityscapes test (77.4) &                       real-time \\ \hline
                             PIDNet &                                                                                                Three-branch network for parsing detailed, context, and boundary information &                                                                                                                                                              Cityscapes test (80.6) &                       real-time \\ \hline
\end{longtable}
\end{center}

\chapter{Deep Models for Instance Segmentation}
\enlargethispage{-\baselineskip}
Instance segmentation adds another component to semantic segmentation - rather than assigning the same label to the pixels in an image belonging to the same object category, each instance of an object class also needs to be distinguished from each other. Instance segmentation can be considered as a combination of object detection and semantic segmentation. It takes the same form of input image as semantic segmentation but produces a richer output label map where the pixels belonging to each distinct instance of the same object category are labeled with distinct instance ID. The mathematical formulation of the instance segmentation problem resembles that of semantic segmentation: given an input image $\mathcal{P}={p\textsubscript{1},p\textsubscript{2}, ..., p\textsubscript{m}}$, and a label space $\mathcal{L}={l\textsubscript{1},l\textsubscript{2}, ..., l\textsubscript{n}}$, it assigns an element from $\mathcal{L}$ to each element of $\mathcal{P}$. However, this label space is also instance-aware, meaning that each instance of the same object category has a separate label $p_i$ in $\mathcal{P}$. Additionally, pixels that do not belong correspond to ``Things'' (defined below) are assigned a ``dummy'' (background) label, or can simply be ignored. Finally, bounding boxes of each object instance can be included in or derived from the output. 

The keen reader will notice that certain objects usually cannot have instances, such as ``csky'' and ``wall'', which are defined as ``Stuff'', and are usually not assigned an instance label. Conversely, objects such as ``Chairs'' and ``Tables'' that are countable and are usually labelled during instance segmentation are defined as ``Things''. These two concepts are useful for understanding the difference between the various types of segmentation problems. Semantics segmentation focuses on ``Stuff''; instance segmentation focuses on ``Things''; whereas panoptic segmentation introduced in Section \ref{chap6} tackles both types of objects simultaneously. Instance segmentation is widely applied in many similar areas as semantic segmentation, but it is particularly useful in satellite imagery and tumor detection (whereas semantic segmentation is often used for segmenting organs and cataract surgery instruments). In this section, methods developed to solve instance segmentation are reviewed.

\section{R-CNN-based Methods (Two-stage)}
R-CNN-based object detection models have been extended for this task and remain to be the most popular family of models for instance segmentation. These models separate the task of detection and classification. For instance segmentation, the additional task of object mask generation is also performed. Region Proposal is a key element in this line of approaches as it generates candidates for downstream classification of object masks. Mask generation and classification are achieved based on the features encoded through the CNN layers. These methods are typically computationally more expensive than the one-stage methods discussed in the next section but achieve better performance as well.

The Simultaneous Detection and Segmentation (\textbf{SDS}) \citep{hariharan2014simultaneous} is developed based on R-CNN and used for instance segmentation. It replaces selective search with a faster Multiscale Combinatorial Grouping (MCN) \citep{arbelaez2014multiscale} algorithm for region proposals. Each proposed region is then processed through two parallel streams – one generating a feature vector for bounding box and the other a feature vector that masks background pixels. The two feature vectors are combined and an SVM is applied to classify the object. Finally, a refinement step composing of non-max suppression and category-specific coarse mask prediction using CNN features is taken to further improve the segmentation outcome. 

\textbf{DeepMask} \citep{pinheiro2015learning} is a discriminative CNN-based approach for generating object proposals for instance segmentation. The idea is originated from R-CNN based approaches, where the two main components are region proposals and object classification. DeepMask is trained with samples of image patches. First, the network generates object segmentation masks without class labels from the input image patch. Then, a score is generated to determine whether the input patch contains an object. Specifically, the patch contains an object if the object is roughly centered and fully contained in the patch. At inference time, the network outputs multiple segmentation masks for a given input image along with the corresponding score indicating the likelihood of an object existing in the patch. DeepMask has superior performance compared to other object proposal algorithms and can be generalized to unseen objects \citep{pinheiro2015learning}. 

\textbf{SharpMask} \citep{pinheiro2016learning} is designed to refine coarse object segmentation masks with an improved feature extraction module based on DeepMask's architecture. The refinement module is based on a top-down structure that progressively augment the encoder feature map output through the lower layers. This approach is much faster than models that leverage skip connections and achieves new state-of-art performance on the COCO instance segmentation dataset \citep{lin2014microsoft}.

Both methods above are developed by the Facebook AI research team to address several challenges associated with the instance dataset \citep{lin2014microsoft}. Specifically, the dataset contains objects with varying scales, occlusions and clutters, and requiring precise localization.

The \textbf{MultiPathNet} \citep{zagoruyko2016multipath} is also R-CNN-based and a derivation from the DeepMask model. Three main modifications are made to further address the challenges associated with varying scales and localization of objects. First, a foveal structure consisting of four pooling regions at different resolutions are used to capture multiscale contexts. Second, skip connections are used to recover spatial information from earlier layers that are lost in Fast R-CNN where the features from the conv5 layer have been down-sampled by a large factor prior to ROI pooling. Third, a novel loss function integrating metrics at multiple IOU thresholds is designed to improve object localization by assigning higher scores to proposals with more overlapped regions with the ground truth. When using DeepMask \citep{pinheiro2015learning} as the object proposal method, this system can be adapted for instance segmentation and achieves 2nd place on the COCO 2015 challenges. It boosts small object detection results by 4x and 66\% overall compared with the Fast R-CNN \citep{girshick2015fast} baseline.

\begin{figure}[h]
\centering
\includegraphics[width=0.9\textwidth]{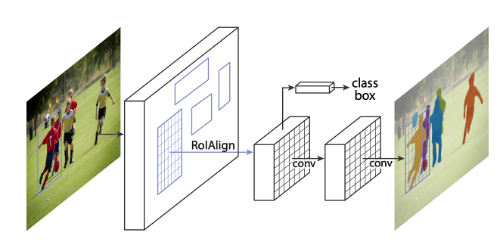}
\caption{Mask R-CNN. From \citet{he2017mask}.}
\label{fig:maskrcnn}
\end{figure}

\enlargethispage{-\baselineskip}
Based on Faster R-CNN and also developed by the Facebook AI team, \textbf{Mask R-CNN} \citep{he2017mask} is one of the most widely used instance segmentation approaches. It adds another branch responsible for predicting object masks in parallel with bounding box regression and object type classification. The loss function combines classification loss, bounding-box loss, and mask loss. For each class label, a binary mask is produced where the value is 1 and 0 for pixels corresponding to the foreground object and background, respectively. Per-pixel sigmoid output for each of the K classes are calculated, and the mask loss is defined as the average binary cross-entropy loss over the K classes (Figure \ref{fig:maskrcnn}). Mask R-CNN has comparable training and inference speed with Faster R-CNN and achieved state-of-art on the COCO \citep{lin2014microsoft} 2015 and 2016 challenge datasets.

\textbf{Mask Scoring R-CNN} \citep{huang2019mask} extends on Mask R-CNN’s \citep{he2017mask} architecture to include an additional MaskIoU head to the network to evaluate the quality of the segmentation mask. The MaskIoU head is trained with simple regression based on the mask head and ROI feature head’s outputs. This head is integrated with Mask R-CNN and the network can be trained from end-to-end. At inference time, the MaskIoU head generates a calibrated mask score to replace the mask confidence. Mask Scoring R-CNN outperforms Mask R-CNN on the COCO \citep{lin2014microsoft} benchmark dataset.

Authors of the Path Aggregation Network (\textbf{PANet}) \citep{liu2018path} propose three additional components to improve the information propagation in Mask R-CNN \citep{he2017mask}. First, they create a bottom-up path to provide localization signals in lower levels to augment the top-down path in the FPN. Second, adaptive feature pooling is used to aggregate information from all levels so that various feature resolutions can benefit each other. Third, tiny fully-connected layers that complement FCN with spatially-adaptable information are used to augment mask prediction at the final layer. These three improvements allow PANet to place 1st in the COCO \citep{lin2014microsoft} 2017 Instance Segmentation challenge and achieve top performance on Cityscapes \citep{cordts2016cityscapes} dataset.

Multi-task Network Cascades (\textbf{MNCs}) \citep{dai2016instance} is a three-stage model solely based on CNNs that solves instance segmentation. Three networks responsible for differentiating instances, estimating masks, and categorizing objects, respectively, share convolutional features and form a cascading structure to predict categorized instance masks in stages. This model can be trained from end-to-end and more stages can be attached to further improve segmentation quality \citep{dai2016instance}.

In \citet{hu2018learning}, the authors develop a novel training paradigm that is partially supervised – requiring labeled masks for only a small fraction of the training dataset. The key component of this paradigm is a weight transfer function that can be trained to map a category’s bounding box detection parameters to its instance segmentation parameters. When given bounding box annotations for all training categories, both the bounding box head and the mask head can be trained jointly from end-to-end. This paradigm is integrated with Mask R-CNN \citep{he2017mask} (thus named \textbf{Mask\textsuperscript{x} R-CNN}) but can be used with any instance segmentation model with a box detection and mask prediction component.

\textbf{MaskLab} is an instance segmentation model built on top of Faster R-CNN and is composed of three separate branches – box detection, semantic segmentation, and direction prediction \citep{chen2018masklab}. The semantic segmentation branch generates logits for pixel-wise segmentation, and the direction prediction branch generates logits for predicting each pixel’s direction toward its corresponding instance \citep{chen2018masklab}. For each region of interest generated from the box detection branch, semantic segmentation logits are cropped from the channel of the predicted class, and directional pooling is used to assemble logits from each channel. The combined cropped features are then passed through 1x1 convolution to produce final instance segmentation.

\citet{hayder2017boundary} design an Object Mask Network (\textbf{OMN}), which infers a representation of the object segment based on distance to object boundary, which is encoded into a multi-valued map and decoded into a final object mask that can go beyond the bounding box. The OMN is integrated into the MNCs \citep{dai2016instance} and replaces the original mask prediction network. This approach achieved state-of-the-art on the PASCAL VOC \citep{everingham2010pascal} 2012 dataset and the Cityscapes \citep{cordts2016cityscapes} dataset in both instance segmentation and object proposal tasks.

\textbf{Cascade Mask R-CNN} \citep{cai2019cascade} is developed based on the fact that the false positives and true positives in object detection problem can be challenging to differentiate and is often done via thresholding using IoU score. The typical IoU threshold of 0.5 leads to issues such as noisy bounding boxes during inference while using a higher threshold instead causes overfitting. The Cascade R-CNN \citep{cai2019cascade} is a multi-stage R-CNN with a sequence of detector structures designed to address these problems. The detectors are trained sequentially, where the output of the previous stage is provided as a training set to the next, which is trained with a higher IoU threshold based on the assumption that the output bounding box from the box regressor is usually better than the input. This resampling process progressively improves bounding box quality while reducing overfitting. Cascade Mask R-CNN performs instance segmentation on top of bounding box detections with an added mask head. Cascade Mask R-CNN achieves superior performance on the COCO \citep{lin2014microsoft} dataset compared to state-of-the-art instance segmentation approaches including Mask R-CNN \citep{he2017mask}.

\section{One-stage Methods}
In contrast to R-CNN based methods that generate bounding boxes before segmentation, these methods directly build upon semantic segmentation to generate instance boundaries and do not require the feature localization step. One key intuition is that a bounding box is a contour in the form of a minimum encompassing rectangle, and by extending the number of possible edges (parameters), better contours can be imposed on object instances to approximate segmentation masks. Many one-stage methods can be directly applied in real-time applications due to their fast inference speeds.

\begin{figure}[h]
\centering
\includegraphics[width=\textwidth]{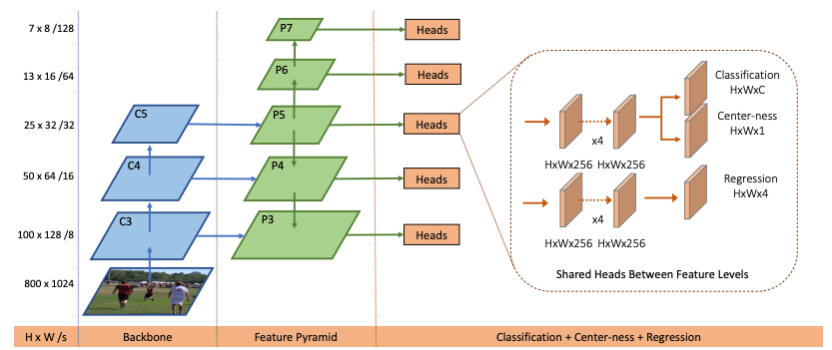}
\caption{The FCOS framework. From \citet{tian2019fcos}.}
\label{fig:fcos}
\end{figure}

The Fully Convolutional One-Stage Object Detection (\textbf{FCOS}) framework \citep{tian2019fcos} is one of the first anchor box and proposal free method for object detection and can be extended to solve instance segmentation with minimum modification. It is composed of a backbone CNN for feature extraction, a Feature Pyramid Network for incorporating multiscale contexts, and five shared object detection heads that optimize an overall loss function which is the sum of classification loss (of location), bounding box regression loss, and the center-ness loss which measures how “centered” is the pixel location in its bounding box \citep{tian2019fcos}. Figure \ref{fig:fcos} shows the overall architecture of FCOS.

\enlargethispage{\baselineskip}
\textbf{ExtremeNet} \citep{zhou2019bottom} generates one center point and four extreme points (top-most, left-most, bottom-most, right-most) for each detection, and fits an octagon to these points to approximate an instance segmentation mask. This segmentation mask can be further refined with Deep Extreme Cut (DEXTR) \citep{maninis2018deep} – a pretrained deep network that can convert a set of extreme points into a segmentation mask. This approach is fast and efficient compared with R-CNN based approaches and achieves competitive performance on the COCO \citep{lin2014microsoft} 2017 dataset.

\textbf{ESE-Seg} \citep{maninis2018deep} is another approach that generates segmentation boundaries based on baseline contours (e.g. bounding box). Specifically, an inner center radius shape signature \citep{xu2019explicit} is designed to parametrize the contour and Chebyshev polynomials \citep{mason2002chebyshev} are used to fit the signatures. 

\textbf{PolarMask} \citep{xie2020polarmask} uses center classification and regression in the polar coordinate to generate masks based on contours. In this case, segmentation is formulated as a regression problem and the Polar IoU Loss \citep{xie2020polarmask} is introduced to optimize the problem.

In \textbf{FourierNet} \citep{riaz2021fouriernet}, five heads are connected to different spatial resolutions of the FPN. These heads predict the centerness, classification scores, and coefficients of Fourier series. An Inverse Fast Fourier Transform (IFFT) algorithm is then applied on the coefficients to obtain the contour points. IFFT is a differentiable shape decoder and allows the optimizer to achieve automatic weight balancing during training. Overall, contour-based methods can produce fast segmentation inference and can be trained efficiently. However, these masks are coarse and cannot match the performance of the R-CNN based methods.

The Mask Encoding based Instance segmentation (\textbf{MEInst}) \citep{zhang2020mask} encodes instance masks into a compact fixed-dimension vector through dictionary learning methods such as PCA. The compression is achieved through removing the redundancy from the original mask. The FCOS \citep{tian2019fcos} pipeline is extended by adding a branch for mask coefficient regression. This approach can also be integrated with other detectors and achieves superior performance than most other one-staged models. In contrast to previous methods discussed in this section, it is not based on contours and thus addresses the problem with artifacts from “disjointed” objects. 

\textbf{TensorMask} \citep{chen2019tensormask} is a novel idea towards instance segmentation with dense mask generation. It uses a tensor representation of segmentation masks over a 4D spatial domain. Two dimensions are used for each possible location of the input image since it follows a sliding window approach. The other two dimensions are used for the mask at each location. Aligned representation and Tensor Bipyramid are introduced to retrieve spatial details during up-sampling. TensorMask achieves comparable results to Mask R-CNN and demonstrates the viability of dense mask prediction (a mask is predicted for each feature map location) \citep{chen2019tensormask}, but takes significantly longer to train.

\textbf{YOLACT} \citep{bolya2019yolact} is one of the first real-time instance segmentation models built on one-stage object detection models by incorporating a parallel branch for mask generation. It first generates prototype masks with a simple FCN backbone and per-instance mask coefficients, and then combines these two elements linearly. A sigmoid function is then used to produce the final masks. Cropping and thresholding are applied to the final mask to generate bounding box and segmentation outputs. Masks generated in this fashion are high-quality because they are based on the entire image and avoids loss of information from “repooling” (feature localization) \citep{bolya2019yolact}. YOLACT is extremely efficient due to its lightweight assembly process and can be attached to most object detectors due to its generalizability. \textbf{YOLACT++} \citep{bolya2020yolact} improves the performance of YOLACT through adding a fast mask re-scoring branch and deformable convolutions, and optimizing the prediction head, while still maintaining real-time inference capability.

\textbf{BlendMask} \citep{chen2020blendmask} and \textbf{CenterMask} \citep{lee2020centermask} are two very similar approaches and extends on YOLACT \citep{bolya2019yolact}. BlendMask uses a modified FCOS \citep{tian2019fcos} object detector as backbone. The bottom module is a decoder that extracts base feature score maps, and the top layer is a single convolution layer that predicts attention masks and bounding boxes. The blender module merges these two branches (Fig) to produce final segmentation masks. \citet{chen2020blendmask} also create a compact version of BlendMask named \textbf{BlendMask-RT} specifically for real-time settings. CenterMask \citep{lee2020centermask} uses CenterNet \citep{duan2019centernet} as backbone. It attaches five heads to the backbone network – Shape, Size, Saliency, Heatmap, and Offset \citep{lee2020centermask}. Center points are generated from the Heatmap and Offset heads and are then used to extract feature representation to form local shapes, which also aggregate information from the Shape and Size heads. The Global Saliency branch produces a Global Saliency Map based on the Saliency Head. Final masks are generated by multiplying the cropped Saliency Map and the Local Shapes. \citet{lee2020centermask} also design \textbf{CenterMask-Lite} by downsizing the backbone, box head, and mask head to achieve real-time performance.

\citet{cao2020sipmask} introduce \textbf{SipMask}, which leverages a lightweight spatial preservation (SP) module to enable accurate delineation of spatially adjacent instances and improve mask predictions by generating a separate set of spatial coefficients for subregions of a bounding box. In addition, a mask alignment weighting loss and a feature alignment scheme are introduced to improve correlations between mask predictions and high quality bounding-box detections \citep{cao2020sipmask}. SipMask achieves satisfactory performance in real-time video instance segmentation \citep{cao2020sipmask}.

\textbf{CondInst} \citep{tian2020conditional} is another approach built on YOLACT’s \citep{bolya2019yolact} architecture. It bypasses the need for bounding box generation, and instead of prototype masks, it uses controller head to dynamically generate convolutional filters conditioned on target instances. These filters form the lightweight mask FCN head that is applied to the feature map output from the backbone network to produce final instance masks. Similar to FCOS \citep{tian2019fcos}, shared heads connected to the FPN feature maps also generate class probabilities and center-ness of the target instances. Compared to Mask R-CNN, it achieves both higher accuracy and speed on the COCO \citep{lin2014microsoft} benchmark dataset.

\textbf{SOLO} \citep{wang2020solo} and \textbf{SOLOv2} \citep{wang2020solov2} are a family of novel but extremely simple instance segmentation approaches. SOLO introduces a new notion of “instance category”, which defines a pixel within an instance by the instance’s size and location and based on the intuition that different instances have either different shapes or positions. An object’s location is assigned to a cell in the SxS grid which divides the input image based on the object’s center. The object’s size is determined by the level of Feature Pyramid Network it is assigned to. Thus, SOLO essentially transforms instance segmentation into two pixel-level classification problems – each pixel is assigned to a grid cell and an FPN level. The Mask branch of the network is responsible for instance mask segmentation. For each positive grid cell, an instance mask is predicted with an FCN. The pixel coordinates are added to the FPN’s output features to preserve spatial information since in this architecture, segmentation is formulated as a classification task which is spatially variant \citep{wang2020solo}. SOLO achieves comparable accuracy with Mask R-CNN and outperforms most of the single-shot models discussed above. SOLOv2 \citep{wang2020solov2} improves the mask branch by decomposing mask learning to feature learning and dynamic convolution kernel generation, similar to the approach in CondInst \citep{tian2020conditional}. Additionally, SOLOv2 develops Matrix NMS that runs non-max suppression with parallel matrix operations that significantly boosts mask post-processing speed for real-time settings while also improving its performance. Both SOLO and SOLOv2 can be easily adapted for panoptic segmentation.

In Segmenting Objects with Transformers (\textbf{SOTR}), \citet{guo2021sotr} combine CNN with transformer to tackle instance segmentation. Specifically, two parallel branches are attached to the FPN backbone to perform two subtasks: one branch uses a transformer to predict category for each instance based on the FPN features, capturing long-range context dependencies; the other branch fuses features from a multi-level upsampling module applied to the FPN feature map and generates final segmentation mask with dynamic convolution kernels produced by the transformer branch. A novel twin attention mechanism also replaces the attention matrix in the original transformer \citep{vaswani2017attention} with sparse representation using only a row and a column attention. With this simple but efficient architecture, SOTR achieves state-of-the art instance segmentation performance on the MS COCO \citep{lin2014microsoft} dataset.

\section{Query-based Models}
\begin{figure}[h]
\centering
\includegraphics[width=0.95\textwidth]{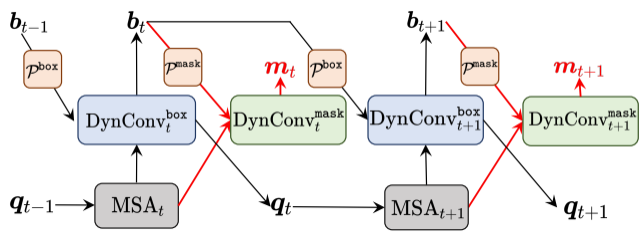}
\caption{The QueryInst framework. From \citet{yang2021tracking}.}
\label{fig:query}
\end{figure}

Query-based methods have recently been proposed to reformulate object detection as a query-based direct set prediction problem given 100 learned object queries as input \citep{yang2021tracking}. \textbf{DETR} \citep{carion2020end} is a query-based method built with transformers. Deformable DETR \citep{zhu2020deformable}, TSP \citep{sun2021rethinking}, ACT \citep{zheng2020end}, SOLQ \citep{dong2021solq}, and UP-DETR \citep{dai2021up} build on DETR and further improve its performance. Sparse R-CNN \citep{sun2021sparse} is a query-based object detection framework built on top of R-CNN architectures. Instance as Queries (\textbf{QueryInst}) \citep{yang2021tracking} is built on Sparse R-CNN, but can also be based on any multi-stage query-based detector. In addition to the query-based detector, QueryInst consists of six parallel dynamic mask heads that transform each mask RoI feature adaptively according to the corresponding query (Figure \ref{fig:query}), and are simultaneously trained in all stages \citep{yang2021tracking}. QueryInst is driven by two insights: 1) there is an intrinsic one-to-one correspondence across different stages, and 2) a one-to-one correspondence between mask RoI features and object queries in the same stage \citep{yang2021tracking}. This approach eliminates the need for multi-stage mask head connection and the issues of proposal distribution inconsistency \citep{yang2021tracking}. It outperforms multi-stage schemes such as Cascade Mask R-CNN \citep{he2017mask} and HTC \citep{chen2019hybrid}.

\section{Other Methods}
\subsubsection{Deep Watershed Transform}
\citet{bai2017deep} combine deep learning with the traditional watershed transform algorithm to achieve instance segmentation. Grayscale images can be considered as a topographic surface \citep{bai2017deep}. Traditional watershed transform is based on flooding this surface from its minima while preventing merging of waters from multiple sources, the image can be partitioned into components \citep{bai2017deep}. However, this often over-segments the image and cannot effectively address instance segmentation. This paper develops a Watershed Transform Net that learns an energy map with deep network based on original image and semantic segmentation input. This energy map represents each object instance as energy basins so that each instance can be extracted with a cut at a single energy level. This approach achieved more than double the performance of the state-of-the-art approaches at the time.

\subsubsection{Deep Metric Learning}
\begin{figure}[h]
\centering
\includegraphics[width=\textwidth]{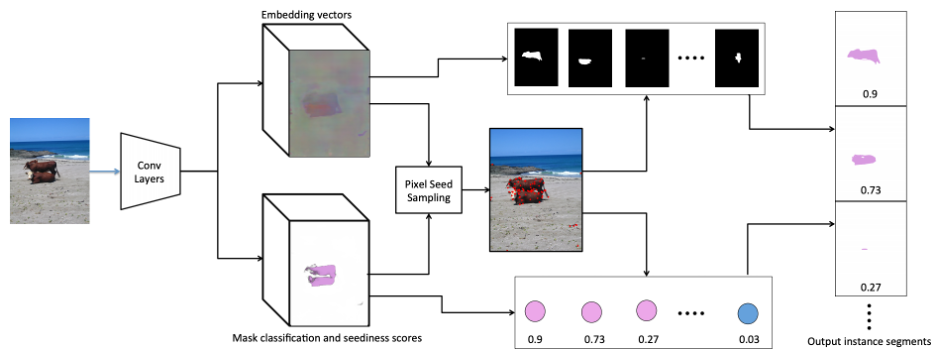}
\caption{The Deep Metric Learning framework. From \citet{fathi2017semantic}.}
\label{fig:deepmetric}
\end{figure}

\citet{fathi2017semantic} propose a clustering-based algorithm for instance segmentation. First embedding vectors for each pixel in the input image are generated with fully convolutional layers. These embeddings are used to compute similarity scores between each pair of pixels and are used to group pixels together. Specifically, seediness scores for pixels are calculated using a deep scoring model to select seed points, each of which generates a mask and its associated class label and confidence score based on the embedding vectors (Figure \ref{fig:deepmetric}).

\subsubsection{Soft Teacher and Box Jittering}
\citet{xu2021end} propose two techniques – soft teacher and box jittering in a semi-supervised end-to-end framework for object detection. The teacher model performs pseudo-labeling of the unlabeled images, whereas the student model is responsible for detection training, taking in as input both labeled and pseudo-labeled images. The exponential moving average (EMA) of the student model is used to update the teacher model during the end-to-end training process. Instead of assigning hard category labels to the box candidates generated by the student model, the “soft” mechanism weighs the classification loss with a “reliability” measure produced by the teacher model itself. The latter approach has significantly better performance than the former hard foreground/background assignment methods \citep{zoph2020rethinking,liu2021swin}. The box jittering approach is used to select reliable pseudo boxes to train the student model’s box regression branch. Candidate pseudo boxes are first jittered several times before regressed with the teacher model. A reliability score is calculated based on the variance of the regressed boxes, and candidates with high reliability scores are used in training the student’s regression branch. When integrated with the Swin Transformer based object detector \citep{liu2021swin}, these two techniques improve instance segmentation accuracy by 1.2 mAP on the COCO benchmark dataset \citep{lin2014microsoft}.

\subsubsection{SparseInst}
\citet{cheng2022sparse} design a novel object representation method which uses instance activation maps that highlight informative regions of objects. According to the highlighted regions, features are aggregated to obtain instance-level features \citep{cheng2022sparse}. Additionally, these instance activation maps predict objects in a one-to-one style and remove the need for non-max suppression (NMS) in post-processing \citep{cheng2022sparse}. This lightweight design allows SparseInst to achieve an extremely fast inference speed of 40 FPS on the COCO dataset while maintaining a decent segmentation accuracy of 37.9 AP \citep{cheng2022sparse}.

\section{Discussion}
The variety and complexity of the real-world scenes including image resolutions, varying object scales and sizes, occlusions and deformations, lighting conditions and background noise all pose different challenges for instance segmentation. It is especially important to address these challenges during instance segmentation as it requires precise separation and categorization of each individual object. Various architectures including feature pyramids and effective utilization of features from different layers of the network allow the models to solve some of these problems, but many still remain under-explored. 

R-CNN based approaches lay the foundation for two-stage methods that separate object localization and classification. These methods achieve good segmentation accuracy and are simple to implement, but are very memory and time-consuming and are not suitable for real-time applications. Semantic segmentation approaches provide the basis for one-stage instance segmentation models that do not require feature localization. They have better generalization capabilities, fast training and inference speed, but usually perform worse in terms of segmentation accuracy. Some of the most challenging issues addressed by both classes of algorithms include how to incorporate short- and long-range contextual information, how to deal with clutters and occlusions, and how to refine object boundaries to improve the quality of localization and categorization. Many algorithms focus on improving the quality of the segmentation output, whereas others attempt to reduce the computational intensity and focus on coarse but fast detection and segmentation. Despite these different focuses, each class of models follow a similar basic design of encoder architectures using convolution kernels while varying on the other components such as long-range skip connections and refinement modules. There remains to be a lot of potential for balancing the speed and quality of instance segmentation algorithms, but the efforts will likely continue to focus on addressing these challenges.

The main contribution, benchmark performances, and categorization of the list of methods for instance segmentation discussed in this section are summarized in Table \ref{tab:instance}.

\begin{center}
\begin{scriptsize}
\begin{longtable}{|L{1.2cm}|L{3.9cm}|L{3.6cm}|L{1.3cm}|} 
\caption{Summary of Popular Instance Segmentation Methods.} \label{tab:instance} \\

\hline \textbf{Name} &                                                                                                                                                           \textbf{Main Contribution} &                                                                                                                                                         \textbf{Benchmarks Performances} &                  \textbf{Categorization} \\
                               \hline
\endfirsthead

\caption* {\textbf{Table 4.1 Continued}}\\

\hline \textbf{Name} &                                                                                                                                                           \textbf{Main Contribution} &                                                                                                                                                         \textbf{Benchmarks Performances} &                  \textbf{Categorization} \\
                               \hline
\endhead

\hline \multicolumn{4}{|r|}{{Cont.}} \\ \hline
\endfoot

\hline \hline
\endlastfoot

                                                               SDS &                                                                                                        Multiscale Combinatorial Grouping (MCN) algorithm for region proposals &                                                                          PASCAL VOC (Mask AP=52.6) &      Two-stage \\ \hline
                      DeepMask &                                                                                                                                    Region proposals and object classification &                                            MS COCO (AR= 36.6), PASCAL VOC 2007 (AR = 43.3) &      Two-stage \\ \hline
                     SharpMask &                                                                                 Augmenting feedforward nets for object segmentation with a novel top-down refinement approach &                                                                         MS COCO (AR= 39.3) &      Two-stage \\ \hline
                  MultiPathNet & Capturing multi-scale contexts with foveal structure; skip connections for recovering spatial information; novel loss function integrating metrics at multiple IOU thresholds &                                                                     MS COCO (Mask AP=25.0) &      Two-stage \\ \hline
                    Mask R-CNN &                                       Extends Faster R-CNN by adding a branch for predicting an object mask in parallel with the existing branch for bounding box recognition & Cityscapes (Mask AP=36.4), MS COCO (ResNeXt-101: Mask AP=37.1, SpineNet-190: Mask AP=46.1) &      Two-stage \\ \hline
            Mask Scoring R-CNN &                                                                                          MaskIoU head trained with simple regression for improving segmentation mask quality &                                                                     MS COCO (Mask AP=39.6) &      Two-stage \\ \hline
                         PANet &                                                                                                 Bottom-up path, adaptive feature pooling, and fully-connected layers that improve information propagation in Mask-R-CNN &                                          MS COCO (Mask AP=42.0), Cityscapes (Mask AP=41.4) &      Two-stage \\ \hline
                          MNCs &                                                                        Three-stage network consisting of instance differentiation, mask estimation, and object categorization &                                 PASCAL VOC (Mask AP=63.5), MS COCO (Mask AP=24.6) &      Two-stage \\ \hline
                   MaskX R-CNN &                                                     Weight transfer function that maps a category’s bounding box detection parameters to its instance segmentation parameters &                                                                     MS COCO (Mask AP=29.5) &      Two-stage \\ \hline
                       MaskLab &                                                                                   Three branches consisting of box detection, semantic segmentation, and direction prediction &                                                                     MS COCO (Mask AP=43.0) &      Two-stage \\ \hline
                           OMN &                                                                                               Inference of object segment representation based on distance to object boundary &                             PASCAL VOC (MASK AP=65.69), Cityscapes (Mask AP=17.4) &      Two-stage \\ \hline
            Cascade Mask R-CNN &                                                                         Progressively improves boudning box quality with sequential detectors with increasing IoU thresholds &                                                         MS COCO (2017 val)  (Mask AP=42.3) &      Two-stage \\ \hline
                          FCOS &                                                                                          Center-ness loss measuring how “centered” is the pixel location in its bounding box  &                                                                     MS COCO (Mask AP=44.7) &      One-stage \\ \hline
                    ExtremeNet &                                                                                                                 Extreme points used to approximate an instance segmentation mask &                                                                     MS COCO (Mask AP=34.6) &      One-stage \\ \hline
                      ESE-Seg  &                                                         Using inner center radius shape signature to parameterize the contour and Chebyshev polynomials  to fit the signatures &                                                                     MS COCO (Mask AP=21.6) &      One-stage \\ \hline
                     PolarMask &                                                                                                      Formulating segmentation as  a regression problem and the Polar IoU Loss &                                                                     MS COCO (Mask AP=32.9) &      One-stage \\ \hline
                    FourierNet &                                                                                                 Obtaining contour points with Inverse Fast Fourier Transform (IFFT) algorithm &                                                                     MS COCO (Mask AP=30.6) &      One-stage \\ \hline
                        MEInst &                                                                                  Mask representation with compact fixed-dimension vector through dictionary learning methods  &                                                                     MS COCO (Mask AP=38.2) &      One-stage \\ \hline
                    TensorMask &                                                    Representation of image content (e.g., masks) with structured high-dimensional tensors in a set of densely sliding windows &                                                                     MS COCO (Mask AP=37.3) &      One-stage \\ \hline
                        YOLACT &                                                                                                 Real-time instance segmentation through linear combination of prototype masks &                                                                     MS COCO (Mask AP=31.2) &      One-stage \\ \hline
                        YOLACT++ &                                                                   Adds a fast mask re-scoring branch and deformable convolutions to YOLACT, and optimizes the prediction head &                                                                     MS COCO (Mask AP=34.1) &      One-stage \\ \hline
                       SipMask &                                       Accurate delineation of spatially adjacent instances and improved mask predictions with a lightweight spatial preservation (SP) module  &                                                                     MS COCO (Mask AP=32.8) &      One-stage \\ \hline
                    SparseInst &                                                                                                                          Object representation with instance activation maps  &                                                                     MS COCO (Mask AP=37.9) &      One-stage \\ \hline
                     BlendMask &                                                                              Blender module incorporating rich instance-level information with accurate dense pixel features &                                                                     MS COCO (Mask AP=41.3) &      One-stage \\ \hline
                    CenterMask &                 Predicting segmentation mask inside detected box with Spatial Attention-Guided Mask (SAG-Mask)  by focusing on the informative pixels while suppressing noise &                                                                     MS COCO (Mask AP=41.8) &      One-stage \\ \hline
                      CondInst &                                                                                                   Dynamically generated convolutional filters conditioned on target instances &                                                                     MS COCO (Mask AP=40.1) &      One-stage \\ \hline
                          SOLO &                                                                                                 Transforms instance segmentation into two pixel-level classification problems &                                                                     MS COCO (Mask AP=37.8) &      One-stage \\ \hline
                        SOLOv2 &                                                                                       Decomposing mask learning to feature learning and dynamic convolution kernel generation &                                                                     MS COCO (Mask AP=41.7) &      One-stage \\ \hline
                          SOTR &                                                 Twin attention mechanism with sparse representation; dynamic convolution kernel generation and segmentation with Transformers &                                                                     MS COCO (Mask AP=42.1) &      One-stage \\ \hline
                          DETR &                                                                                                   Set-based global loss that forces unique predictions via bipartite matching &                                                              MS COCO (Panoptic Quality=46) &    Query-based \\ \hline
                     QueryInst &                                                                                                                                    Parallel supervision on dynamic mask heads &              MS COCO (Mask AP=42.8), Cityscapes (Mask AP=34.4), YouTube-VIS (Mask AP=36.2) &    Query-based \\ \hline
          Deep Metric Learning &                                                                                                                         Clustering pixel embeddings for instance segmentation &                                                        PASCAL VOC (Mask AP=62.21) &        Other \\ \hline
      Deep Watershed Transform &                                                                                                                       Extraction of object instance segments with energy map &                                                                  Cityscapes (Mask AP=19.4) &          Other \\ \hline
Soft Teacher and Box Jittering &                                                                                                          Semi-supervised training with both labeled and pseudo-labeled images &                                                                     MS COCO (Mask AP=53.0) &          Other \\ \hline
\end{longtable}
\end{scriptsize}
\end{center}

\chapter{Deep Learning Models for 3D and Video Segmentation}
3D geometric data come in many forms of representations such as point clouds, meshes, voxel grids, depth maps, parametric models, and RGB-D. Segmentation of 3D data can be challenging, and different approaches exist for different representations.  Video Sequences can also be considered as a form of 3D data since time can be considered as a third dimension. As still-image segmentation techniques mature over the recent years, video segmentation has started to gain more attention. Naively applying still-image segmentation methods to video frames is computationally costly and tends to miss the inherent temporal continuity of videos. Segmentation of 3D/video data involves taking an input and assigning an output label to each minimum unit of the input. Formally, given an 3D/video input $\mathcal{P}={p\textsubscript{1},p\textsubscript{2}, ..., p\textsubscript{m}}$, where $p_i$ denotes a minimum unit (e.g. voxel, point, mesh, pixel within a video frame), a label space $\mathcal{L}={l\textsubscript{1},l\textsubscript{2}, ..., l\textsubscript{n}}$, it assigns an element from $\mathcal{L}$ to each element of $\mathcal{P}$. Note that the family of 3D/video segmentation methods can also be further categorized into semantic, instance, and panoptic segmentation, and thus the specific formulation for each subcategory of the problem will be slightly different from the general formulation above.

In addition to the practical applications of image segmentation methods, 3D segmentation is also utilized in robotics and augmented/virtual reality. Object segmentation from videos is a challenging task where objects are segmented while being tracked across frames. It is particularly important for applications such as person identification, tracking in security videos and object tracking for self-driving. This section reviews deep-learning-based approaches that have developed specifically for 3D data and video frame segmentation.

\section{Voxel-based Semantic Segmentation}
Voxels are volumetric grids that divide the 3D space, serving a similar function as pixels in the 2D space. High resolution voxels are necessary when representing detailed structure information, thus demanding increased computational power that often becomes the bottleneck for fine-grained object segmentation. The \textbf{3DCNN-DQN-RNN} \citep{liu20173dcnn} is designed to address this challenge. It combines 3D convolutional neural network (3DCNN), Deep Q-Network (DQN), and Residual recurrent neural network (RNN). The 3DCNN network first learns visual, spatial, and contextual features from the point cloud data at multiple scales and produces a 3DCNN feature representation. The DQN controls an eye window which learns through trial-and-error to localize class objects: the eye window attempts to envelop the points of a class object accurately and the 3DCNN evaluates a reward based on the points in the eye window, allowing important features to be captured efficiently. Colors and coordinates of the points in the eye window are then combined with the 3DCNN feature representation into one input vector and fed into the RNN, which produces the final class labels for the input points. This method achieves state-of-the-art performance on the Stanford 3D semantic parsing data set (S3DIS) \citep{armeni20163d} and the SUNCG \citep{song2017semantic} datasets.

\begin{figure}[h]
\centering
\includegraphics[width=0.90\textwidth]{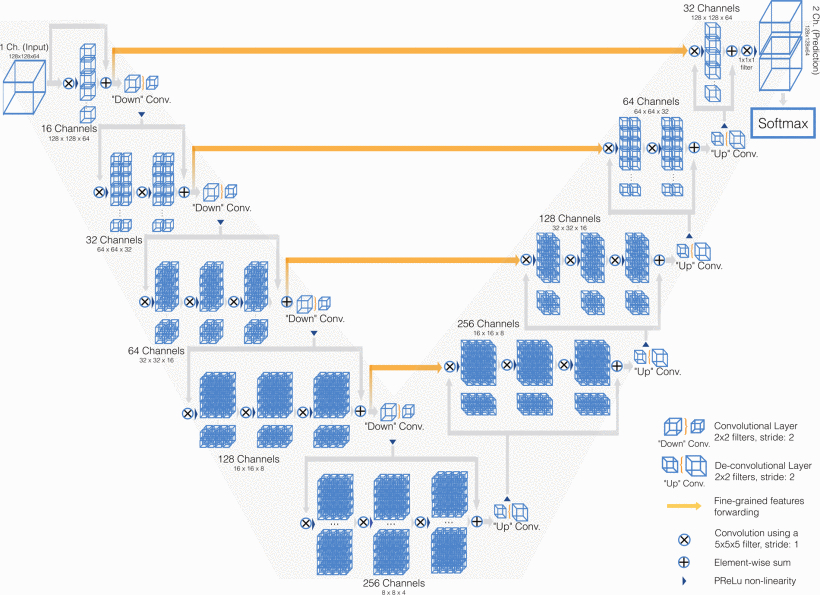}
\caption{The V-Net architecture. From \citet{milletari2016v}.}
\label{fig:vnet}
\end{figure}

\textbf{V-Net} \citep{milletari2016v} is based on fully convolutional neural networks and developed for 3D image segmentation for segmentation of prostate MRI volumes. This task is challenging because prostate’s appearance varies significantly in different scans while artifacts and distortions also affect MRI volumes. Unlike many other approaches which process 3D input in a slice-wise fashion, V-Net uses volumetric convolutions instead. The network architecture includes a compression and a decompression path. The compression path performs volumetric convolutions in stages to reduce resolution of the data and replaces pooling with convolution layers that serve the same functions. The decompression path up-samples the compressed feature map and generates a final two-channel volumetric segmentation. Similar to U-Net \citep{ronneberger2015u}, features extracted from the left part of the network are passed to the right part to add fine-grained detail to improve final contour predictions (Figure \ref{fig:vnet}). A novel loss function based on the Dice coefficient is also introduced to reduce the effect of background-foreground voxel imbalance. V-Net achieves competitive performance on the PROMISE2012 \citep{litjens2014evaluation} challenge dataset.

A different line of approaches to reduce the computational complexity associated with large-scale regular voxel representations is through non-uniform voxel representation of the 3D space. These approaches are based on the intuition that volumetric representations in the 3D space are naturally sparse and therefore should not require the application of dense convolutions. \textbf{OctNet} \citep{riegler2017octnet} utilizes unbalanced octrees to divide the 3D space hierarchically and non-uniformly to allow efficient allocation of memory to the important voxels. The submanifold sparse convolution (\textbf{SSC}) developed by \citet{graham20183d} explore this idea further by completely eliminating computations in the empty spaces, allowing spatially-sparse data to be processed even more efficiently.

\section{Point Cloud-based Semantic Segmentation}

Point cloud is a sparse representation method for 3D scenes. It is composed of unordered and unevenly sampled 3D data points captured from sensors such as Lidars, and suffers from high redundancy and the lack of explicit structure \citep{6758588}. As a result, it is not possible to directly apply CNNs to this type of 3D data, hence either transformation to other data formats or modified network structures taking into account of possible data permutations are required to process point clouds. A key intuition is the distances between points can be a useful signal for determining objects or object clusters due to its variant nature, thus segmentation models could leverage distance information to localize objects in the 3D space.

\enlargethispage{\baselineskip}
\textbf{3D-CNN} was initially applied to video data analysis as the time dimension can be considered as the third dimension in addition to 2D image frames. \citet{huang2016point} first introduce 3D-CNN for 3D point cloud segmentation. During training, point cloud input is first transformed into 3D voxels. Random keypoints balanced across different categories are generated, and the occupancy voxel grids centered around each keypoint are labeled based on the dominating category in the cell around the keypoint \citep{huang2016point}. The 3D-CNN, composed of two 3D convolutional layers, two 3D pooling layers, and a fully connected layer followed by logistic regression, is trained using the voxels and their labels. During inference, center points of voxel grids are densely sampled at a distance. All points in a cell are labeled based on the label of the local voxel box the cell is close to.

\begin{figure}[h]
\centering
\includegraphics[width=\textwidth]{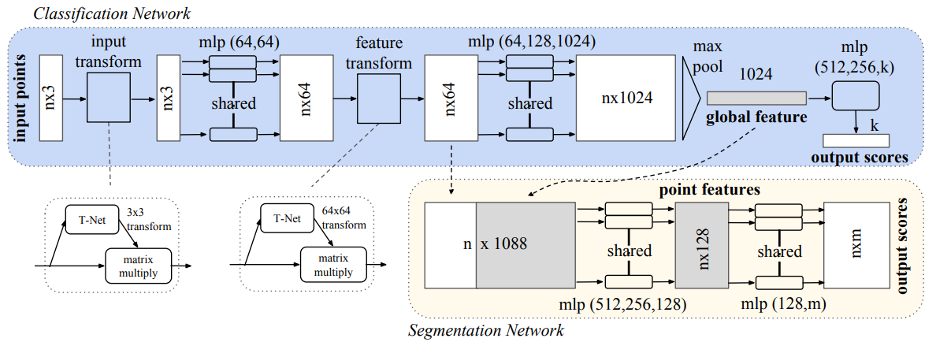}
\caption{The PointNet framework. From \citet{qi2017pointnet}.}
\label{fig:pointnet}
\end{figure}

Point clouds is an irregular format for representing 3D data, and most existing methods transform point clouds to regular 3D voxel grids or collections of images before applying a deep net architecture \citep{qi2017pointnet}. \textbf{PointNet} \citep{qi2017pointnet} is developed to directly consume raw point clouds and generate segmentation labels for each input point. The network first applies input and feature transformations on the points. Next, point features are aggregated through max pooling. A multi-layer perceptron (MLP) is used to generate the final results. For object classification, it outputs classification scores for each candidate class. For semantic segmentation, an extended structure concatenates global and local features before applying the mlp to generate per point scores for each input point and each semantic category (Figure \ref{fig:pointnet}). PointNet achieves the competitive performance on the ShapeNet part \citep{yi2016scalable} dataset. 

The Recurrent Slice Network (\textbf{RSNet}) \citep{huang2018recurrent} is designed to segment raw 3D point cloud data. The key component of RSNet, the local dependency module, is composed of a slice pooling layer, followed by RNN layers, and finally a slice unpooling layer. The sliced pooling layer maps features extracted from unordered points to ordered features. RNN layers are then applied to these aggregated features to extract feature dependencies. Finally, the slice unpooling layer performs a reverse mapping to assign features back to the original points. RSNet achieves competitive performances on the S3DIS \citep{armeni20163d},
ScanNet \citep{dai2017scannet}, and ShapeNet \citep{chang2015shapenet} datasets. In addition, its simple architecture exhibits high computional efficiency as well.

\citet{landrieu2018large} develop Superpoint Graphs (\textbf{SPGs}) as a novel framework for representing contextual relationships between object parts in 3D point clouds. This representation framework treats objects parts as a whole during classification, captures detailed relationship between adjacent objects, and scales with the number of simple object structures rather than the number of points, making it both effective and memory-efficient for segmentation of large-scale 3D point cloud data. When combined with a graph convolutional network based on Edge-Conditioned Convolutions \citep{simonovsky2017dynamic} for contextual segmentation, the SPG framework sets competitive performances on Semantic3D \citep{hackel2017semantic3d} and S3DIS \citep{armeni20163d} datasets.

Using random sampling and effective local feature aggregation, \textbf{RandLA-Net} \citep{hu2020randla} is designed for efficiently segmenting large-scale point clouds. The key approach to improving computational efficiency is through random sampling to reduce point density. However, this downsampling process may discard important features, thus a novel local feature aggregation module is designed to address this challenge. Specifically, this module extracts features from input points using a location spatial encoding (LocSE) block, and weights the important neighboring features with the attentive pooling mechanism based on local context and geometry \citep{hu2020randla}. These two components are chained together within a residual block to increase the receptive field size for each randomly sampled point, so that it is more likely that the geometric details of the input point clouds are reserved. RandLA-Net achieves competitive performance on both Semantic3D \citep{hackel2017semantic3d} and SemanticKITTI \citep{behley2019semantickitti} datasets.

\citet{fan2021scf} design a novel module that learns Spatial Contextual Feature (\textbf{SCF}) from large-scale point clouds. The SCF module can be incorporated into various network architectures for 3D semantic segmentation. It consists of three main components: 1) the Local Polar Representation (LPR) that learns for each 3D point a z-axis rotation invariant representation; 2) the Dual-Distance Attentive Pooling (DDAP) block that learns effective local features automatically based on the neighboring points’ geometric and feature distances; and 3) the Global Contextual Feature (GCF) block that learns the global context of each point based on its spatial location and its neighborhood’s volume ratio to the global point cloud \citep{fan2021scf}. SCF-Net achieves competitive performances on S3DIS \citep{armeni20163d} and Semantic3D \citep{hackel2017semantic3d} datasets.

\citet{qiu2021semantic} introduce Bilateral Augmentation and Adaptive Fusion (\textbf{BAAF-Net}) for segmenting 3D point clouds. BAAF-Net addresses the challenging raw nature of 3D data through two key components: a Bilateral Context Block that augments local context of points with geometric and semantic features and the Adaptive Fusion Module that up-samples the bilateral blocks’ outputs and adaptively fuses these multi-resolutional features to produce an output feature map representing comprehensive point-wise features. BAAF-Net achieves state-of-the-art performance on the S3DIS dataset \citep{armeni20163d}.

The \textbf{Point Transformer} layer \citep{zhao2021point} is a permutation- and cardinality-invariant layer designed for processing 3D point clouds. The Point Transformer network is constructed based on the point transformer layer and can serve as the backbone for many tasks including semantic segmentation and object classification. It applies vector self-attention to each point’s local neighborhood and to the encoding of the network’s positional information, enabling the exchange of information and ensuring high expressivity with adaptive modulation of individual feature channels.

\citet{zhu2021cylindrical} propose a novel framework named \textbf{Cylinder3D} for outdoor LiDAR segmentation to address outdoor point clouds’ sparsity and varying density issues. Specifically, this framework incorporates Cylindrical Partition, which uses a cylinder coordinate where point clouds are divided dynamically based on distance from the origin, producing a more balanced point distribution against varying density. Asymmetrical 3D convolution networks that power the horizontal and vertical weights are then applied to generate the voxel-wise outputs that match the point distribution of the scene while enhancing robustness to sparsity. Finally, outputs are refined via a point-wise module that alleviates the interference of lossy voxel-based label encoding \citep{zhu2021cylindrical}. This framework achieved state-of-the-art performance on the SemanticKITTI \citep{behley2019semantickitti} and the nuScenes \citep{caesar2020nuscenes} datasets.

\citet{peng2022mass} propose a multi attentional semantic segmentation model with LiDAR point cloud data as input. Their proposed framework comprises of two parts: a) PillarSegNet model for generating dense top-view segmentation maps using pillar-level features; b) Multi-attention framework for aggregating local and global features to generate attention maps that are useful for the segmented output which is a dense 360$^{\circ}$
segmentation mask. Experiments show that the proposed model outperforms the state of the art by 19.0\% on
SemanticKITTI \citep{behley2019semantickitti} and reaches 30.4\% in mIoU on nuScenesLidarSeg \citep{caesar2020nuscenes} datasets. 

\section{RGB-D-based Semantic Segmentation}
Use of low-cost RGB-D sensors allows depth information to be extracted from scenes, providing  geometric cues that help improve the accuracy of semantic segmentation. RGB-D data are often considered 2.5D data due to their shared components with 2D images. Previously introduced \textbf{LSD-Net} \citep{cheng2017locality} and \textbf{LSTM-CF} \citep{li2016lstm} are tailored for segmenting RGB-D data. \citet{zeng2017multi} and \citet{ma2017multi} leverage the multiple views obtained from a moving RGB-D camera for improving 3D segmentation. The Data Associated Recurrent Neural Networks (\textbf{DA-RNN}) \citep{xiang2017rnn} utilizes recurrent neural network structures for 3D semantic segmentation of RGB-D videos. It uses a Data Associated Recurrent Unit (DA-RU) that takes as input a pixel, accumulates its information through time by incorporating its DA-RU’s hidden state from the previous frame, and injects semantic labels into the 3D scene reconstructed with KineticFusion \citep{henry2012rgb}, which also provides the necessary data associations to help connect the recurrent units between RGB-D frames \citep{xiang2017rnn}. This work also generates a new synthetic dataset based on the ShapeNet \citep{chang2015shapenet} repository for future 3D semantic mapping research. \citet{sun2020real} propose RFNet, a real-time RGB-D semantic segmentation framework trained using multiple datasets. It achieves high accuracy in detecting unexpected small obstacles through effective usage of depth complementary features, making it very useful in autonomous vehicles applications. \citet{chen2020bi} introduces \textbf{SA-Gate} to effectively address the uncertainty in depth measurements as well as the variations between RGB and Depth modalities, allowing for efficient cross-modality feature aggregation \citep{chen2020bi}. \citet{liu2022cmx} extends the utility of multi-sensor data by designing a Cross-Modal Feature Rectification Module (CM-FRM) that calibrates bi-modal features through their spatial- and channel-wise correlations. They name this cross-modal fusion framework \textbf{CMX} for its application in RGB-X semantic segmentation \citep{liu2022cmx}. \citet{cao2021shapeconv} propose a model-agnostic Shape-aware Convolutional layer (\textbf{ShapeConv}) that decomposes depth feature into a shape-component and a base-component before effectively re-merging them through a weighted combination. They argue that the shape component of the depth feature has a stronger connection to semantics and thus is more critical for segmentation accuracy \citep{cao2021shapeconv}. Other works that exploit multi-modal feature fusion include \textbf{RDFNet} \citep{park2017rdfnet}, \textbf{ACNet} \citep{hu2019acnet}, \textbf{NANet} \citep{zhang2021non}, and \textbf{(SGNet)} \citep{chen2021spatial}.

In addition to semantic segmentation, multi-task learning have been successful using RGB-D data. \citet{chen2020bi} propose a multi-task guided prediction-and-distillation network (\textbf{PAD-Net}) that achieves simultaneous depth estimation and scene parsing. A set of intermediate auxiliary tasks are utilized as multi-modal input to the proposed multi-modal distillation modules to accomplish the final tasks. The Pattern-affinitive Propagation (\textbf{PAP}) \citep{zhang2019pattern} method aims to achieve three joint prediction tasks of depth estimation, surface normal prediction and semantic segmentation using RGB-D input. It effectively models the affinitive patterns among tasks in pair-wise similarities to boost and regularize joint-task learning.

\section{Other Models for 3D Data Segmentation }
The advance of 3D scanners such as LiDAR continues to contribute to the increasing abundance of 3D data and the development of new methods for 3D data segmentation. Other approaches for 3D data segmentation include \citet{wang2019dynamic}, \citet{thomas2019kpconv}, \citet{engelmann2020dilated}, and \citet{liu2019point2sequence} where point cloud data are directly processed, and \citet{liu2019relation}, \citet{yan2020pointasnl}, \citet{qiu2021geometric}, and \citet{qiu2021dense}, where semantic knowledge is learned through feature representations. The main contribution, benchmark performances, and categorization of the popular 3D segmentation methods are summarized in Table \ref{tab:threed}.

\begin{scriptsize}
\begin{longtable}{|L{1cm}|L{4.2cm}|L{3.8cm}|L{1cm}|} 
\caption{Summary of Popular 3D Segmentation Methods.} \label{tab:threed} \\
\hline \textbf{Name} &                                                                                                                                                           \textbf{Main Contribution} &                                                                                                                                                         \textbf{Benchmarks Performances (default metric: mIoU (\%))} &                  \textbf{Categorization} \\
                               \hline
\endfirsthead

\caption* {\textbf{Table 5.1 Continued}}\\

\hline \textbf{Name} &                                                                                                                                                           \textbf{Main Contribution} &                                                                                                                                                         \textbf{Benchmarks Performances} &                  \textbf{Categorization} \\
                               \hline
\endhead

\hline \multicolumn{4}{|r|}{{Cont.}} \\ \hline
\endfoot

\hline \hline
\endlastfoot

                                                               3DCNN-DQN-RNN &                                                                                  Combining CNN, RNN, and Deep Q-Network for 3D object recognition &            Stanford 3D semantic parsing (70.76), SUNCG (33.21) &       Voxel based \\ \hline
            V-Net &                                                        Loss function based on Dice coefficient for tackling background-foreground voxel imbalance &                                         PROMISE2012 (Dice score=82.39) &       Voxel based \\ \hline
           OctNet &                  Hierarchical and nonuniform division of 3D space using unbalanced octrees for efficient allocation of memory to important voxels &                                                RueMonge2014 (81.5) &       Voxel based \\ \hline
              SSC &                                                             Eliminating computations in empty spaces to efficiently process spatially-sparse data &                                              NYU Depth test (68.5) &       Voxel based \\ \hline
           3D-CNN &                                                                                     3D point cloud data segmentation through voxel transformation & Large Lidar point cloud dataset of the urban area of Ottawa (93.0) & Point cloud based \\ \hline
         PointNet &                                                              Raw point clouds segmentation with feature transformation and multi-layer-perceptron &                                                   ShapeNet part (83.7) & Point cloud based \\ \hline
            RSNet &                                                                              Extraction of feature dependencies with slice pooling and RNN layers &                   S3DIS (51.93),  ScanNet (39.35), ShapeNet (84.9) & Point cloud based \\ \hline
              SPG &                                                       Novel 3D point cloud representation capturing contextual relationships between object parts &                                       S3DIS (62.1),  Semantic3D (76.2) & Point cloud based \\ \hline
       RandLA-Net &                                                                                     Random sampling of point clouds and local feature aggregation &                                Semantic3D (77.4), SemanticKITTI (53.9) & Point cloud based \\ \hline
          SCF-Net &                                                                                 Spatial Contextual Feature learning from large-scale point clouds &                                       S3DIS (71.6),  Semantic3D (77.6) & Point cloud based \\ \hline
         BAAF-Net &                                                 Local context augmentation with bilateral block and adaptive fusion of multi-resolutional feature &                 S3DIS (72.2),  Semantic3D (76.2), SemanticKITTI (59.9) & Point cloud based \\ \hline
Point Transformer &                                                                                                     Vector self-attention based transformer layer &                            S3DIS (73.5), ShapeNetPart (ins. mIoU=86.6) & Point cloud based \\ \hline
       Cylinder3D &                                                                             Cylindrical partition of point clouds and asymmetrical 3D convolution &                                 SemanticKITTI (67.8), nuScenes (76.1)  & Point cloud based \\ \hline
          LSD-Net &                                                                                                              Boundary refinement and RGB-D fusion &                                                   NYU-Depth v2 (45.9) &       RGB-D based \\ \hline
          LSTM-CF & Capturing and fusion of contextual information from multiple channels of photometric and depth data with Long Short-Term Memorized Context Fusion &                                         SUN-RGBD (48.1), NYUDv2 (49.7) &       RGB-D based \\ \hline
           DA-RNN &                                                                   Injection of semantic labels into 3D scenes with Data Associated Recurrent Unit &                            Synthetic ShapeNet Scene dataset (85.9) &       RGB-D based \\ \hline
\end{longtable}
\end{scriptsize}

\section{Video Object Segmentation}
\enlargethispage{\baselineskip}
Early deep-learning-based video object segmentation techniques \citep{fragkiadaki2015learning,tsai2016video,tokmakov2017learning,li2017primary} rely on multi-stage methods where visual features extracted with CNNs are passed into a classification network to generate object segmentation masks. Later works take an object embedding-centric approach (\citeauthor{li2018instance}, \citeyear*{li2018instance}; \citeyear*{li2018unsupervised}). More recent works have started to design more unified architectures that are end-to-end trainable \citep{tokmakov2017learning,jain2017fusionseg,cheng2017segflow}. These works leverage a variety of convolutional \citep{li2019motion,zhou2020motion,lu2020zero,zhang2020unsupervised,ren2021reciprocal}, recurrent \citep{zhou2020motion,song2018pyramid}, and Graph Neural Network-based \citep{wang2019zero} structures for modeling the contextual information and spatial-temporal relationships among objects across frames for better segmentation. 
\textbf{SSTVOS} is a Transformer-based video segmentation framework developed by \citet{duke2021sstvos} that uses self-attention that learns to search for similar and reference masks in the temporal history in order to segment a frame. \citet{khoreva2018video} augment the DAVIS 2016 \citep{perazzi2016benchmark} and DAVIS 2017 \citep{pont20172017} with textual descriptions of objects for segmentation with natural language referring expressions.

\citet{li2018unsupervised} propose a motion-based bilateral network (\textbf{MBNM}) that identifies regions with motion patterns similar to those of non-object regions to associate static objects with the background. Consequently, MBNM helps reduce the false positives from static but semantically similar objects for the Video Object Segmentation problem \citep{li2018unsupervised}. \textbf{DFNet} is a novel unsupervised object segmentation framework developed by \citet{zhen2020learning} that learns discriminative features (D-features) that reveal feature distribution from a global view. An attention module under a CRF formulation is then used to extract correlations between the input images for inference of common foreground objects. While the majority of existing work leverage 2D convolutional networks for per-frame segmentation of videos, \citet{mahadevan2020making} design a simple encoder-decoder framework (\textbf{3DC-Seg}) composed entirely of 3D convolutions for video object segmentation. This model can be trained from end-to-end with the standard cross-entropy loss and achieves state-of-the-art performance on the DAVIS 2016 \citep{perazzi2016benchmark} dataset.

\section{Video Semantic Segmentation}
Semantic segmentation of videos can be achieved naively by applying 2D semantic segmentation approaches for each frame of the video. However, this approach neglects the continuity along the temporal axis and the dependencies among objects across different video frames. \citet{hur2016joint}, \citet{kundu2016feature}, \citet{jin2017video}, \citet{gadde2017semantic}, \citet{huang2018efficient}, and \citet{nilsson2018semantic} are some approaches developed for video semantic segmentation that focus on capturing temporal relationships among different frames to achieve better segmentation accuracy. \citet{chandra2018deep} integrate CNN-based per-frame feature extraction and CRF-based spatio-temporal reasoning into a unified framework (\textbf{VideoGCRF}) for end-to-end learning.

On the other hand, \citet{zhu2017deep}, \citet{li2018low}, \citet{xu2018dynamic}, and \citet{mahasseni2017budget} are approaches that focus on increasing the efficiency of video semantic segmentation by mining the temporal cues. Another example is \textbf{Clockwork Convnets} \citep{shelhamer2016clockwork}, which is developed based on the observation that semantic contents of scenes in consecutive frames evolve slower than actual pixel changes. In this framework, fixed or adaptive clock signals are used to yield purpose-fit computation schedules that determine when the FCNs process different layers based on their semantic stability and at what update rates. Fixed rate schedules reduce overall computation and limit the execution rate of deep layers due to slower semantics updates. Adaptive rate schedules are determined based on input and network states and vary with dynamics of the scenes.

A more recent work, \textbf{TMANet} \citep{wang2021temporal}, is the first framework applying memory and self-attention to build the temporal relation in videos. \textbf{ETC-MobileNet} incorporates a motion guided \textit{temporal loss} that enforces the network to assign consistent labels for a pixel across time frames \citep{liu2020efficient}. Through this approach, temporal consistency is explicitly encoded into the segmentation network as a constraint during the training process. The trained network is employed for per-frame inference with no computation overhead. In contrast with 2D-based approaches, \textbf{Voxel2Voxel} \citep{tran2016deep} is one of the first works leveraging a 3D Convolutional Neural Network architecture with 3D deconvolutional layers developed for video voxel prediction. Lower part of the V2V architecture contains layers from C3D \citep{tran2015learning} which has proven to be effective for video recognition. These layers can be either fine-tuned based on pretrained weights from C3D or learned from scratch. Upper part of the architecture includes additional 3D convolutional layers and 3D deconvolutional layers. The 3D deconvolutional layers deconvolve the filters spatio-temporally, as compared to the 2D variant where filters are deconvolved only spatially. This approach achieves competitive performance on three different tasks – video semantic segmentation, optical flow estimation, and video coloring.

Table \ref{tab:video} summarizes the main contribution, benchmark performances, and categorization of the popular video semantic and instance segmentation methods.

\begin{scriptsize}
\begin{longtable}{|L{1.5cm}|L{3.7cm}|L{3.7cm}|L{1cm}|} 
\caption{Summary of Popular Video Segmentation Methods.} \label{tab:video} \\

\hline \textbf{Name} &                                                                                                                                                           \textbf{Main Contribution} &                                                                                                                                                         \textbf{Benchmarks Performances} &                  \textbf{Categorization} \\
                               \hline
\endfirsthead

\multicolumn{4}{c}%
{{\bfseries \tablename\ \thetable{} Continued}} \\
\hline \textbf{Name} &                                                                                                                                                           \textbf{Main Contribution} &                                                                                                                                                         \textbf{Benchmarks Performances} &                  \textbf{Categorization} \\
                               \hline
\endhead

\hline \multicolumn{4}{|r|}{{Cont.}} \\ \hline
\endfoot

\hline \hline
\endlastfoot

             SSTVOS &                                                                Uses self-attention to learn to search for similar regions and reference masks in the temporal history of video frames & DAVIS17 val (Jaccard Mean: 75.4), Youtube-VOS 2018 val (Jaccard Seen: 80.9; Jaccard Unseen: 76.6) &   Video Object Segmentation \\ \hline
              MBNM &                                                Identifies regions with motion patterns similar to those of non-object regions to associate static objects with the background &                                                                      DAVIS16 val (Avg MAE: 0.031) &   Video Object Segmentation \\ \hline
             DFNet &                                                                        Capture inherent correlation among video frames through feature distribution from a global perspective &                                                                  DAVIS16 val (Jaccard Mean: 83.4) &   Video Object Segmentation \\ \hline
           3DC-Seg &                                                                                                                      End-to-end 3D convolutions for video object segmentation &                                                                  DAVIS16 val (Jaccard Mean: 84.3) &   Video Object Segmentation \\ \hline
         VideoGCRF &                                                                                                CNN-based per-frame feature extraction and CRF-based spatio-temporal reasoning &                                                                               CamVId (mIoU: 75.2) & Video Semantic Segmentation \\ \hline
Clockwork Convnets & Clockwork convnets driven by fixed or adaptive clock signals that schedule the processing of different layers at different update rates according to their semantic stability &                      Youtube-Objects (mIoU: 70.0), NYUD (mIoU: 28.9), and Cityscapes (mIoU: 64.4) & Video Semantic Segmentation \\ \hline
            TMANet &                                                                                   First framework applying memory and self-attention to build the temporal relation in videos &                                                  CamVId (mIoU: 76.5), Cityscapes val (mIoU: 80.3) & Video Semantic Segmentation \\ \hline
     ETC-MobileNet &                                                              Motion guided temporal loss that enforces the network to assign consistent labels for a pixel across time frames &                                                                               CamVId (mIoU: 76.3) & Video Semantic Segmentation \\ \hline
       Voxel2Voxel &                                                                                                 End-to-end trainable 3D convolutional architecture for voxel-level prediction &                                                                                     GATECH (accuracy: 76.0) & Video Semantic Segmentation \\ \hline
\end{longtable}
\end{scriptsize}

\section{Discussion}
Despite the multitude of challenges presented by the volume and complexity of 3D and video data, recent developments in virtual and augmented reality and related applications such as self-driving and robotics have continued to attract research interests in the fields of 3D and video segmentation. 

One of the major topics that emerge from these deep learning-based approaches when dealing with 3D data is the processing of the various forms of representations. Direct processing of irregular data formats such as point clouds require careful design of network architectures, whereas transformation-based approaches usually take up more computational resources. Another major focus is selecting important features to focus on to reduce computational load and increase model performance. Unevenly distributed 3D data present the opportunity for unequal weight assignment to different spatial locations. 

For video segmentation, it is of vital importance to capture both spatial and temporal dependency between objects as continuity is key among the frames in the videos. Effective architectures have been designed to capture these connections thanks to the continuous advance of deep learning techniques. Another major challenge in video segmentation is the availability of high-quality datasets and annotations. Current methods have relied heavily on older datasets such as DAVIS 2016 \citep{perazzi2016benchmark} and 2017 \citep{pont20172017}. More recent and complex datasets with increased variety need to be built in order to continue to improve the performance of models so that they can be readily deployed in real-world situations. Fortunately, these datasets should become increasingly available as self-driving becomes more widespread.

For tasks as complex as 3D and video segmentation, reproducibility is especially challenging as codebases require difficult setup and experiments are hard to replicate across different platforms. Better documentations need to be produced in order to facilitate the systematic and continuous development of the different approaches, in particular for the reduction in memory consumption and improvement in training and inference efficiency. These elements are essential and can often outweigh the importance of segmentation accuracy depending on the particular application. Finally, because these experiments are expensive to carry out, automatic exploration of optimized network architectures could become the driving force for design of future models.

\chapter{Deep Learning Models for Panoptic Segmentation}\label{chap6}
\begin{figure}[h]
\centering
\includegraphics[width=0.9\textwidth]{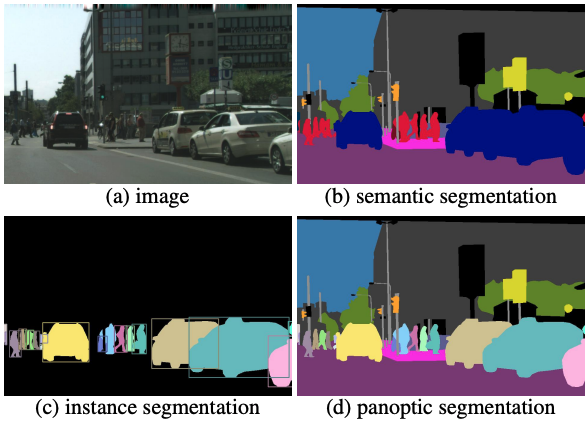}
\caption{Given an input image (a), ground truth label maps for (b) semantic segmentation, (c) instance segmentation, and (d) panoptic segmentation. From \citet{kirillov2019panoptic}.}
\label{fig:segtypes}
\end{figure}

Even though semantic segmentation and instance segmentation have their use cases in the vision community, both methods have some inherent limitations. Recall that semantic segmentation assigns a class label to every pixel in the image, and instance segmentation segments out each object instance separately. Thus, semantic segmentation is not able to separate different objects within the same class and instance segmentation is not able to detect ``stuff'' or background regions (that are not objects). To overcome these limitations, a unified segmentation approach called panoptic segmentation is proposed by \citet{kirillov2019panoptic}. Panoptic segmentation aims to generate a coherent and unified scene representation for visual understanding. The word panoptic means ``including everything visible in one view'' \citep{kirillov2019panoptic}. The task is to assign a semantic label as well as an instance ID to each pixel in the image, thus resulting in a unified segmentation. Panoptic segmentation can be formulated as follows: given an input image $\mathcal{P}={p\textsubscript{1},p\textsubscript{2}, ..., p\textsubscript{m}}$, a semantic label space $\mathcal{S}={s\textsubscript{1},s\textsubscript{2}, ..., s\textsubscript{n}}$, and $n$ instance ID spaces $\mathcal{D}_i={d\textsubscript{i1},d\textsubscript{i2}, ..., d\textsubscript{il}}$ for $i\in{1...n}$, it assigns an element $s_i$ from $\mathcal{S}$ and an element $d_{il}$ from $\mathcal{D}_i$ to each element of $\mathcal{P}$. One small difference between panoptic and instance segmentation when separating instances is that instance segmentation allows overlapping of objects while panoptic segmentation does not, since it assigns a unique instance ID to each pixel in the input image. Given an input image, Figure \ref{fig:segtypes} illustrates the ground truths label maps for semantic, instance, and panoptic segmentation. 

\enlargethispage{\baselineskip}
Panoptic segmentation is particularly useful when separation of ``background'' and ``foreground'' and of each foreground object instance are desired. Several practical applications of panoptic segmentation include: 1) cancer cell detection in medical images, where identification of each individual instance of tumors and locating them within amorphous regions are crucial for understanding the context; 2) image processing, where functions such as auto-focusing and background blurring can be achieved only after separating the foreground from the background and identifying the ``subject'' instance; and 3) autonomous driving, where distance between each object (car, person, tree) and between objects and the background (road, wall) need to be properly calculated. This section reviews the methods that solve panoptic segmentation.

\section{Top-down Methods (Two-stage)}
These methods perform detection first and then segment out the objects. The two-stage approach within this class of methods involve generating region proposals as the first step (or stage) followed by refining the proposals to generate the final segmentation mask. \citet{kirillov2019panoptic}, who propose the task of panoptic segmentation and achieve it by performing instance segmentation, semantic segmentation and adding processing steps on top of the segmentation outputs to unify the results. They formulate the task as assigning each pixel in the image, a label as well as an instance ID. They also propose a new metric for evaluating panoptic segmentation task: panoptic quality (PQ) metric. \citet{de2018panoptic} design the Joint Semantic and Instance Segmentation Network (\textbf{JSIS-Net}) panoptic segmentation using a single network (instead of two networks and then combine the results as was proposed by \citet{kirillov2019panoptic}. This end-to-end panoptic segmentation model is more memory efficient and takes less time than the original model. However, the semantic segmentation network architecture differs greatly from the model architecture that performs instance segmentation.

\enlargethispage{-\baselineskip}
To overcome the challenge of designing a unified panoptic segmentation model at the architectural level, \citet{kirillov2019panoptic2} propose a single network that generates instances and segments simultaneously. They use a Feature Pyramid Network (FPN) \citep{lin2017feature} backbone for extracting multiscale features and add a region-based branch on top of the FPN backbone. In addition, they incorporate Mask-RCNN \citep{he2017mask} as another branch stemming from the same FPN multiscale features for instance segmentation. They call this method \textbf{Panoptic FPN}. Further architectural improvements are made by \textbf{EfficientPS} \citep{mohan2021efficientps}. They propose a bidirectional FPN as the feature encoder and add the instance and segmentation branches on top of it. The outputs from these are fused and the whole network is optimized end-to-end. \textbf{Auto-Panoptic} \citep{wu2020auto} adds architectural search to the pipeline following the one-shot Network Architecture Search (NAS) paradigm which automatically searches for the best backbone, segmentation/instance branches and fusion module. 

Several methods have been proposed to fuse the outputs from the instance and semantic segmentation branches to produce the final result efficiently.  Occlusion Aware Network (\textbf{OANet}) \citet{liu2019end} propose a spatial ranking module which addresses occlusions among the predicted instances. \textbf{OCFusion} \citep{lazarow2020learning} propose to model the overlap between two instances as a binary relation problem and train it with ground truth from dataset annotations. \textbf{SOGNet} \citep{yang2020sognet} model the occlusions via a scene overlapping graph where objects are points and relations between them are edges. By explicitly  modeling the overlaps and resolving the occlusions, their approach does not require object relations supervision. 

\citet{wu2020bidirectional} model the correlations between objects and the background, something the prior methods have not explored. They propose a Bidirectional Graph Reasoning Network (\textbf{BGRNet}) which incorporates graph CNNs into the panoptic segmentation architecture to yield superior results. \citet{Xiong_2019_CVPR} propose a parameter-free panoptic head to merge the predictions from the semantic and instance segmentation networks (\textbf{UPSNet}). \citet{li2020unifying} propose an end-to-end panoptic segmentation model without requiring post-processing. Their panoptic head consists of a dynamic potential head representing panoptic instances and a dense affinity head that predicts the probability of pairs of pixels to belong to the same class (thing or stuff). 

Attention mechanisms are also incorporated within panoptic segmentation models with \textbf{AUNet} \citep{li2019attention}, a unified architecture with a foreground and background segmentation branches and a third RPN branch. Attention modules named Proposal Attention Module (PAM) and Mask Attention Module (MAM) are added respectively to the foreground and background branches. 

Some approaches propose new loss functions to the panoptic segmentation task. \citet{porzi2021improving} propose a bounding box regression loss that addresses the problem caused by using crops during training (e.g truncating large objects). \citet{zhang2020ada} propose \textbf{Ada-Segment} to adjust multiple training losses automatically during training the network for panoptic segmentation. Their model does not require manually combining the different losses.

\section{Top-down Methods (One-stage)}
In this category of methods, the detection and then segmentation are performed but not in two stages. There is typically only one stage in which detection and segmentation occur and there is no need for proposal generation. \citet{de2020fast} propose \textbf{FPSNet}, where the panoptic segmentation module is formulated into a dense pixel classification module. Hence, their approach does not require heavy post-processing in terms of merging the predictions. \citet{Hou_2020_CVPR} propose a parameter free method to compute the mask by distilling information from the semantic segmentation and detection branches. Other one-stage approaches use RetinaNet \citep{lin2017focal} as their object detection network \citep{weber2020single,chen2020spatialflow}. \textbf{EPSNet} \citep{Chang_2020_ACCV} is an efficient panoptic segmentation model where a linear combination of prototype masks and mask coefficients are used to generate the masks. Another efficient approach is proposed by \citet{9466324} where an additional module is added to the detection network to predict the category- and instance-aware pixel embedding. 

\section{Bottom-up Methods}
This class of methods aims to avoid the computationally expensive instance mask prediction via bounding boxes. Therefore, they use segmentation masks to group or cluster pixels into different instances. \textbf{DeeperLab} \citep{yang2019deeperlab} is the first bottom-up panoptic segmentation approach proposed that parses the image in a single pass over a fully convolutional network. These predictions are then fused for the final output. \textbf{Panoptic-DeepLab} \citep{cheng2020panoptic} has the same semantic segmentation architecture as DeeperLab \citep{yang2019deeperlab} but it also incorporates dual ASPP for semantic segmentation and dual decoder network for instance segmentation. Other bottom-up methods use attention (\textbf{Axial-DeepLab}) \citep{wang2020axial}, pixel consensus voting (PCV) \citep{Wang_2020_CVPR} and position sensitive embedding for grouping instances \citep{chen2020panonet}.

\section{Single-path Methods}
These methods aim to unify the two tasks of semantic segmentation and instance segmentation. \textbf{Panoptic FCN} \citep{Li_2021_CVPR} address this by using a fully convolutional network for unified representation. By encoding an instance as a kernel, Panoptic FCN can predict the things and stuff directly via convolutions. However, a kernel fusion method is required for resolving overlaps. To avoid this, \citet{zhang2021k} restrict a kernel to predict a single mask for one object (\textbf{K-Net}). \textbf{DR1Mask} \citep{chen2020unifying} incorporates a dynamic rank-1 convolution (DR1Conv) module to learn a shared feature representation for both semantic and instance segmentation.

Transformer architecture has also proved useful for segmentation tasks. \citet{carion2020end} propose a transformer-based encoder-decoder architecture and bipartite matching for obtaining unique predictions called DEtection TRansformer (\textbf{DETR}). \citet{Zhang_2021_ICCV} propose a wearable system for assisting the visually impaired which consists of a transformer-based encoder network and a Transformer Parsing Module (TPM) for the decoder network. \citet{Wang_2021_CVPR} propose \textbf{MaX-DeepLab}, the first panoptic segmentation model where a mask transformer is trained with a panoptic quality metric based loss. \textbf{MaskFormer} \citep{cheng2021per} and \textbf{Mask2Former} \citep{cheng2022masked} propose that the mask prediction network is sufficient to solve both instance and semantic segmentation tasks using the same network, training parameters and loss function.

Other methods include \textbf{Bipartite CRF} \citep{jayasumana2019bipartite}, which address panoptic segmentation and cast it as a CRF model. \citet{chen2020scaling} study Wide Residual Networks (\textbf{Wide-ResNets}) as the backbone architecture for panoptic segmentation. Recent works on panorama panoptic segmentation (\cite{jaus2021panoramic}; \citeauthor{yang2020omnisupervised}, \citeyear*{yang2020omnisupervised}; \citeyear*{yang2021capturing}; \citeyear*{yang2021context}) also provide the benefits of wider field-of-view. Domain adaptive panoptic segmentation is studied by \citet{Huang_2021_CVPR} who use inter-style consistency and inter-task regularization to learn and generalize across domains.

\section{Weakly-supervised Learning}
\citet{Li_2018_ECCV} propose a weakly supervised panoptic segmentation model. Using GrabCut \citep{rother2004grabcut} and multiscale combinatorial grouping (MCG) \citep{arbelaez2014multiscale} to extract masks from annotated bounding boxes, the method assigns a label to only pixels with high confidence scores. Others are marked as ``ignore regions''. The ``stuff'' pixels are extracted by weakly supervising the model via image-level tags. A special case of weakly-supervised learning is omni-supervised learning, which uses both labeled and unlabeled data to train the model. \citet{yang2020omnisupervised} propose an omni-supervised learning segmentation model applied to panoramic images. Their proposed model uses unlabeled panoramic images as input along with labeled pinhole images, finally generating segmentation output on panoramic images. 

\section{Video Panoptic Segmentation}
\citet{kim2020video} propose to extend the task of panoptic segmentation to videos with \textbf{VPSNet}. Unlike image-based panoptic segmentation methods, this task is much more challenging. The reason is that this model needs to predict object categories, masks, bounding boxes, instance IDs, and semantic segmentation simultaneously, along with assigning a unique label to each pixel in the video. The authors also contribute two datasets and propose a new metric called Video Panoptic Quality (VPQ) for this task's evaluation. \textbf{ViP-DeepLab} \citep{Qiao_2021_CVPR} extends Panoptic-DeepLab \citep{cheng2020panoptic} and jointly performs video panoptic segmentation and monocular depth estimation \citep{saxena2005learning}. This method is proposed to solve the inverse projection problem. The inverse projection problem arises due to the ambiguous mapping from the retinal sources to the sources of retinal stimulation \citep{Qiao_2021_CVPR}. \citet{graber2021panoptic} study the task of panoptic segmentation forecasting where the goal is to predict the panoptic segmentation of future frames (of video), given a set of observed frames. 

\section{Panoptic Segmentation on LiDAR Point Clouds}
Panoptic segmentation on point clouds should generate a semantic class for each point and assign an instance ID to it. This is especially applicable in the field of autonomous driving. \citet{9340837} propose a single-stage real-time approach using a shared encoder along with a semantic and instance decoder. \textbf{DS-Net} \citep{hong2021lidar} use a dynamic shifting module to address the problems caused by uneven point cloud distribution. \citet{Zhou_2021_CVPR} propose a LiDAR point cloud panoptic segmentation framework named \textbf{Panoptic-PolarNet} where the backbone network based on PolarNet \citep{Zhang_2020_CVPR} generates the 3D semantic predictions and the 2D instance head is on top of it. Predictions from the two networks are then combined through majority voting. \citet{gasperini2021panoster} incorporate learning the clustering via a loss function into their network instead of grouping pixels in multiple steps. Their model, named \textbf{Panoster}, directly outputs the instance IDs and hence reduces model complexity.

\section{Discussion}
Panoptic segmentation has recently gained popularity due to its increased use cases and development of benchmark datasets and well-defined evaluation metrics. Nevertheless, it is still in its early stages and there remains big room for improvement. Table \ref{tab:panoptic} summarizes the main contribution, benchmark performances, and categorization of the popular panoptic segmentation methods. 

It is evident that methods with real-time capacity are preferred for many applications such as autonomous driving, and despite some methods having success in segmentation quality \citep{9466324}, there is still a large gap between most faster one-stage methods to the slower but better-performing two-stage methods. Single-path methods that unify instance and semantic segmentation tasks emerged as a new perspective and is drawing more attention as it tends to achieve a balance between speed and accuracy.

Panoptic segmentation should continue to draw its experience from new developments in semantic and instance segmentation as the combination of the two latter tasks is the key to achieving the former. However, panoptic segmentation benefits from additional information learned in multiscale \citep{porzi2021improving}, multitask \citep{saxena2005learning}, multi-modal \citep{Li_2018_ECCV} and multi-sensory \citep{yang2020omnisupervised} frameworks. It is crucial to keep exploring these learning approaches as better performance will be required for more complex and challenging applications.

\begin{scriptsize}
\begin{longtable}{|L{1.5cm}|L{3.7cm}|L{3.7cm}|L{1cm}|} 
\caption{Summary of Popular Panoptic Segmentation Methods.} \label{tab:panoptic} \\

\hline                         \textbf{Name} &                                                                                                                                                           \textbf{Main Contribution} &                                                                                                                                                         \textbf{Benchmarks Performances (default metric: PQ (\%))} &                  \textbf{Categorization} \\ \hline
\endfirsthead

\caption* {\textbf{Table 6.1 Continued}}\\

\hline  \textbf{Name} &                                                                                                                                                           \textbf{Main Contribution} &                                                                                                                                                         \textbf{Benchmarks Performances (default metric: PQ (\%))} &                  \textbf{Categorization} \\
\endhead

\hline \multicolumn{4}{|r|}{{Cont.}} \\ \hline
\endfoot

\hline \hline
\endlastfoot

            JSIS-Net &                                                                                   Joint semantic and instance segmentation with a single network &                                                                                                   COCO panoptic val (26.9) &        Top-down (two-stage) \\ \hline
                 Panoptic FPN &                                        Single network incorporating Feature Pyramid Network that generates instances and segments simultaneously &                                                                                COCO test-dev (40.9); Cityscapes val (58.1) &        Top-down (two-stage) \\ \hline
                  EfficientPS &                                 Shared backbone with 2-way FPN and parallel semantic and instance segmentation heads followed by panoptic fusion &                                                                                        Cityscapes val (67.1); KITTI (43.7) &        Top-down (two-stage) \\ \hline
                Auto-Panoptic &                                           Automatic architectural search for the best backbone, segmentation/instance branches and fusion module &                                                                                COCO panoptic val (44.8); ADE20K val (32.4) &        Top-down (two-stage) \\ \hline
                        OANet &                                                                  Spatial ranking module which addresses occlusions among the predicted instances &                                                                             COCO panoptic val (40.7); COCO test-dev (41.3) &        Top-down (two-stage) \\ \hline
                     OCFusion &                         Model overlap between two instances as a binary relation problem and train it with ground truth from dataset annotations &                                                      COCO panoptic val (46.3); COCO test-dev (46.7); Cityscapes val (60.2) &        Top-down (two-stage) \\ \hline
                       SOGNet &                                 Model the occlusions via a scene overlapping graph where objects are points and relations between them are edges &                                                      COCO panoptic val (43.7); COCO test-dev (47.8); Cityscapes val (60.0) &        Top-down (two-stage) \\ \hline
                       BGRNet &                                                                         Model the correlations between objects and the background with graph CNN &                                                                                COCO panoptic val (43.2); ADE20K val (31.8) &        Top-down (two-stage) \\ \hline
                       UPSNet &                                       Parameter-free panoptic head to merge the predictions from the semantic and instance segmentation networks &                                                      COCO panoptic val (43.2); COCO test-dev (46.6); Cityscapes val (61.8) &        Top-down (two-stage) \\ \hline
        \citet{li2020unifying} &                                                                                   End-to-end panoptic segmentation model without post-processing &                                                      COCO panoptic val (43.4); COCO test-dev (47.2); Cityscapes val (61.4) &        Top-down (two-stage) \\ \hline
                        AUNet &                                               Unified architecture with a foreground and background segmentation branches and a third RPN branch &                                                      COCO panoptic val (39.6); COCO test-dev (46.5); Cityscapes val (59.0) &        Top-down (two-stage) \\ \hline
    \citet{porzi2021improving} &                                                    Bounding box regression loss that addresses the problem caused by using crops during training &                                          Mapillary Vistas val (45.1); Indian Driving Dataset (50.7); Cityscapes val (66.7) &        Top-down (two-stage) \\ \hline
                  Ada-Segment &                                                                                    Automatically adjust multiple training losses during training &                                                                             COCO panoptic val (43.7); COCO test-dev (48.5) &        Top-down (two-stage) \\ \hline
                       FPSNet &                                                              Panoptic segmentation module is formulated into a dense pixel classification module &                                                                                   PASCAL VOC (57.8); Cityscapes val (55.1) &        Top-down (one-stage) \\ \hline
         \citet{Hou_2020_CVPR} &                        Parameter-free method to compute the mask by distilling information from the semantic segmentation and detection branches &                                                                            Cityscapes val (58.8); COCO panoptic val (37.1) &        Top-down (one-stage) \\ \hline
                       EPSNet &                                                 Linear combination of prototype masks and mask coefficients are used to generate the final masks &                                                                             COCO panoptic val (38.6); COCO test-dev (38.9) &        Top-down (one-stage) \\ \hline
               \citet{9466324} &                                  Additional module is added to the detection network to predict the category- and instance-aware pixel embedding &                                                                             COCO panoptic val (45.7); COCO test-dev (46.3) &        Top-down (one-stage) \\ \hline
                    DeeperLab &                        First bottom-up panoptic segmentation approach which parses the image in a single pass over a fully convolutional network &                                                                                               Mapillary Vistas val (31.95) &                 Single-path \\ \hline
             Panoptic-DeepLab &                                               Incorporates dual ASPP for semantic segmentation and dual decoder network for instance segmentation & COCO panoptic val (41.2); COCO test-dev (41.4); Cityscapes val (67.0); Cityscapes test (65.5); Mapillary Vistas val (40.3) &                 Single-path \\ \hline
                Axial-DeepLab &                                                                      Factorizing 2D self-attention into two 1D self-attentions for global region & COCO panoptic val (43.9); COCO test-dev (44.2); Mapillary Vistas val (41.1); Cityscapes val (68.5); Cityscapes test (66.6) &                 Single-path \\ \hline
                 Panoptic FCN &                                                              Using a fully convolutional network for unifying semantic and instance segmentation &                         COCO panoptic val (44.3); COCO test-dev (47.5); Mapillary Vistas val (36.9); Cityscapes val (61.4) &                 Single-path \\ \hline
                        K-Net &                                                                                        Restrict a kernel to predict a single mask for one object &                                                                             COCO panoptic val (54.6); COCO test-dev (55.2) &                 Single-path \\ \hline
                      DR1Mask &                                              Dynamic module to learn a shared feature representation for both semantic and instance segmentation &                                                                             COCO panoptic val (42.9); COCO test-dev (46.1) &                 Single-path \\ \hline
                         DETR &                                            Transformer-based encoder-decoder architecture and bipartite matching for obtaining unique predictions &                                                                                                   COCO panoptic val (45.1) &                 Single-path \\ \hline
                  MaX-DeepLab &                                   First panoptic segmentation model where a masktransformer is trained with a panoptic quality metric based loss &                                                                                                       COCO test-dev (51.3) &                 Single-path \\ \hline
                   MaskFormer &                                                                     Solves semantic and instance segmentation in unison with mask classification &                                                                                                   COCO panoptic val (52.7) &                 Single-path \\ \hline
                   Mask2Former &                                            Masked attention that restricts the attention to localized features centered around predicted segment &                                                                                                   COCO panoptic val (57.8) &                 Single-path \\ \hline
                Bipartite CRF &                                                                                                     Casting panoptic segmentation as a CRF model &                                                      PASCAL VOC (71.76); Cityscapes val (50.299); COCO panoptic val (41.7) &                 Single-path \\ \hline
          \citet{Li_2018_ECCV} &    First weakly-supervised method that jointly produces non-overlapping instance and semantic segmentation for both “thing” and “stuff” classes &                                                                                   PASCAL VOC (63.1); Cityscapes val (53.8) &  Weakly-super\-vised Learning \\ \hline
\citet{yang2020omnisupervised} & Uses unlabeled panoramic images and labeled pinhole images to generate segmentation on panoramic images in an omni-supervised learning framework &                                                                                                      Currently unavailable &  Weakly-super\-vised Learning \\ \hline
                       VPSNet &                                                    Proposes the task of video panoptic segmentation and re-organizes the synthetic VIPER dataset &                                                          VIPER (PQ=55.4, VPQ=51.9), Cityscapes-VPS val (PQ=62.2, VPQ=56.1) & Video Panoptic Segmentation \\ \hline
                  ViP-DeepLab &                            Jointly performs video panoptic segmentation and monocular depth estimation and solves the inverse projection problem &                                                                                              Cityscapes-VPS val (VPQ=63.1) & Video Panoptic Segmentation \\ \hline
    \citet{graber2021panoptic} &                                                       Predict the panoptic segmentation of future frames of video given a set of observed frames &                                                                                                      Cityscapes val (49.0) & Video Panoptic Segmentation \\ \hline
               \citet{9340837} &                              Single-stage real-time point cloud panoptic segmentation using a shared encoder and a semantic and instance decoder &                                                                                                  SemanticKITTI test (65.8) &          LiDAR Point Clouds \\ \hline
                       DS-Net &                                                        Dynamic shifting module to address the problems caused by uneven point cloud distribution &                                                   SemanticKITTI val (57.7); SemanticKITTI test (55.9); nuScenes val (42.5) &          LiDAR Point Clouds \\ \hline
            Panoptic-PolarNet &                                                  Combines 3D semantic predictions with PolarNet and 2D instance predictions with majority voting &                                                   SemanticKITTI val (59.1); SemanticKITTI test (54.1); nuScenes val (67.7) &          LiDAR Point Clouds \\ \hline
                     Panoster &                                                               Clustering pixels via a loss function instead of grouping pixels in multiple steps &                                                                        SemanticKITTI val (55.6); SemanticKITTI test (52.7) &          LiDAR Point Clouds \\ \hline
\end{longtable}
\end{scriptsize}

\chapter{Datasets}
In this section, we review the wide range of datasets that are often used as benchmarks for different segmentation methods. Many of these datasets serve as benchmarks for multiple tasks. We compare their main usages, scenes, sizes, and various other features, and discuss associated challenges and concerns.

The \textbf{Cityscapes} \citep{cordts2016cityscapes} dataset covers 30 object classes in urban street scenes and was collected from 50 cities spanning a duration of several months. About 5,000 images are finely annotated, and about 20,000 are coarsely annotated. It is widely used for urban street scene understanding and self-driving tasks. \textbf{Cityscapes Panoptic Parts} \citep{de2021part} extends the original Cityscapes dataset by adding part-aware panoptic segmentation annotations.

The \textbf{ADE20K} \citep{zhou2017scene} dataset is a densely annotated dataset spanning a total of 3,169 object classes across 1,072 complex everyday scenes. There are a total of 25k images with an average of 19.5 annotated instances and 10.5 annotated object classes per image. The Scene Parsing Benchmark (SceneParse150) is a smaller subset of the full ADE20K dataset containing 20k training images, 2k validation images, and a small test set with 150 semantic categories.

\textbf{ShapeNet} \citep{chang2015shapenet} is a large-scale 3D object database containing 3,135 classes. The ShapeNet Parts subset includes 31,693 meshes across 16 common object classes and is used primarily for 3D shape segmentation.

The \textbf{Stanford 3D Indoor Scene (S3DIS)} \citep{armeni20163d} dataset contains RGB-D and point cloud data from 5 large-scale indoor areas from three different buildings, covering multiple room types and diverse architectural styles. Each point of data is labeled with one of the 12 semantic classes representing structural elements of buildings and common furniture.

The \textbf{NYU-Depth V2 (NYUv2)} \citep{silberman2012indoor} dataset contains 1449 densely labeled pairs of aligned RGB and depth images from a variety of indoor scenes and is widely used for depth estimation and object segmentation tasks.

\textbf{ScanNet} \citep{dai2017scannet} is an RGB-D video dataset with both 2D and 3D data. The dataset is annotated with instance-level semantic segmentations and can be used for 3D scene understanding and semantic voxel labeling tasks. The newest version, ScanNet v2 \citep{dai2017scannet}, contains 1513 annotated scans covering 20 classes of annotated 3D voxelized objects.

The \textbf{Densely Annotation Video Segmentation} \textbf{(DAVIS)} \citep{perazzi2016benchmark} dataset is a benchmark dataset for video segmentation task consisting of high-quality video sequences. Per-frame dense pixel annotation is available for each video sequence. 

The \textbf{Sun RGB-D} \citep{song2015sun} dataset contains 10,335 densely annotated RGB-D images, totaling 146k 2D polygon and 64k 3D bounding box annotations. Each image is also labeled with additional 3D room layout and scene category, allowing it to serve as the benchmark for 3D object detection tasks.

\textbf{Semantic3D} \citep{hackel2017semantic3d} is a 3D point cloud dataset covering a wide range of urban scenes including churches, villages, streets, soccer fields, railroad tracks, etc. The dataset includes 15 training and 15 test scenes annotated with 8 class labels.

The \textbf{PASCAL Visual Object Classes Challenge} \textbf{(PASCAL VOC)} \citep{everingham2010pascal} dataset is built for image recognition and covers 20 object classes spanning five major categories – person, animal, vehicle, household, and other. Each image in this dataset is annotated with pixel-level segmentation, bounding boxes, and object classes. It includes 1,464 images for training, 1,449 images for validation, and a private test set. The PASCAL Context dataset \citep{mottaghi2014role} extends the PASCAL VOC 2010 detection challenge by annotating all pixels for training images, increasing the number of labeled classes to over 400. A subset of 59 classes are usually selected for use due to the sparsity of the remaining categories. \textbf{Pascal Panoptic Parts} \citep{de2021part} extends this dataset further by merging scene-level labels from PASCAL-Context with part-level labels from PASCAL-Part to create annotations for part-aware panoptic segmentation task.

The \textbf{Cambridge-driving Labeled Video Database} \textbf{(CamVid)} \citep{brostow2009semantic} is a driving/road scene understanding database containing 701 frames sampled from five video sequences captured with a 960x720 resolution camera on a car. These images are densely-labeled with 32 object classes including sidewalk, road, pedestrian, car, motorcycle, etc. 

The \textbf{Look into Person (LIP)} \citep{gong2017look} dataset contains 50k images of humans annotated in fine-grained details. It covers 19 human part category labels.

The \textbf{Microsoft COCO} \citep{lin2014microsoft} dataset contains more than 320k photos of 80 objects types and 40 scene categories. The COCO-Stuff \citep{caesar2018coco} dataset is constructed with per-pixel annotation on the original COCO dataset and spans 172 categories including 80 things, 91 stuff, and 1 unlabeled class. In addition, COCO has also been extended for panoptic segmentation by merging the 80 thing category and a subset of the 91 stuff category annotations.

The \textbf{Karlsruhe Institute of Technology and Toyota Technological Institute} \textbf{(KITTI)} \citep{geiger2012we} dataset is originally collected for the task of 3D object detection and tracking in autonomous driving and robotics. It has later been extended to include semantic \citep{alvarez2012road,zhang2015sensor,ros2015vision,behley2019semantickitti} and instance segmentation \citep{qi2019amodal} annotations for various research purposes.

The \textbf{YouTube-VOS (Youtube Video Object Segmentation)} dataset \citep{xu2018youtube} contains 4,453 videos totaling a duration of 340 minutes labeled with dense pixel-level annotations every 5th frame. More than 7,800 unique objects are also labeled for instance segmentation tasks.

The \textbf{nuScenes} \citep{caesar2020nuscenes} is a 3D video segmentation dataset containing 1000 scenes collected using a 32-beam LiDAR, 6 cameras and 5 radars covering full 360° views on the streets of Boston and Singapore. Each scene is 20 seconds long and annotated with 3D bounding boxes at 2Hz for 23 object classes, resulting in over 28k training samples, 6k validation samples, and 6k test samples. The nuImages \citep{caesar2020nuscenes} dataset contains 150h of video data collected using the same sensor suite in the same two cities. From these video frames, 93k images are annotated with 2D bounding boxes and instance masks for foreground and background classes.

The \textbf{Large Vocabulary Instance Segmentation (LVIS)} \citep{gupta2019lvis} dataset is a large-scale dataset containing 164k images annotated with more than 2 million instance segmentation masks. It poses the challenging task of long-tail object segmentation, covering more than 1200 categories with sparse per-category data.

\textbf{PartNet} \citep{mo2019partnet} is a 3D object dataset annotated with both instance-level and fine-grained hierarchical 3D parts segmentation. It contains over 570k part instances covering 24 object categories. PartNet serves as a benchmark dataset for fine-grained 3D semantic segmentation, hierarchical semantic segmentation, and instance segmentation tasks.

The \textbf{Digital Retinal Images for Vessel Extraction (DRIVE)} dataset \citep{staal2004ridge} consists a set of 40 fundus images collected through a diabetic retinopathy screening program in The Netherlands. 20 training images are labeled with retinal vessel segmentation by ophthalmological experts, and 20 test images are labeled by two different observers where one serves as the gold standard. This dataset enables comparative studies on blood vessels segmentation in retinal images which is utilized for diagnosis, treatment, and screening of ophthalmologic and cardiovascular diseases such as diabetes, hypertension, and arteriosclerosis.

The \textbf{PROMISE12} \citep{litjens2014evaluation} dataset contains T2-weighted Magnetic Resonance (MR) images of the prostates. The dataset was collected with several MRI vendors and scanning protocols from 50 patients with various diseases. Each image contains ground truth segmentation of the prostate, which is key to computer-aided diagnostic and detection of prostate cancer and prediction of the disease’s pathological stage. These insights are especially valuable for prognosis and treatment intervention.

The \textbf{Mapillary Vistas} dataset \citep{neuhold2017mapillary} is a diverse, large-scale street-level image dataset collected under various ambient conditions on a multitude of imaging devices by photographers with varying degrees of experience. It contains 25,000 high-resolution images with pixel-level annotations over 66 object categories and instance-specific labels for 37 classes. It is widely used for street scene understanding and provide benchmark for semantic, instance, and panoptic segmentation tasks.

The \textbf{VIsual PERception (VIPER)} \citep{richter2017playing} dataset is collected from a realistic virtual world and contains over 250K high-resolution video frames with annotations for various vision tasks including optical flow, object-level 3D scene layout, visual odometry, and semantic, instance, and panoptic segmentation.

The \textbf{Apolloscape} \citep{huang2019apolloscape} is a self-driving dataset consisting of over 140,000 video frames collected from various locations in China under varying weather conditions. The dataset contains pixel-wise semantic annotation and lane marking annotations in 2D, with point-wise semantic annotation in 3D for 28 classes.

The \textbf{Audi Autonomous Driving Dataset (A2D2)} \citep{geyer2020a2d2} is a multi-modal dataset recorded with camera, LiDAR, and autonomous bus. There are 41,277 non-sequential video frames with semantic segmentation, of which 12,497 frames also have 3D bounding box annotations.

The \textbf{BDD} \citep{yu2020bdd100k} dataset is a large-scale driving dataset containing 100K videos and supports 10 image recognition and autonomous driving - related tasks. It contains diverse scene types including city streets, residential areas, and highways, and diverse weather conditions at different times of the day \citep{yu2020bdd100k}.

The \textbf{Indian Driving Dataset (IDD)} \citep{varma2019idd} is another autonomous driving dataset containing 10K images annotated with 34 semantic classes collected from 182 driving sequences in multiple Indian cities.

The \textbf{WildDash} \citep{zendel2018wilddash} dataset is another autonomous driving dataset and provides benchmark for panoptic, semantic, and instance segmentation. It is a diverse dataset with 5000+ traffic scenarios collected in city, highway and rural locations from more than 100 countries under various driving conditions such as poor weather, overexposure, and darkness.

The \textbf{Dense Panoramic Semantic Segmentation (DensePASS}) \citep{ma2021densepass} is a benchmark dataset for panoramic semantic segmentation. It contains 100 labelled 2,000 unlabeled panoramic images 
collected using Google Street View from locations around the world. It is used for the task of Pinhole to Panoramic domain transfer and the labelled images are annotated with 19 categories present in the source domain dataset (i.e. Cityscapes \citep{cordts2016cityscapes}).

The \textbf{SYNTHIA} dataset \citep{ros2016synthia} is a synthetic dataset containing 9,400 photo-realistic frames rendered from a virtual city. It covers a diverse set of scenes including European style town, modern city, highway and green areas, and comes with pixel-level semantic annotations for 13 classes.

The \textbf{Grand Theft Auto 5 (GTA5)} dataset is another synthetic dataset containing 24,966 synthetic images rendered using the open-world video game Grand Theft Auto 5, all from the car perspective in the streets of American-style virtual cities \citep{richter2016playing}. The images are labeled with pixel level semantic annotation for 19 categories present in the Cityscapes dataset.

The \textbf{2D-3D-Semantic (2D-3D-S)} is a large-scale indoor dataset that enables joint and cross-modal learning \citep{armeni2017joint}. It consists of over 70,000 RGB images with corresponding depths, surface normals, semantic annotations, global XYZ images along with camera information. The dataset also includes registered raw and semantically annotated 3D meshes and point clouds \citep{armeni2017joint}.

The list of tasks each dataset can be applied to is summarized in Table \ref{tab:datatask}. Table \ref{tab:datasize} summarizes the important information associated with each dataset. It is evident that there is a huge variety of datasets, and that there is no ``one-size-fits-all'' solution to the various segmentation problems. A certain model performing well on one benchmark dataset might fail miserably on another, which makes it critical to perform a comprehensive evaluation of a trained model before declaring it production-ready, particularly for applications with significant health, security, and safety concerns. Moreover, monitoring model performance after deployment is equally important as 1) data drift can be rapid in many fast-developing industrial applications, and 2) deep learning models are often susceptible to minor perturbations in the input signals \citep{akhtar2018threat}.

\begin{scriptsize}
\begin{longtable}{|L{1.6cm}|L{0.23cm}|L{0.23cm}|L{0.23cm}|L{0.23cm}|L{0.23cm}|L{0.23cm}|L{5.1cm}|} 
\caption{Summary of Dataset and Associated Tasks} \label{tab:datatask} \\

\hline Dataset Name & \rot{Object Detection} &  \rot{Semantic Segmentation} &  \rot{Instance Segmentation} &  \rot{Panoptic Segmentation} &  \rot{3D segmentation} &  \rot{Video segmentation} & Others \\ \hline
\endfirsthead

\caption* {\textbf{Table 7.1 Continued}}\\

\hline Dataset Name & \rot{Object Detection} &  \rot{Semantic Segmentation} &  \rot{Instance Segmentation} &  \rot{Panoptic Segmentation} &  \rot{3D segmentation} &  \rot{Video segmentation} & Others \\ \hline
\endhead

\hline \multicolumn{8}{|r|}{{Cont.}} \\ \hline
\endfoot

\hline \hline
\endlastfoot

Cityscapes     &                  &  \checkmark &            \checkmark &            \checkmark &                 &         \checkmark &                                     Image-to-image Translation, Real-Time Semantic Segmentation \\ \hline

ShapeNet       &                  &            \checkmark &                       &                       &      \checkmark &                    &                                                                               3D Reconstruction \\ \hline

S3DIS             & &            \checkmark &                       &                       &      \checkmark &                    &                                                                                                 \\ \hline

 NYUv2           &       \checkmark &            \checkmark &            \checkmark &                       &                 &                    &                       Monocular Depth Estimation, Depth Completion, Surface Normals Estimation \\ \hline

ScanNet          &       \checkmark &            \checkmark &                       &            \checkmark &      \checkmark &                    &              Depth Estimation, Scene Recognition, 3D Reconsturction, Surface Normals Estimation \\ \hline

DAVIS 	  &       \checkmark &                       &                       &                       &                 &         \checkmark &                               Interactive Segmentation, Video Denoising, Video Object Detection \\ \hline

ADE20K           &                  &            \checkmark &                       &                       &                 &                    &                                                 Image-to-Image Translation, Scene Understanding \\ \hline

Sun RGB-D       &       \checkmark &            \checkmark &                       &                       &                 &                    &                                                                          Room Layout Estimation \\ \hline

Semantic3D      &                 &            \checkmark &                       &                       &                 &                    &                                                                                                 \\ \hline

PASCAL VOC    & \checkmark &            \checkmark &            \checkmark &            \checkmark &                 &                    &                                                                                  Graph Matching \\ \hline

CamVid             &                 &            \checkmark &                       &                       &                 &         \checkmark &                                                                                                 \\ \hline

LIP 		    &               &            \checkmark &                       &                       &                 &                    &                                                                                                 \\ \hline

COCO		    &       \checkmark &            \checkmark &            \checkmark &            \checkmark &                 &                    &        Pose Estimation, Keypoint Detection, Text-to-Image Generation, Visual Question Answering \\ \hline

KITTI 		    &       \checkmark &            \checkmark &                       &            \checkmark &                 &                    & Vehicle Pose Estimation, Monocular Depth Estimation, Multiple Object Tracking, Depth Completion \\ \hline

YouTube-VOS    &                  &                       &                       &                       &                 &         \checkmark &                                                        Visual Object Tracking, Video Inpainting \\ \hline

nuScenes           &       \checkmark &            \checkmark &            \checkmark &                       &      \checkmark &                    &                                                3D Multi-Object Tracking,  Trajectory Prediction \\ \hline

nuImages          &       \checkmark &            \checkmark &            \checkmark &                      &                 &                    &                                                                                                 \\ \hline

LVIS                   &       \checkmark &                       &            \checkmark &                       &                 &                    &                                                                                                 \\ \hline

PartNet             &                &                       &                       &                       &      \checkmark &                    &                                                                                                 \\ \hline

DRIVE               &                  &            \checkmark &                       &                       &                 &                    &                                                                                                 \\ \hline

Mapillary Vistas &       \checkmark &            \checkmark &            \checkmark &            \checkmark &                 &                    &                                                                                                 \\ \hline

VIPER                &       \checkmark &            \checkmark &            \checkmark &            \checkmark &      \checkmark &         \checkmark &  Optical Flow, Visual Odometry \\ \hline

PROMISE12      &                  &            \checkmark &                       &                       &                 &                    &                                                                                                 \\ \hline

Apolloscape      &                  &            \checkmark &            \checkmark &                       &                 &                    &                                                        Self-Localization, Lanemark Segmentation \\ \hline

A2D2                &       \checkmark &            \checkmark &            \checkmark &                      &      \checkmark &                    &                                                                                                 \\ \hline

BDD                  &       \checkmark &            \checkmark &            \checkmark &                       &                 &         \checkmark &                                        Multi-Object Tracking, Lane Detection, Domain Adaptation \\ \hline

IDD                  &       \checkmark &            \checkmark &                        &                       &                 &                    &                                                                                                 \\ \hline

WildDash         &                 &            \checkmark &            \checkmark &            \checkmark &                 &                    &                                                                                                 \\ \hline

DensePASS     &                  &            \checkmark &                       &                       &                 &                    &                                                       Panoramic Segmentation, Domain Adaptation \\ \hline

SYNTHIA         &                  &            \checkmark &                       &                       &                 &                    &                                                   Domain Adaptation, Image-to-Image Translation \\ \hline

GTA5               &                  &            \checkmark &                       &                       &                 &                    &                                                   Domain Adaptation, Image-to-Image Translation \\ \hline

2D-3D-S          &       \checkmark &            \checkmark &                       &                       &      \checkmark &                    &                              Depth Estimation, Surface Normals Estimation, Scene Reconstruction \\ \hline
\end{longtable}
\end{scriptsize}
\begin{scriptsize}
\begin{longtable}{|L{1.5cm}|L{2.7cm}|L{4.5cm}|L{1cm}|} 
\caption{Dataset Scene, Size, and Variety.} \label{tab:datasize} \\

\hline     \textbf{Dataset Name} &                                                                                                        \textbf{\# of object classes} &                                                                                                                        \textbf{Size} &   \textbf{Scene} \\ \hline
\endfirsthead

\caption* {\textbf{Table 7.2 Continued}}\\

\hline     \textbf{Dataset Name} &                                                                                                        \textbf{\# of object classes} &                                                                                                                        \textbf{Size} &   \textbf{Scene} \\ \hline
\endhead

\hline \multicolumn{4}{|r|}{{Cont.}} \\ \hline
\endfoot

\hline \hline
\endlastfoot

  Cityscapes &                                                                                                                         30 &                                                                        5k fine annotated images, 20k coarse annotated images &  street \\ \hline
        ShapeNet &                                                                                                                       3135 &                                                                                            Over 300M models, 220k classified & variety \\ \hline
           S3DIS &                                                                                                                         12 &                                                                                    6 large-scale indoor areas with 271 rooms &  indoor \\ \hline
           NYUv2 &                                                                                                                         26 & 1,449 densely labeled pairs of aligned RGB and depth images, 464 new scenes taken from 3 cities 407,024 new unlabeled frames &  indoor \\ \hline
         ScanNet &                                                                                                                         20 &                                             1,513 annotated scans of labeled voxels with an approximate 90\% surface coverage &  indoor \\ \hline
           DAVIS &                                                                                                                            &                                                        50 video sequences with 3,455 densely annotated frames in pixel level & variety \\ \hline
          ADE20K &                                                                                                                        150 &                             20K scene-centric images exhaustively annotated with pixel-level objects and object parts labels & variety \\ \hline
       Sun RGB-D &                                                                                                                        700 &                      10,335 real RGB-D images of room scenes. Each RGB image has a corresponding depth and segmentation map &  indoor \\ \hline
      Semantic3D &                                                                                                                          8 &                                                                        Over 3 billion points, 15 training and 15 test scenes & outdoor \\ \hline
      PASCAL VOC &                                                                                                                         20 &                                             1,464 images for training, 1,449 images for validation and a private testing set & variety \\ \hline
          CamVid &                                                                                                                         32 &                                                                                                                   701 frames  &  street \\ \hline
             LIP &                                                                                                                         19 &                                                                            50k images with elaborated pixel-wise annotations &  people \\ \hline
            COCO & object detection: 80, keypoints detection: 17, stuff image segmentation: 91, panoptic: 80 thing and 3 stuff categories  &                                                                                                                  328K images & variety \\ \hline
           KITTI &                                                                                                                            &                                                                        180 GB sensor streams with images, OXTS, and Velodyne &  street \\ \hline
     YouTube-VOS &                                                                                                                       7800 & 4,453 videos, 340+ minutes, labeled with dense pixel-level annotations every 5th frame, has instance segmentaion annotations &         \\ \hline
        nuScenes &                                                                                                                         10 &                                                                                                28,130 samples from 1k scenes  &  street \\ \hline
        nuImages &                                                                                                                         10 &                                                                                                         93k annotated images  &  street \\ \hline
            LVIS &                                                                                                                        1k+ &                                                                                                                  164k images & variety \\ \hline
         PartNet &                                                                                                                         24 &                                                                                 573,585 part instances over 26,671 3D models &  indoor \\ \hline
           DRIVE &                                                                                                                            &                                                                                                                    40 images  & medical \\ \hline
Mapillary Vistas &                                                                             66 semantic categories and 37 instance classes &                                                                                                  25k high-resolution images  &  street \\ \hline
           VIPER &                                                                                                                         11 &                                                                                            250K high-resolution video frames & virtual \\ \hline
       PROMISE12 &                                                                                                                            &                                                                                                            50 training cases & medical \\ \hline
            Apolloscape &                                                                                35 semantic classes and 8 instance classes  &                                                                                                         140,000 video frames &  street \\ \hline
            A2D2 &                                                                                                                         38 &                                                                          41,277 annotated frames, 392,556 unannotated frames &  street \\ \hline
             BDD &                                                                                                                         40 &                                                                                                                  100k videos  &  street \\ \hline
             IDD &                                                                                                                         34 &                                                                                                                10,004 images  &  street \\ \hline
        WildDash &                                                                                 28 semantic classes and 9 instance classes &                                                                                                                 1,800 frames &  street \\ \hline
       DensePASS &                                                                                                                         19 &                                                                            100 labelled and 2,000 unlabeled panoramic images &  street \\ \hline
         SYNTHIA &                                                                                                                         13 &                                                                                                                  9,400 frames  & virtual \\ \hline
            GTA5 &                                                                                                                         19 &                                                                                                       24,966 synthetic images  & virtual \\ \hline
         2D-3D-S &                                                                                                                         13 &                                                                                                            70,496 RGB images  &  indoor \\ \hline
\end{longtable}
\end{scriptsize}

\chapter{Evaluation Metrics}
\section{Pixel Accuracy / Accuracy}
Pixel accuracy is the most straightforward metric for evaluating segmentation quality. It is defined as the percentage of pixels that are classified correctly in an image. This metric is simple to calculate but has many drawbacks. Similar to the accuracy used in traditional machine learning classification tasks, pixel accuracy suffers from the class imbalance issue. For example, if over 90\% of the pixels belong to class A and the rest belongs to class B in the ground truth image, and all the pixels in the predicted segmentation are assigned to class A, then the pixel accuracy would be over 90\%. However, this model is practically useless as it does not show any ability to distinguish B from A.

\section{Mean Intersection Over Union (mIoU)}
\enlargethispage{\baselineskip}
Intersection over union (IoU) is defined as the overlap (intersection) area divided by the union area between the ground truth and predicted segmentation. Mean IoU (mIoU) is widely used for multi-class semantic segmentation and is calculated by averaging the IoUs calculated for each class. It is a much more effective measure than the pixel accuracy and does not suffer from the class imbalance issue.

\section{Mean Average Precision (mAP)}
In traditional classification tasks, precision measures the quality of positive predictions, and is defined as the ratio of true positives and the total predicted positives. Recall, on the other hand, is defined by dividing the number of true positives by the total number of ground truth positives. In computer vision, mean average precision (mAP) is a widely used metric for evaluating object detection and instance segmentation tasks. To calculate the mAP, IoU is first calculated for each prediction, and a threshold (e.g. 0.5) along with the ground truth and predicted class label are used to determine whether the prediction is a true positive, false positive, or false negative. A precision-recall (PR) curve is then created with these three values for each class using interpolated precision, which is the maximum precision value measured at each recall level. Each detected instance also has an associated confidence value that is used to rank the output when drawing the PR curve. Average precision (AP) can be calculated by taking the area under the curve, and several variations are often used. The primary challenge metric for the most widely-used benchmark – MS-COCO \citep{lin2014microsoft} – is AP@[.5:.95], which means that AP is averaged over IoU thresholds from 0.5 to 0.95 with steps of 0.05 (10 different AP values). Alternatively, AP@.5 and AP@.75 are also widely used metrics that correspond to AP values calculated at IoU=0.5 and IoU=0.75 respectively. Finally, mAP can be obtained by computing AP for each object class and averaging them. It is worth to note that for MS-COCO, AP and mAP are often used interchangeably.

\section{Mean Average Recall (mAR)}
Average Recall (AR) is a less-used metric for image segmentation tasks. Originally, it is designed to evaluate the performance of region proposals \citep{hosang2015makes}. Similar to AP, AR is averaged over the values measured at the same 10 IoU thresholds, but unlike AP, it does not rely on the confidence score for each detection, and all detections are considered positive. For MS-COCO \citep{lin2014microsoft}, mean AR is calculated by averaging the largest recall values over IoUs and object categories given a fixed number of detections per image \citep{lin2014microsoft}. AR is also used interchangeably with mAR.

\section{Dice Score (Coefficient)}
Dice Score is calculated similarly to F1-score. It is equivalent to the ratio between two times the overlapping area and the total number of pixels of the ground truth image and the segmentation output.

\section{Panoptic Quality (PQ)}
\begin{figure}[h]
\centering
\includegraphics[width=0.9\textwidth]{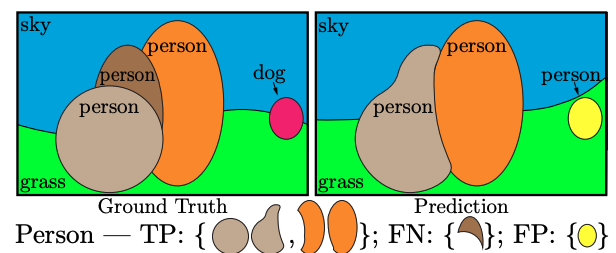}
\caption{Toy example showing ground truth and predicted panoptic segmentations of an image. From \citet{kirillov2019panoptic}.}
\label{fig:pq}
\end{figure}

Panoptic Quality (PQ) is designed to measure the performance of both semantic (stuff) and instance (things) segmentation in a unified manner. It is calculated in two steps: (1) segment matching and (2) PQ computation, and can be seen as a combination of two terms - Segmentation Quality (SQ) and Recognition Quality (RQ). An IoU threshold of 0.5 is used to determine whether a predicted and a ground truth segment match. Then, RQ evaluates the quality of detections based on precision and recall (which results in a term resembling the F1-score), and SQ evaluates the quality of predicted segments based on how much they overlap with the ground truth based on the IoU. Figure~\ref{fig:pq} illustrates ground truth and predicted panoptic segmentations of an image with a toy example.

\section{Video Panoptic Quality (VPQ)}
\begin{figure}[h]
\centering
\includegraphics[width=0.9\textwidth]{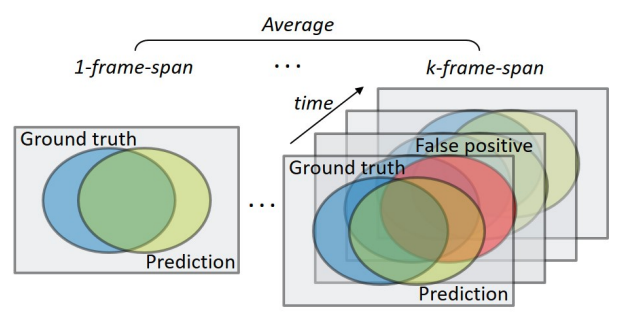}
\caption{Tube matching and video panoptic quality (VPQ) metric. From \citet{kim2020video}.}
\label{fig:vpq}
\end{figure}

Video Panoptic Quality (VPQ) \citep{kim2020video} is adapted from Panoptic Quality \citep{kirillov2019panoptic} and aims to take into account the temporal element when considering the task of video panoptic segmentation. VPQ is obtained over a span of \textit{k} frames with 3D tube matching, where each tube is the sequence of panoptic segments over the \textit{k} spans (Figure \ref{fig:vpq}). IoU is calculated between matching predicted and ground truth tubes, and false positive segments from each frame penalizes the whole predicted tube. VPQ\textsuperscript{k} is defined as the average of PQ computed for each of the \textit{k} frames. Multiple VPQ\textsuperscript{k} values are obtained for each selection of \textit{k} and the average of these values produces the final VPQ score. This metric captures how video panoptic segmentation is challenged by the object segment tracking across frames, especially for longer spans.

\chapter{Challenges and Future Directions}
The landscape of object segmentation has been ever-changing as a result of newly developed deep learning architectures and expanding domains of applications. In this section, we identify and summarize the major challenges associated with object segmentation and discuss open questions and future directions in this field.

\section{Challenges}
\textbf{Balance between speed, resource, and accuracy.} As the variety of object segmentation approaches continues to increase, the question of how to maintain high accuracy while reducing computational time and other resources has become more manageable. \citet{li2019dfanet}, \citet{orsic2019defense}, \citet{bolya2019yolact}, and \citet{wang2020solo} are examples of semantic and instance segmentation methods that achieve good balances between efficiency and accuracy. However, for panoptic, 3D, and video segmentation, the availability of such methods still remains low. For panoptic segmentation, the main problem that needs to be addressed before a good balance can be found is how to merge the results from the semantic and instance segmentation branches. For 3D segmentation, heuristics-based methods such as K-nearest neighbors search have shown to improve efficiency in capturing point-based features, but accuracy suffers as a result of considering only local features. For video segmentation, due to the nature of the problem, despite the improvements from exploring temporal continuity, the efficiency of the existing approaches remains to be the biggest limitation.

\textbf{Low-quality images and complex scenes.} Many state-of-the-art segmentation approaches still have difficulty when dealing with low-quality images and complex scenes. This is an important problem to address because real-world situations are often complex and can be subject to various adverse conditions such as a crowded room with many occlusions, a nighttime scene with low light and blurry vision, or an accident scene at a busy street. While human perception is enabled by our visual systems to quickly interpolate and identify obscure or occluded objects given other visual cues \citep{palmer1999vision}, this process is challenging for our machine counterparts mainly due to the deterministic nature of most visual recognition algorithms, which lack the ability to transfer knowledge \citep{hao2020brief}. Several lines of approaches have been explored to tackle these issues. For example, GANs have been used to synthesize occluded parts \citep{ehsani2018segan} and transforming scenes \citep{orsic2019defense}, and amodal perception datasets \citep{zhu2017semantic} have been created for the task of amodal semantic segmentation, which aims to reason about interactions of objects and achieve de-occlusion.

\textbf{Availability of high-quality densely annotated datasets.}
The majority of existing segmentation models are trained based on a fully supervised approaches, relying heavily on high-quality densely annotated datasets, which are extremely time-consuming to obtain due to the difficulty associated with annotation of irregular object boundaries, especially in complex scenes and low-resolution images or videos. For 3D and panoptic segmentation tasks with multiple objectives, it is even more challenging to construct datasets that allow for unified training schemes. This problem has been somewhat alleviated due to the development of semi-supervised and unsupervised approaches, but performance of these algorithms are not as competitive.

\textbf{Domain adaptation.}
The increasingly wide range of applications calls for increased adaptability of segmentation models given the difficulty involved in annotating datasets and the high requirements for computational resources. It is not sustainable to modify existing model architectures for a domain-specific task, or even to perform transfer learning due to the lack of high-quality training data. The variety of the existing datasets alleviates this issue by providing multiple benchmarks, but it will be increasingly challenging to develop more complex datasets to meet higher requirements. Recently, synthetic datasets \citep{brodeur2017home, wu2018building} have started to gain momentum for their low cost of development, particularly for 3D segmentation tasks. Utilization of these datasets requires more careful consideration of the domain adaptation issue. Finally, video segmentation models must be robust to unforeseen but common object categories in real-world circumstances, as opposed to the fixed categories they were trained to recognize. This is particularly important because video segmentation applications such as self-driving cars cannot afford to miss hazardous objects that may lead to catastrophic accidents.

\textbf{Reproducibility.}
While many existing works describe their experimental setup in detail and provide source code repositories for implementation of their algorithms, others do poorly in these documentations. In particular, there is a significant shortage of well-documented 3D and video segmentation approaches, likely due to the complex nature of these tasks. The lack of reproducibility prevents meaningful comparative analysis to be performed and hinders the continuous advances in these areas.

\textbf{Ethical Concerns.}
Successfully deploying a segmentation model to practical situations requires many considerations. One of the questions that often arise is what constitutes a ``correct'' segmentation. This is a challenging problem, and the answer is arguably task-dependent. Particularly, for applications that are safety- or security-critical such as medical diagnosis and autonomous driving, the requirement for the quality of segmentation is often much higher than for other applications such as fashion and home improvement. Deep learning models, often considered as black boxes, pose ethical concerns when applied to these safety-critical tasks. What further exacerbates these concerns are that these models are often susceptible to adversarial attacks \citep{akhtar2018threat} or even extraneous signals \citep{zech2018radiological}. These perturbations are often subtle, but can significantly impact the predicted outcome and pose threats to human health and safety when applied without proper defense mechanisms in place. Due to these concerns, discussions about transparency and interpretability of deep models have become increasingly prevalent. One example is the push for better understanding of the specific features these deep models rely on when making predictions, thus allowing us to make informed decisions based on these model outputs.

\section{Future Directions}
\textbf{Learning from weakly-labeled and unlabeled datasets.} Because dense, high-quality annotations for object segmentation tasks are very difficult to acquire, particularly for video and 3D datasets, there is a growing trend towards learning from unlabeled datasets using unsupervised and semi-supervised approaches. Semi-supervised approaches such as partially-labeled datasets, image-level labels and bounding box annotations have proven to be sufficient for training segmentation models \citep{papandreou2015weakly, khoreva2017simple, souly2017semi, Li_2018_ECCV}, and unsupervised approaches such as clustering \citep{ji2019invariant} and automatic label generation \citep{li2018so,zhao20193d} have also shown promises. This realm of research is particularly important for certain domains of application such as medical image segmentation, where scarce data and noisy and sparse annotations are often the limiting factors for a production-ready segmentation model.

\textbf{Synthetic datasets and domain adaptation.} Synthetic datasets \citep{wu2018building, ros2016synthia} have recently begun to shift the landscape of the computer vision by reducing the effort and cost of manually labeling image datasets and improving the variety of the training data, both of which are key limitations of real-world datasets. Manually curated datasets for video segmentation, particularly for autonomous driving applications such as CamVid \citep{brostow2009semantic}, suffer from bias introduced by their acquisition in specific locations and the low variety of objects and scenes. Similarly, real-world 3D datasets are collected with expensive equipment and are often subject to noise and occlusion \citep{silberman2012indoor, song2015sun, dai2017scannet, armeni20163d, behley2019semantickitti}, whereas synthetic 3D datasets benefit from clear, 360-degree views of objects \citep{wu2018building}. On the other hand, domain discrepancy between synthetic and real-world data can often cause issues when adapting from research to applications and should be resolved before synthetic datasets can truly realize their potential. Recently, growing efforts have been made to address this challenge \citep{kamnitsas2017unsupervised, huang2018domain, hong2018conditional, sankaranarayanan2018learning, huang2021cross}.

\textbf{3D, long-term and open world video segmentation.} Video segmentation is unique in many aspects and there are many lines of related research that are still at their infant stage. 3D video segmentation, for example, has been studied only briefly \citep{choy20194d, shi2020spsequencenet}, but have already shown great potential as spatio-temporal features have proven valuable for improving the quality of segmentation output. Long-term and open world video segmentation are both crucial for practical applications, but remain under-explored due to the lack of longer and more diverse sequences of training data \citep{wang2021survey}. With the advent of self-driving cars and the availability of more open-world video data, there is a lot of promise for more powerful and adaptable video segmentation techniques.

\textbf{Real-time segmentation.}
Real-time methods are becoming increasingly desirable as object segmentation methods mature and applications become more demanding. They lay the critical foundations for self-driving cars, augmented reality, robots and even medical imaging \citep{ouahabi2021deep}. Despite the success of recent real-time segmentation approaches \citep{li2019dfanet, pham2019real, yu2018bisenet, lee2020centermask, bolya2019yolact}, there remains to be a significant performance gap to models that are built for offline tasks in terms of accuracy. Several key factors that should be considered in this research direction include improving the efficiency of feature representations, unifying training objectives, adapting network architectures for parallel processing, and maximizing the utilization of advanced GPU hardware.

\textbf{Multi-objective and multi-modal learning.} The continuous development of panoptic segmentation shows the trend towards multi-objective learning. Semantic segmentation can be naturally extended to accomplish other tasks such as object detection \citep{wu2019detectron2, meyer2019sensor}, instance segmentation \citep{he2017mask}, pose estimation \citep{xia2017joint}, depth estimation \citep{cheng2020panoptic}, scene completion \citep{dai2018scancomplete}, representation learning \citep{liao2013representation}, and image editing \citep{chen2020harnessing, xu2017deep}. Through exploiting a diverse set of features, deep learning models can integrate these tasks into a unified framework that has the potential to achieve better performance than when attempting to achieve each objective individually. Multi-objective learning also calls for multi-modal learning, where multi-sensory (e.g. RGB-D, LiDAR) information and multi-representations (voxels, point clouds) can lead to better scene understanding and higher accuracy, especially for complex and noisy datasets \citep{chiang2019unified, dai20183dmv}.

\textbf{Automated search for optimal network architectures.}
As the requirements from various application domains continue to rise, it will become increasingly difficult to handcraft suitable network architectures for domain-specific tasks despite the abundance and variety of existing models. Automated search of optimal structures should gradually become the desired technique for the design of new segmentation networks in order to meet larger-scale applications. This will be made increasingly more possible through research in neural architecture search \citep{elsken2019neural} and reinforcement learning \citep{zoph2016neural}.

\chapter{Conclusion}
In this monograph, we reviewed the development and many variations of segmentation approaches. Traditional feature-based approaches were surpassed by deep-learning-based approaches as convolutional neural networks started to dominate the field of computer vision. Earlier works focused on semantic segmentation problems, while instance segmentation and panoptic segmentation approaches started to develop as R-CNN based object detection approaches began laying the important groundwork. 3D and Video segmentation tasks have also become increasingly important as 3D and LiDAR scanners became more widespread and large-scale video segmentation datasets became increasingly available. We performed detailed reviews of the most impactful and signature works and provided overviews of a wide range of papers in each category. We categorized each method based on its architecture, and compared the main components and performances of the various approaches both qualitatively and quantitatively. In addition, we reviewed the main metrics for evaluating and comparing segmentation approaches and surveyed the major datasets that serve as benchmarks for each task. Finally, we identified the major challenges facing various object segmentation methods today and proposed several promising future directions. The goal of this monograph is to broaden the readers' understanding of the object segmentation landscape and provide them with options to quickly select the appropriate approaches for their research or applications. Finally, researchers can leverage this review to further advance these approaches in the future.


\begin{acknowledgements}
The authors are grateful to Osama Sakhi, who assisted in formatting and generating the tables.
\end{acknowledgements}

\backmatter  

\printbibliography

@article{hochreiter1997long,
  title={Long short-term memory},
  author={Hochreiter, Sepp and Schmidhuber, J{\"u}rgen},
  journal={Neural computation},
  volume={9},
  number={8},
  pages={1735--1780},
  year={1997},
  publisher={MIT Press}
}

@inproceedings{deng2009imagenet,
  title={Imagenet: A large-scale hierarchical image database},
  author={Deng, Jia and Dong, Wei and Socher, Richard and Li, Li-Jia and Li, Kai and Fei-Fei, Li},
  booktitle={2009 IEEE conference on computer vision and pattern recognition},
  pages={248--255},
  year={2009},
  organization={Ieee}
}

@article{biederman1985human,
  title={Human image understanding: Recent research and a theory},
  author={Biederman, Irving},
  journal={Computer vision, graphics, and image processing},
  volume={32},
  number={1},
  pages={29--73},
  year={1985},
  publisher={Elsevier}
}

@article{huang2019apolloscape,
  title={The apolloscape open dataset for autonomous driving and its application},
  author={Huang, Xinyu and Wang, Peng and Cheng, Xinjing and Zhou, Dingfu and Geng, Qichuan and Yang, Ruigang},
  journal={IEEE transactions on pattern analysis and machine intelligence},
  volume={42},
  number={10},
  pages={2702--2719},
  year={2019},
  publisher={IEEE}
}

@article{geyer2020a2d2,
  title={A2d2: Audi autonomous driving dataset},
  author={Geyer, Jakob and Kassahun, Yohannes and Mahmudi, Mentar and Ricou, Xavier and Durgesh, Rupesh and Chung, Andrew S and Hauswald, Lorenz and Pham, Viet Hoang and M{\"u}hlegg, Maximilian and Dorn, Sebastian and others},
  journal={arXiv preprint arXiv:2004.06320},
  year={2020}
}

@inproceedings{yu2020bdd100k,
  title={Bdd100k: A diverse driving dataset for heterogeneous multitask learning},
  author={Yu, Fisher and Chen, Haofeng and Wang, Xin and Xian, Wenqi and Chen, Yingying and Liu, Fangchen and Madhavan, Vashisht and Darrell, Trevor},
  booktitle={Proceedings of the IEEE/CVF conference on computer vision and pattern recognition},
  pages={2636--2645},
  year={2020}
}

@inproceedings{zendel2018wilddash,
  title={Wilddash-creating hazard-aware benchmarks},
  author={Zendel, Oliver and Honauer, Katrin and Murschitz, Markus and Steininger, Daniel and Domin\-guez, Gustavo Fernandez},
  booktitle={Proceedings of the European Conference on Computer Vision (ECCV)},
  pages={402--416},
  year={2018}
}

@inproceedings{richter2016playing,
  title={Playing for data: Ground truth from computer games},
  author={Richter, Stephan R and Vineet, Vibhav and Roth, Stefan and Koltun, Vladlen},
  booktitle={European conference on computer vision},
  pages={102--118},
  year={2016},
  organization={Springer}
}

@article{armeni2017joint,
  title={Joint 2d-3d-semantic data for indoor scene understanding},
  author={Armeni, Iro and Sax, Sasha and Zamir, Amir R and Savarese, Silvio},
  journal={arXiv preprint arXiv:1702.01105},
  year={2017}
}

@inproceedings{ma2021densepass,
  title={Densepass: Dense panoramic semantic segmentation via unsupervised domain adaptation with attention-augmented context exchange},
  author={Ma, Chaoxiang and Zhang, Jiaming and Yang, Kailun and Roitberg, Alina and Stiefelhagen, Rainer},
  booktitle={2021 IEEE International Intelligent Transportation Systems Conference (ITSC)},
  pages={2766--2772},
  year={2021},
  organization={IEEE}
}

@inproceedings{varma2019idd,
  title={IDD: A dataset for exploring problems of autonomous navigation in unconstrained environments},
  author={Varma, Girish and Subramanian, Anbumani and Namboodiri, Anoop and Chandraker, Manmohan and Jawahar, CV},
  booktitle={2019 IEEE Winter Conference on Applications of Computer Vision (WACV)},
  pages={1743--1751},
  year={2019},
  organization={IEEE}
}

@article{everingham2010pascal,
  title={The pascal visual object classes (voc) challenge},
  author={Everingham, Mark and Van Gool, Luc and Williams, Christopher KI and Winn, John and Zisserman, Andrew},
  journal={International journal of computer vision},
  volume={88},
  number={2},
  pages={303--338},
  year={2010},
  publisher={Springer}
}

@inproceedings{perazzi2016benchmark,
  title={A benchmark dataset and evaluation methodology for video object segmentation},
  author={Perazzi, Federico and Pont-Tuset, Jordi and McWilliams, Brian and Van Gool, Luc and Gross, Markus and Sorkine-Hornung, Alexander},
  booktitle={Proceedings of the IEEE conference on computer vision and pattern recognition},
  pages={724--732},
  year={2016}
}

@inproceedings{de2021part,
  title={Part-aware panoptic segmentation},
  author={de Geus, Daan and Meletis, Panagiotis and Lu, Chenyang and Wen, Xiaoxiao and Dubbelman, Gijs},
  booktitle={Proceedings of the IEEE/ CVF Conference on Computer Vision and Pattern Recognition},
  pages={5485--5494},
  year={2021}
}

@inproceedings{richter2017playing,
  title={Playing for benchmarks},
  author={Richter, Stephan R and Hayder, Zeeshan and Koltun, Vladlen},
  booktitle={Proceedings of the IEEE International Conference on Computer Vision},
  pages={2213--2222},
  year={2017}
}

@inproceedings{neuhold2017mapillary,
  title={The mapillary vistas dataset for semantic understanding of street scenes},
  author={Neuhold, Gerhard and Ollmann, Tobias and Rota Bulo, Samuel and Kontschieder, Peter},
  booktitle={Proceedings of the IEEE international conference on computer vision},
  pages={4990--4999},
  year={2017}
}

@inproceedings{song2015sun,
  title={Sun rgb-d: A rgb-d scene understanding benchmark suite},
  author={Song, Shuran and Lichtenberg, Samuel P and Xiao, Jianxiong},
  booktitle={Proceedings of the IEEE conference on computer vision and pattern recognition},
  pages={567--576},
  year={2015}
}

@article{hackel2017semantic3d,
author = {Hackel, Timo and Savinov, Nikolay and Ladicky, Lubor and Wegner, Jan and Schindler, Konrad and Pollefeys, Marc},
year = {2017},
%% month = {04},
pages = {},
title = {Semantic3D.net: A new Large-scale Point Cloud Classification Benchmark},
volume = {IV-1/W1},
journal = {ISPRS Annals of Photogrammetry, Remote Sensing and Spatial Information Sciences},
doi = {10.5194/isprs-annals-IV-1-W1-91-2017}
}

@article{brostow2009semantic,
  title={Semantic object classes in video: A high-definition ground truth database},
  author={Brostow, Gabriel J and Fauqueur, Julien and Cipolla, Roberto},
  journal={Pattern Recognition Letters},
  volume={30},
  number={2},
  pages={88--97},
  year={2009},
  publisher={Elsevier}
}

@inproceedings{gong2017look,
  title={Look into person: Self-supervised structure-sensitive learning and a new benchmark for human parsing},
  author={Gong, Ke and Liang, Xiaodan and Zhang, Dongyu and Shen, Xiaohui and Lin, Liang},
  booktitle={Proceedings of the IEEE Conference on Computer Vision and Pattern Recognition},
  pages={932--940},
  year={2017}
}

@inproceedings{caesar2018coco,
  title={Coco-stuff: Thing and stuff classes in context},
  author={Caesar, Holger and Uijlings, Jasper and Ferrari, Vittorio},
  booktitle={Proceedings of the IEEE conference on computer vision and pattern recognition},
  pages={1209--1218},
  year={2018}
}

@inproceedings{silberman2012indoor,
  title={Indoor segmentation and support inference from rgbd images},
  author={Silberman, Nathan and Hoiem, Derek and Kohli, Pushmeet and Fergus, Rob},
  booktitle={European conference on computer vision},
  pages={746--760},
  year={2012},
  organization={Springer}
}

@article{peng2022mass,
  title={MASS: Multi-attentional semantic segmentation of LiDAR data for dense top-view understanding},
  author={Peng, Kunyu and Fei, Juncong and Yang, Kailun and Roitberg, Alina and Zhang, Jiaming and Bieder, Frank and Heidenreich, Philipp and Stiller, Christoph and Stiefelhagen, Rainer},
  journal={IEEE Transactions on Intelligent Transportation Systems},
  year={2022},
  publisher={IEEE}
}

@inproceedings{deng2017cnn,
  title={CNN based semantic segmentation for urban traffic scenes using fisheye camera},
  author={Deng, Liuyuan and Yang, Ming and Qian, Yeqiang and Wang, Chunxiang and Wang, Bing},
  booktitle={2017 IEEE Intelligent Vehicles Symposium (IV)},
  pages={231--236},
  year={2017},
  organization={IEEE}
}

@inproceedings{blott2018semantic,
  title={Semantic segmentation of fisheye images},
  author={Blott, Gregor and Takami, Masato and Heipke, Christian},
  booktitle={Proceedings of the European Conference on Computer Vision (ECCV) Workshops},
  pages={0--0},
  year={2018}
}

@inproceedings{yang2021capturing,
  title={Capturing omni-range context for omnidirectional segmentation},
  author={Yang, Kailun and Zhang, Jiaming and Rei{\ss}, Simon and Hu, Xinxin and Stiefelhagen, Rainer},
  booktitle={Proceedings of the IEEE/CVF Conference on Computer Vision and Pattern Recognition},
  pages={1376--1386},
  year={2021}
}

@article{yang2020omnisupervised,
  title={Omnisupervised omnidirectional semantic segmentation},
  author={Yang, Kailun and Hu, Xinxin and Fang, Yicheng and Wang, Kaiwei and Stiefelhagen, Rainer},
  journal={IEEE Transactions on Intelligent Transportation Systems},
  year={2020},
  publisher={IEEE}
}

@article{zhang2021transfer,
  title={Transfer beyond the Field of View: Dense Panoramic Semantic Segmentation via Unsupervised Domain Adaptation},
  author={Zhang, Jiaming and Ma, Chaoxiang and Yang, Kailun and Roitberg, Alina and Peng, Kunyu and Stiefelhagen, Rainer},
  journal={IEEE Transactions on Intelligent Transportation Systems},
  year={2021},
  publisher={IEEE}
}

@inproceedings{sun2021aerial,
  title={Aerial-PASS: panoramic annular scene segmentation in drone videos},
  author={Sun, Lei and Wang, Jia and Yang, Kailun and Wu, Kaikai and Zhou, Xiangdong and Wang, Kaiwei and Bai, Jian},
  booktitle={2021 European Conference on Mobile Robots (ECMR)},
  pages={1--6},
  year={2021},
  organization={IEEE}
}

@article{wang2022high,
  title={High-performance panoramic annular lens design for real-time semantic segmentation on aerial imagery},
  author={Wang, Jia and Yang, Kailun and Gao, Shaohua and Sun, Lei and Zhu, Chengxi and Wang, Kaiwei and Bai, Jian},
  journal={Optical Engineering},
  volume={61},
  number={3},
  pages={035101},
  year={2022},
  publisher={SPIE}
}

@inproceedings{blin2019road,
  title={Road scenes analysis in adverse weather conditions by polarization-encoded images and adapted deep learning},
  author={Blin, Rachel and Ainouz, Samia and Canu, St{\'e}phane and Meriaudeau, Fabrice},
  booktitle={2019 IEEE Intelligent Transportation Systems Conference (ITSC)},
  pages={27--32},
  year={2019},
  organization={IEEE}
}

@inproceedings{blanchon2019outdoor,
  title={Outdoor Scenes Pixel-wise Semantic Segmentation using Polarimetry and Fully Convolutional Network.},
  author={Blanchon, Marc and Morel, Olivier and Zhang, Yifei and Seulin, Ralph and Crombez, Nathan and Sidib{\'e}, D{\'e}sir{\'e}},
  booktitle={VISIGRAPP (5: VISAPP)},
  pages={328--335},
  year={2019}
}

@inproceedings{yan2021nlfnet,
  title={NLFNet: Non-Local Fusion Towards Generalized Multimodal Semantic Segmentation across RGB-Depth, Polarization, and Thermal Images},
  author={Yan, Ran and Yang, Kailun and Wang, Kaiwei},
  booktitle={2021 IEEE International Conference on Robotics and Biomimetics (ROBIO)},
  pages={1129--1135},
  year={2021},
  organization={IEEE}
}

@article{xiang2021polarization,
  title={Polarization-driven semantic segmentation via efficient attention-bridged fusion},
  author={Xiang, Kaite and Yang, Kailun and Wang, Kaiwei},
  journal={Optics Express},
  volume={29},
  number={4},
  pages={4802--4820},
  year={2021},
  publisher={Optica Publishing Group}
}

@inproceedings{blin2020new,
  title={A new multimodal RGB and polarimetric image dataset for road scenes analysis},
  author={Blin, Rachel and Ainouz, Samia and Canu, St{\'e}phane and Meriaudeau, Fabrice},
  booktitle={Proceedings of the IEEE/CVF Conference on Computer Vision and Pattern Recognition Workshops},
  pages={216--217},
  year={2020}
}

@article{li2022unconventional,
  title={Unconventional Visual Sensors for Autonomous Vehicles},
  author={Li, You and Moreau, Julien and Ibanez-Guzman, Javier},
  journal={arXiv preprint arXiv:2205. 09383},
  year={2022}
}

@inproceedings{jaus2021panoramic,
  title={Panoramic panoptic segmentation: Towards complete surrounding understanding via unsupervised contrastive learning},
  author={Jaus, Alexander and Yang, Kailun and Stiefelhagen, Rainer},
  booktitle={2021 IEEE Intelligent Vehicles Symposium (IV)},
  pages={1421--1427},
  year={2021},
  organization={IEEE}
}

@article{yang2021context,
  title={Is context-aware cnn ready for the surroundings? panoramic semantic segmentation in the wild},
  author={Yang, Kailun and Hu, Xinxin and Stiefelhagen, Rainer},
  journal={IEEE Transactions on Image Processing},
  volume={30},
  pages={1866--1881},
  year={2021},
  publisher={IEEE}
}

@inproceedings{yang2020ds,
  title={Ds-pass: Detail-sensitive panoramic annular semantic segmentation through swaftnet for surrounding sensing},
  author={Yang, Kailun and Hu, Xinxin and Chen, Hao and Xiang, Kaite and Wang, Kaiwei and Stiefelhagen, Rainer},
  booktitle={2020 IEEE Intelligent Vehicles Symposium (IV)},
  pages={457--464},
  year={2020},
  organization={IEEE}
}

@inproceedings{yang2019can,
  title={Can we pass beyond the field of view? panoramic annular semantic segmentation for real-world surrounding perception},
  author={Yang, Kailun and Hu, Xinxin and Bergasa, Luis M and Romera, Eduardo and Huang, Xiao and Sun, Dongming and Wang, Kaiwei},
  booktitle={2019 IEEE Intelligent Vehicles Symposium (IV)},
  pages={446--453},
  year={2019},
  organization={IEEE}
}

@inproceedings{zhang2022bending,
  title={Bending reality: Distortion-aware transformers for adapting to panoramic semantic segmentation},
  author={Zhang, Jiaming and Yang, Kailun and Ma, Chaoxiang and Rei{\ss}, Simon and Peng, Kunyu and Stiefelhagen, Rainer},
  booktitle={Proceedings of the IEEE/CVF Conference on Computer Vision and Pattern Recognition},
  pages={16917--16927},
  year={2022}
}

@inproceedings{strudel2021segmenter,
  title={Segmenter: Transformer for semantic segmentation},
  author={Strudel, Robin and Garcia, Ricardo and Laptev, Ivan and Schmid, Cordelia},
  booktitle={Proceedings of the IEEE/CVF International Conference on Computer Vision},
  pages={7262--7272},
  year={2021}
}

@article{fu2020scene,
  title={Scene segmentation with dual relation-aware attention network},
  author={Fu, Jun and Liu, Jing and Jiang, Jie and Li, Yong and Bao, Yongjun and Lu, Hanqing},
  journal={IEEE Transactions on Neural Networks and Learning Systems},
  volume={32},
  number={6},
  pages={2547--2560},
  year={2020},
  publisher={IEEE}
}

@inproceedings{yin2020disentangled,
  title={Disentangled non-local neural networks},
  author={Yin, Minghao and Yao, Zhuliang and Cao, Yue and Li, Xiu and Zhang, Zheng and Lin, Stephen and Hu, Han},
  booktitle={European Conference on Computer Vision},
  pages={191--207},
  year={2020},
  organization={Springer}
}

@inproceedings{dufour2020instance,
  title={Instance segmentation in fisheye images},
  author={Dufour, R{\'e}mi and Meurie, Cyril and Strauss, Cl{\'e}ment and Lezoray, Olivier},
  booktitle={2020 Tenth International Conference on Image Processing Theory, Tools and Applications (IPTA)},
  pages={1--6},
  year={2020},
  organization={IEEE}
}

@inproceedings{ye2020universal,
  title={Universal semantic segmentation for fisheye urban driving images},
  author={Ye, Yaozu and Yang, Kailun and Xiang, Kaite and Wang, Juan and Wang, Kaiwei},
  booktitle={2020 IEEE International Conference on Systems, Man, and Cybernetics (SMC)},
  pages={648--655},
  year={2020},
  organization={IEEE}
}

@article{deng2019restricted,
  title={Restricted deformable convolution-based road scene semantic segmentation using surround view cameras},
  author={Deng, Liuyuan and Yang, Ming and Li, Hao and Li, Tianyi and Hu, Bing and Wang, Chunxiang},
  journal={IEEE Transactions on Intelligent Transportation Systems},
  volume={21},
  number={10},
  pages={4350--4362},
  year={2019},
  publisher={IEEE}
}

@inproceedings{cordts2016cityscapes,
  title={The cityscapes dataset for semantic urban scene understanding},
  author={Cordts, Marius and Omran, Mohamed and Ramos, Sebastian and Rehfeld, Timo and Enzweiler, Markus and Benenson, Rodrigo and Franke, Uwe and Roth, Stefan and Schiele, Bernt},
  booktitle={Proceedings of the IEEE conference on computer vision and pattern recognition},
  pages={3213--3223},
  year={2016}
}

@inproceedings{lin2014microsoft,
  title={Microsoft coco: Common objects in context},
  author={Lin, Tsung-Yi and Maire, Michael and Belongie, Serge and Hays, James and Perona, Pietro and Ramanan, Deva and Doll{\'a}r, Piotr and Zitnick, C Lawrence},
  booktitle={European conference on computer vision},
  pages={740--755},
  year={2014},
  organization={Springer}
}

@inproceedings{zhou2017scene,
  title={Scene parsing through ade20k dataset},
  author={Zhou, Bolei and Zhao, Hang and Puig, Xavier and Fidler, Sanja and Barriuso, Adela and Torralba, Antonio},
  booktitle={Proceedings of the IEEE conference on computer vision and pattern recognition},
  pages={633--641},
  year={2017}
}

@inproceedings{mottaghi2014role,
  title={The role of context for object detection and semantic segmentation in the wild},
  author={Mottaghi, Roozbeh and Chen, Xianjie and Liu, Xiaobai and Cho, Nam-Gyu and Lee, Seong-Whan and Fidler, Sanja and Urtasun, Raquel and Yuille, Alan},
  booktitle={Proceedings of the IEEE conference on computer vision and pattern recognition},
  pages={891--898},
  year={2014}
}

@inproceedings{geiger2012we,
  title={Are we ready for autonomous driving? the kitti vision benchmark suite},
  author={Geiger, Andreas and Lenz, Philip and Urtasun, Raquel},
  booktitle={2012 IEEE conference on computer vision and pattern recognition},
  pages={3354--3361},
  year={2012},
  organization={IEEE}
}

@inproceedings{alvarez2012road,
  title={Road scene segmentation from a single image},
  author={Alvarez, Jose M and Gevers, Theo and LeCun, Yann and Lopez, Antonio M},
  booktitle={European Conference on Computer Vision},
  pages={376--389},
  year={2012},
  organization={Springer}
}

@inproceedings{zhang2015sensor,
  title={Sensor fusion for semantic segmentation of urban scenes},
  author={Zhang, Richard and Candra, Stefan A and Vetter, Kai and Zakhor, Avideh},
  booktitle={2015 IEEE international conference on robotics and automation (ICRA)},
  pages={1850--1857},
  year={2015},
  organization={IEEE}
}

@inproceedings{ros2015vision,
  title={Vision-based offline-online perception paradigm for autonomous driving},
  author={Ros, German and Ramos, Sebastian and Granados, Manuel and Bakhtiary, Amir and Vazquez, David and Lopez, Antonio M},
  booktitle={2015 IEEE Winter Conference on Applications of Computer Vision},
  pages={231--238},
  year={2015},
  organization={IEEE}
}

@inproceedings{gould2009decomposing,
  title={Decomposing a scene into geometric and semantically consistent regions},
  author={Gould, Stephen and Fulton, Richard and Koller, Daphne},
  booktitle={2009 IEEE 12th international conference on computer vision},
  pages={1--8},
  year={2009},
  organization={IEEE}
}

@article{liu2011nonparametric,
  title={Nonparametric scene parsing via label transfer},
  author={Liu, Ce and Yuen, Jenny and Torralba, Antonio},
  journal={IEEE Transactions on Pattern Analysis and Machine Intelligence},
  volume={33},
  number={12},
  pages={2368--2382},
  year={2011},
  publisher={IEEE}
}

@inproceedings{wang2015semantic,
  title={Semantic part segmentation using compositional model combining shape and appearance},
  author={Wang, Jianyu and Yuille, Alan L},
  booktitle={Proceedings of the IEEE conference on computer vision and pattern recognition},
  pages={1788--1797},
  year={2015}
}

@article{liang2015deep,
  title={Deep human parsing with active template regression},
  author={Liang, Xiaodan and Liu, Si and Shen, Xiaohui and Yang, Jianchao and Liu, Luoqi and Dong, Jian and Lin, Liang and Yan, Shuicheng},
  journal={IEEE transactions on pattern analysis and machine intelligence},
  volume={37},
  number={12},
  pages={2402--2414},
  year={2015},
  publisher={IEEE}
}

@inproceedings{yamaguchi2012parsing,
  title={Parsing clothing in fashion photographs},
  author={Yamaguchi, Kota and Kiapour, M Hadi and Ortiz, Luis E and Berg, Tamara L},
  booktitle={2012 IEEE Conference on Computer vision and pattern recognition},
  pages={3570--3577},
  year={2012},
  organization={IEEE}
}

@article{de2018panoptic,
  title={Panoptic segmentation with a joint semantic and instance segmentation network},
  author={De Geus, Daan and Meletis, Panagiotis and Dubbelman, Gijs},
  journal={arXiv preprint arXiv:1809.02110},
  year={2018}
}

@inproceedings{gupta2019lvis,
  title={LVIS: A dataset for large vocabulary instance segmentation},
  author={Gupta, Agrim and Dollar, Piotr and Girshick, Ross},
  booktitle={Proceedings of the IEEE/CVF Conference on Computer Vision and Pattern Recognition},
  pages={5356--5364},
  year={2019}
}

@inproceedings{mo2019partnet,
  title={Partnet: A large-scale benchmark for fine-grained and hierarchical part-level 3d object understanding},
  author={Mo, Kaichun and Zhu, Shilin and Chang, Angel X and Yi, Li and Tripathi, Subarna and Guibas, Leonidas J and Su, Hao},
  booktitle={Proceedings of the IEEE/CVF Conference on Computer Vision and Pattern Recognition},
  pages={909--918},
  year={2019}
}

@article{staal2004ridge,
  title={Ridge-based vessel segmentation in color images of the retina},
  author={Staal, Joes and Abr{\`a}moff, Michael D and Niemeijer, Meindert and Viergever, Max A and Van Ginneken, Bram},
  journal={IEEE transactions on medical imaging},
  volume={23},
  number={4},
  pages={501--509},
  year={2004},
  publisher={IEEE}
}

@article{litjens2014evaluation,
  title={Evaluation of prostate segmentation algorithms for MRI: the PROMISE12 challenge},
  author={Litjens, Geert and Toth, Robert and van de Ven, Wendy and Hoeks, Caroline and Kerkstra, Sjoerd and van Ginneken, Bram and Vincent, Graham and Guillard, Gwenael and Birbeck, Neil and Zhang, Jindang and others},
  journal={Medical image analysis},
  volume={18},
  number={2},
  pages={359--373},
  year={2014},
  publisher={Elsevier}
}

@article{menze2014multimodal,
  title={The multimodal brain tumor image segmentation benchmark (BRATS)},
  author={Menze, Bjoern H and Jakab, Andras and Bauer, Stefan and Kalpathy-Cramer, Jayashree and Farahani, Keyvan and Kirby, Justin and Burren, Yuliya and Porz, Nicole and Slotboom, Johannes and Wiest, Roland and others},
  journal={IEEE transactions on medical imaging},
  volume={34},
  number={10},
  pages={1993--2024},
  year={2014},
  publisher={IEEE}
}

@inproceedings{behley2019semantickitti,
  title={Semantickitti: A dataset for semantic scene understanding of lidar sequences},
  author={Behley, Jens and Garbade, Martin and Milioto, Andres and Quenzel, Jan and Behnke, Sven and Stachniss, Cyrill and Gall, Jurgen},
  booktitle={Proceedings of the IEEE/CVF International Conference on Computer Vision},
  pages={9297--9307},
  year={2019}
}

@inproceedings{qi2019amodal,
  title={Amodal instance segmentation with kins dataset},
  author={Qi, Lu and Jiang, Li and Liu, Shu and Shen, Xiaoyong and Jia, Jiaya},
  booktitle={Proceedings of the IEEE/CVF Conference on Computer Vision and Pattern Recognition},
  pages={3014--3023},
  year={2019}
}

@article{xu2018youtube,
  title={Youtube-vos: A large-scale video object segmentation benchmark},
  author={Xu, Ning and Yang, Linjie and Fan, Yuchen and Yue, Dingcheng and Liang, Yuchen and Yang, Jianchao and Huang, Thomas},
  journal={arXiv preprint arXiv:1809.03327},
  year={2018}
}

@inproceedings{caesar2020nuscenes,
  title={nuscenes: A multimodal dataset for autonomous driving},
  author={Caesar, Holger and Bankiti, Varun and Lang, Alex H and Vora, Sourabh and Liong, Venice Erin and Xu, Qiang and Krishnan, Anush and Pan, Yu and Baldan, Giancarlo and Beijbom, Oscar},
  booktitle={Proceedings of the IEEE/CVF conference on computer vision and pattern recognition},
  pages={11621--11631},
  year={2020}
}

@article{chang2015shapenet,
  title={Shapenet: An information-rich 3d model repository},
  author={Chang, Angel X and Funkhouser, Thomas and Guibas, Leonidas and Hanrahan, Pat and Huang, Qixing and Li, Zimo and Savarese, Silvio and Savva, Manolis and Song, Shuran and Su, Hao and others},
  journal={arXiv preprint arXiv:1512. 03012},
  year={2015}
}

@inproceedings{armeni20163d,
  title={3d semantic parsing of large-scale indoor spaces},
  author={Armeni, Iro and Sener, Ozan and Zamir, Amir R and Jiang, Helen and Brilakis, Ioannis and Fischer, Martin and Savarese, Silvio},
  booktitle={Proceedings of the IEEE Conference on Computer Vision and Pattern Recognition},
  pages={1534--1543},
  year={2016}
}

@inproceedings{song2017semantic,
  title={Semantic scene completion from a single depth image},
  author={Song, Shuran and Yu, Fisher and Zeng, Andy and Chang, Angel X and Savva, Manolis and Funkhouser, Thomas},
  booktitle={Proceedings of the IEEE Conference on Computer Vision and Pattern Recognition},
  pages={1746--1754},
  year={2017}
}

@inproceedings{dai2017scannet,
  title={Scannet: Richly-annotated 3d reconstructions of indoor scenes},
  author={Dai, Angela and Chang, Angel X and Savva, Manolis and Halber, Maciej and Funkhouser, Thomas and Nie{\ss}ner, Matthias},
  booktitle={Proceedings of the IEEE conference on computer vision and pattern recognition},
  pages={5828--5839},
  year={2017}
}

@inproceedings{chen2014detect,
  title={Detect what you can: Detecting and representing objects using holistic models and body parts},
  author={Chen, Xianjie and Mottaghi, Roozbeh and Liu, Xiaobai and Fidler, Sanja and Urtasun, Raquel and Yuille, Alan},
  booktitle={Proceedings of the IEEE conference on computer vision and pattern recognition},
  pages={1971--1978},
  year={2014}
}

@article{otsu1979threshold,
  title={A threshold selection method from gray-level histograms},
  author={Otsu, Nobuyuki},
  journal={IEEE transactions on systems, man, and cybernetics},
  volume={9},
  number={1},
  pages={62--66},
  year={1979},
  publisher={IEEE}
}

@article{adams1994seeded,
  title={Seeded region growing},
  author={Adams, Rolf and Bischof, Leanne},
  journal={IEEE Transactions on pattern analysis and machine intelligence},
  volume={16},
  number={6},
  pages={641--647},
  year={1994},
  publisher={IEEE}
}

@inproceedings{qin2010scene,
  title={Scene segmentation based on seeded region growing for foreground detection},
  author={Qin, Hongwu and Zain, Jasni Mohamad and Ma, Xiuqin and Hai, Tao},
  booktitle={2010 Sixth International Conference on Natural Computation},
  volume={7},
  pages={3619--3623},
  year={2010},
  organization={IEEE}
}

@article{ramli2020homogeneous,
  title={Homogeneous tree height derivation from tree crown delineation using Seeded Region Growing (SRG) segmentation},
  author={Ramli, Muhamad Farid and Tahar, Khairul Nizam},
  journal={Geo-spatial Information Science},
  volume={23},
  number={3},
  pages={195--208},
  year={2020},
  publisher={Taylor \& Francis}
}

@inproceedings{rong2014improved,
  title={An improved CANNY edge detection algorithm},
  author={Rong, Weibin and Li, Zhanjing and Zhang, Wei and Sun, Lining},
  booktitle={2014 IEEE international conference on mechatronics and automation},
  pages={577--582},
  year={2014},
  organization={IEEE}
}

@article{canny1986computational,
  title={A computational approach to edge detection},
  author={Canny, John},
  journal={IEEE Transactions on pattern analysis and machine intelligence},
  number={6},
  pages={679--698},
  year={1986},
  publisher={Ieee}
}

@article{mumford1989optimal,
  title={Optimal approximations by piecewise smooth functions and associated variational problems},
  author={Mumford, David Bryant and Shah, Jayant},
  journal={Communications on pure and applied mathematics},
  year={1989},
  publisher={Wiley-Blackwell}
}

@article{osher1988fronts,
  title={Fronts propagating with curvature-dependent speed: Algorithms based on Hamilton-Jacobi formulations},
  author={Osher, Stanley and Sethian, James A},
  journal={Journal of computational physics},
  volume={79},
  number={1},
  pages={12--49},
  year={1988},
  publisher={Elsevier}
}

@article{malladi1995shape,
  title={Shape modeling with front propagation: A level set approach},
  author={Malladi, Ravi and Sethian, James A and Vemuri, Baba C},
  journal={IEEE transactions on pattern analysis and machine intelligence},
  volume={17},
  number={2},
  pages={158--175},
  year={1995},
  publisher={IEEE}
}

@inproceedings{luo2019convex,
  title={Convex shape prior for multi-object segmentation using a single level set function},
  author={Luo, Shousheng and Tai, Xue-Cheng and Huo, Limei and Wang, Yang and Glowinski, Roland},
  booktitle={Proceedings of the IEEE/CVF International Conference on Computer Vision},
  pages={613--621},
  year={2019}
}

@article{yan2020convexity,
  title={Convexity shape prior for level set-based image segmentation method},
  author={Yan, Shi and Tai, Xue-Cheng and Liu, Jun and Huang, Hai-Yang},
  journal={IEEE Transactions on Image Processing},
  volume={29},
  pages={7141--7152},
  year={2020},
  publisher={IEEE}
}

@inproceedings{yang2017level,
  title={A level set method for convexity preserving segmentation of cardiac left ventricle},
  author={Yang, Cong and Shi, Xue and Yao, Donglan and Li, Chunming},
  booktitle={2017 IEEE International Conference on Image Processing (ICIP)},
  pages={2159--2163},
  year={2017},
  organization={IEEE}
}

@article{shi2000normalized,
  title={Normalized cuts and image segmentation},
  author={Shi, Jianbo and Malik, Jitendra},
  journal={IEEE Transactions on pattern analysis and machine intelligence},
  volume={22},
  number={8},
  pages={888--905},
  year={2000},
  publisher={Ieee}
}

@inproceedings{moore2008superpixel,
  title={Superpixel lattices},
  author={Moore, Alastair P and Prince, Simon JD and Warrell, Jonathan and Mohammed, Umar and Jones, Graham},
  booktitle={2008 IEEE conference on computer vision and pattern recognition},
  pages={1--8},
  year={2008},
  organization={IEEE}
}

@inproceedings{liu2011entropy,
  title={Entropy rate superpixel segmentation},
  author={Liu, Ming-Yu and Tuzel, Oncel and Ramalingam, Srikumar and Chellappa, Rama},
  booktitle={CVPR 2011},
  pages={2097--2104},
  year={2011},
  organization={IEEE}
}

@article{d2002robust,
  title={A Robust Approach toward Feature Space Analysis},
  author={d Comaniciu, Meer and Shift, P Mean},
  journal={IEEE Trans. Patt. An. Mach. Intell},
  volume={24},
  year={2002}
}

@book{szeliski2010computer,
  title={Computer vision: algorithms and applications},
  author={Szeliski, Richard},
  year={2010},
  publisher={Spring\-er Science \& Business Media}
}

@article{kass1988snakes,
  title={Snakes: Active contour models},
  author={Kass, Michael and Witkin, Andrew and Terzopoulos, Demetri},
  journal={International journal of computer vision},
  volume={1},
  number={4},
  pages={321--331},
  year={1988},
  publisher={Springer}
}

@inproceedings{chen2019learning,
  title={Learning active contour models for medical image segmentation},
  author={Chen, Xu and Williams, Bryan M and Vallabhaneni, Srinivasa R and Czanner, Gabriela and Williams, Rachel and Zheng, Yalin},
  booktitle={Proceedings of the IEEE/CVF Conference on Computer Vision and Pattern Recognition},
  pages={11632--11640},
  year={2019}
}

@inproceedings{huang2017densely,
  title={Densely connected convolutional networks},
  author={Huang, Gao and Liu, Zhuang and Van Der Maaten, Laurens and Weinberger, Kilian Q},
  booktitle={Proceedings of the IEEE conference on computer vision and pattern recognition},
  pages={4700--4708},
  year={2017}
}

@inproceedings{gur2019unsupervised,
  title={Unsupervised microvascular image segmentation using an active contours mimicking neural network},
  author={Gur, Shir and Wolf, Lior and Golgher, Lior and Blinder, Pablo},
  booktitle={Proceedings of the IEEE/CVF International Conference on Computer Vision},
  pages={10722--10731},
  year={2019}
}

@article{chan2001active,
  title={Active contours without edges},
  author={Chan, Tony F and Vese, Luminita A},
  journal={IEEE Transactions on image processing},
  volume={10},
  number={2},
  pages={266--277},
  year={2001},
  publisher={IEEE}
}

@article{haft2019deep,
  title={Deep convolutional neural networks for segmenting 3D in vivo multiphoton images of vasculature in Alzheimer disease mouse models},
  author={Haft-Javaherian, Mohammad and Fang, Linjing and Muse, Victorine and Schaffer, Chris B and Nishimura, Nozomi and Sabuncu, Mert R},
  journal={PloS one},
  volume={14},
  number={3},
  pages={e0213539},
  year={2019},
  publisher={Public Library of Science San Francisco, CA USA}
}

@article{teikari2016deep,
  title={Deep learning convolutional networks for multiphoton microscopy vasculature segmentation},
  author={Teikari, Petteri and Santos, Marc and Poon, Charissa and Hynynen, Kullervo},
  journal={arXiv preprint arXiv:1606.02382},
  year={2016}
}

@article{le2018reformulating,
  title={Reformulating level sets as deep recurrent neural network approach to semantic segmentation},
  author={Le, T Hoang Ngan and Quach, Kha Gia and Luu, Khoa and Duong, Chi Nhan and Savvides, Marios},
  journal={IEEE Transactions on Image Processing},
  volume={27},
  number={5},
  pages={2393--2407},
  year={2018},
  publisher={IEEE}
}

@article{rupprecht2016deep,
  title={Deep active contours},
  author={Rupprecht, Christian and Huaroc, Elizabeth and Baust, Maximilian and Navab, Nassir},
  journal={arXiv preprint arXiv:1607.05074},
  year={2016}
}

@inproceedings{hatamizadeh2020end,
  title={End-to-end trainable deep active contour models for automated image segmentation: Delineating buildings in aerial imagery},
  author={Hatamizadeh, Ali and Sengupta, Debleena and Terzopoulos, Demetri},
  booktitle={European Conference on Computer Vision},
  pages={730--746},
  year={2020},
  organization={Springer}
}

@article{gerke2014use,
  title={Use of the stair vision library within the ISPRS 2D semantic labeling benchmark (Vaihingen)},
  author={Gerke, Markus},
  year={2014}
}

@inproceedings{marcos2018learning,
  title={Learning deep structured active contours end-to-end},
  author={Marcos, Diego and Tuia, Devis and Kellenberger, Benjamin and Zhang, Lisa and Bai, Min and Liao, Renjie and Urtasun, Raquel},
  booktitle={Proceedings of the IEEE Conference on Computer Vision and Pattern Recognition},
  pages={8877--8885},
  year={2018}
}

@inproceedings{hu2017deep,
  title={Deep level sets for salient object detection},
  author={Hu, Ping and Shuai, Bing and Liu, Jun and Wang, Gang},
  booktitle={Proceedings of the IEEE conference on computer vision and pattern recognition},
  pages={2300--2309},
  year={2017}
}

@inproceedings{hatamizadeh2019deep,
  title={Deep active lesion segmentation},
  author={Hatamizadeh, Ali and Hoogi, Assaf and Sengupta, Debleena and Lu, Wuyue and Wilcox, Brian and Rubin, Daniel and Terzopoulos, Demetri},
  booktitle={International Workshop on Machine Learning in Medical Imaging},
  pages={98--105},
  year={2019},
  organization={Springer}
}

@article{hatamizadeh2019end,
  title={End-to-end deep convolutional active contours for image segmentation},
  author={Hatamizadeh, Ali and Sengupta, Debleena and Terzopoulos, Demetri},
  journal={arXiv preprint arXiv:1909.13359},
  year={2019}
}

@inproceedings{cheng2019darnet,
  title={Darnet: Deep active ray network for building segmentation},
  author={Cheng, Dominic and Liao, Renjie and Fidler, Sanja and Urtasun, Raquel},
  booktitle={Proceedings of the IEEE/CVF Conference on Computer Vision and Pattern Recognition},
  pages={7431--7439},
  year={2019}
}

@inproceedings{wang2019object,
  title={Object instance annotation with deep extreme level set evolution},
  author={Wang, Zian and Acuna, David and Ling, Huan and Kar, Amlan and Fidler, Sanja},
  booktitle={Proceedings of the IEEE/CVF Conference on Computer Vision and Pattern Recognition},
  pages={7500--7508},
  year={2019}
}

@article{hoogi2016adaptive,
  title={Adaptive estimation of active contour parameters using convolutional neural networks and texture analysis},
  author={Hoogi, Assaf and Subramaniam, Arjun and Veerapaneni, Rishi and Rubin, Daniel L},
  journal={IEEE transactions on medical imaging},
  volume={36},
  number={3},
  pages={781--791},
  year={2016},
  publisher={IEEE}
}

@article{akbari2021deep,
author = {Akbarimoghaddam, Parastoo and Ziaei, Atefeh and Azarnoush, Hamed},
year = {2022},
%% month = {01},
pages = {},
title = {Deep active contours using locally controlled distance vector flow},
journal = {Signal, Image and Video Processing},
doi = {10.1007/s11760-022-02134-1}
}

@inproceedings{peng2020deep,
  title={Deep snake for real-time instance segmentation},
  author={Peng, Sida and Jiang, Wen and Pi, Huaijin and Li, Xiuli and Bao, Hujun and Zhou, Xiaowei},
  booktitle={Proceedings of the IEEE/CVF Conference on Computer Vision and Pattern Recognition},
  pages={8533--8542},
  year={2020}
}

@article{sun2020real,
  title={Real-time fusion network for RGB-D semantic segmentation incorporating unexpected obstacle detection for road-driving images},
  author={Sun, Lei and Yang, Kailun and Hu, Xinxin and Hu, Weijian and Wang, Kaiwei},
  journal={IEEE Robotics and Automation Letters},
  volume={5},
  number={4},
  pages={5558--5565},
  year={2020},
  publisher={IEEE}
}

@inproceedings{chen2020bi,
  title={Bi-directional cross-modality feature propagation with separation-and-aggregation gate for RGB-D semantic segmentation},
  author={Chen, Xiaokang and Lin, Kwan-Yee and Wang, Jingbo and Wu, Wayne and Qian, Chen and Li, Hongsheng and Zeng, Gang},
  booktitle={European Conference on Computer Vision},
  pages={561--577},
  year={2020},
  organization={Springer}
}

@inproceedings{zhang2019pattern,
  title={Pattern-affinitive propagation across depth, surface normal and semantic segmentation},
  author={Zhang, Zhenyu and Cui, Zhen and Xu, Chunyan and Yan, Yan and Sebe, Nicu and Yang, Jian},
  booktitle={Proceedings of the IEEE/CVF Conference on Computer Vision and Pattern Recognition},
  pages={4106--4115},
  year={2019}
}

@inproceedings{cao2021shapeconv,
  title={ShapeConv: Shape-aware Convolutional Layer for Indoor RGB-D Semantic Segmentation},
  author={Cao, Jinming and Leng, Hanchao and Lischinski, Dani and Cohen-Or, Daniel and Tu, Changhe and Li, Yangyan},
  booktitle={Proceedings of the IEEE/CVF International Conference on Computer Vision},
  pages={7088--7097},
  year={2021}
}

@article{liu2022cmx,
  title={CMX: Cross-Modal Fusion for RGB-X Semantic Segmentation with Transformers},
  author={Liu, Huayao and Zhang, Jiaming and Yang, Kailun and Hu, Xinxin and Stiefelhagen, Rainer},
  journal={arXiv preprint arXiv:2203.04838},
  year={2022}
}

@inproceedings{cheng2022masked,
  title={Masked-attention mask transformer for universal image segmentation},
  author={Cheng, Bowen and Misra, Ishan and Schwing, Alexander G and Kirillov, Alexander and Girdhar, Rohit},
  booktitle={Proceedings of the IEEE/CVF Conference on Computer Vision and Pattern Recognition},
  pages={1290--1299},
  year={2022}
}

@article{zhang2021non,
  title={Non-local aggregation for RGB-D semantic segmentation},
  author={Zhang, Guodong and Xue, Jing-Hao and Xie, Pengwei and Yang, Sifan and Wang, Guijin},
  journal={IEEE Signal Processing Letters},
  volume={28},
  pages={658--662},
  year={2021},
  publisher={IEEE}
}

@inproceedings{park2017rdfnet,
  title={Rdfnet: Rgb-d multi-level residual feature fusion for indoor semantic segmentation},
  author={Park, Seong-Jin and Hong, Ki-Sang and Lee, Seungyong},
  booktitle={Proceedings of the IEEE international conference on computer vision},
  pages={4980--4989},
  year={2017}
}

@inproceedings{dai2018dark,
  title={Dark model adaptation: Semantic image segmentation from daytime to nighttime},
  author={Dai, Dengxin and Van Gool, Luc},
  booktitle={2018 21st International Conference on Intelligent Transportation Systems (ITSC)},
  pages={3819--3824},
  year={2018},
  organization={IEEE}
}

@article{sakaridis2020map,
author = {Sakaridis, Christos and Dai, Dengxin and Gool, Luc},
year = {2020},
%% month = {12},
pages = {1-1},
title = {Map-Guided Curriculum Domain Adaptation and Uncertainty-Aware Evaluation for Semantic Nighttime Image Segmentation},
volume = {PP},
journal = {IEEE Transactions on Pattern Analysis and Machine Intelligence},
doi = {10.1109/TPAMI.2020.3045882}
}

@inproceedings{sakaridis2019guided,
  title={Guided curriculum model adaptation and uncertainty-aware evaluation for semantic nighttime image segmentation},
  author={Sakaridis, Christos and Dai, Dengxin and Gool, Luc Van},
  booktitle={Proceedings of the IEEE/CVF International Conference on Computer Vision},
  pages={7374--7383},
  year={2019}
}

@inproceedings{romera2019bridging,
  title={Bridging the day and night domain gap for semantic segmentation},
  author={Romera, Eduardo and Bergasa, Luis M and Yang, Kailun and Alvarez, Jose M and Barea, Rafael},
  booktitle={2019 IEEE Intelligent Vehicles Symposium (IV)},
  pages={1312--1318},
  year={2019},
  organization={IEEE}
}

@inproceedings{nag2019s,
  title={What’s there in the dark},
  author={Nag, Sauradip and Adak, Saptakatha and Das, Sukhendu},
  booktitle={2019 IEEE International Conference on Image Processing (ICIP)},
  pages={2996--3000},
  year={2019},
  organization={IEEE}
}

@inproceedings{wu2021dannet,
  title={Dannet: A one-stage domain adaptation network for unsupervised nighttime semantic segmentation},
  author={Wu, Xinyi and Wu, Zhenyao and Guo, Hao and Ju, Lili and Wang, Song},
  booktitle={Proceedings of the IEEE/CVF Conference on Computer Vision and Pattern Recognition},
  pages={15769--15778},
  year={2021}
}

@inproceedings{lengyel2021zero,
  title={Zero-Shot Day-Night Domain Adaptation with a Physics Prior},
  author={Lengyel, Attila and Garg, Sourav and Milford, Michael and van Gemert, Jan C},
  booktitle={Proceedings of the IEEE/CVF International Conference on Computer Vision},
  pages={4399--4409},
  year={2021}
}

@inproceedings{sun2019see,
  title={See clearer at night: towards robust nighttime semantic segmentation through day-night image conversion},
  author={Sun, Lei and Wang, Kaiwei and Yang, Kailun and Xiang, Kaite},
  booktitle={Artificial Intelligence and Machine Learning in Defense Applications},
  volume={11169},
  pages={111690A},
  year={2019},
  organization={International Society for Optics and Photonics}
}

@article{zhang2021exploring,
  title={Exploring Event-Driven Dynamic Context for Accident Scene Segmentation},
  author={Zhang, Jiaming and Yang, Kailun and Stiefelhagen, Rainer},
  journal={IEEE Transactions on Intelligent Transportation Systems},
  year={2021},
  publisher={IEEE}
}

@inproceedings{luo2022towards,
  title={Towards Robust Semantic Segmentation of Accident Scenes via Multi-Source Mixed Sampling and Meta-Learning},
  author={Luo, Xinyu and Zhang, Jiaming and Yang, Kailun and Roitberg, Alina and Peng, Kunyu and Stiefelhagen, Rainer},
  booktitle={Proceedings of the IEEE/CVF Conference on Computer Vision and Pattern Recognition},
  pages={4429--4439},
  year={2022}
}

@article{sakaridis2018semantic,
  title={Semantic foggy scene understanding with synthetic data},
  author={Sakaridis, Christos and Dai, Dengxin and Van Gool, Luc},
  journal={International Journal of Computer Vision},
  volume={126},
  number={9},
  pages={973--992},
  year={2018},
  publisher={Springer}
}

@article{dai2020curriculum,
  title={Curriculum model adaptation with synthetic and real data for semantic foggy scene understanding},
  author={Dai, Dengxin and Sakaridis, Christos and Hecker, Simon and Van Gool, Luc},
  journal={International Journal of Computer Vision},
  volume={128},
  number={5},
  pages={1182--1204},
  year={2020},
  publisher={Springer}
}

@inproceedings{sakaridis2018model,
  title={Model adaptation with synthetic and real data for semantic dense foggy scene understanding},
  author={Sakaridis, Christos and Dai, Dengxin and Hecker, Simon and Van Gool, Luc},
  booktitle={Proceedings of the european conference on computer vision (ECCV)},
  pages={687--704},
  year={2018}
}

@inproceedings{hahner2019semantic,
  title={Semantic understanding of foggy scenes with purely synthetic data},
  author={Hahner, Martin and Dai, Dengxin and Sakaridis, Christos and Zaech, Jan-Nico and Van Gool, Luc},
  booktitle={2019 IEEE Intelligent Transportation Systems Conference (ITSC)},
  pages={3675--3681},
  year={2019},
  organization={IEEE}
}

@article{chen2021spatial,
  title={Spatial information guided convolution for real-time RGBD semantic segmentation},
  author={Chen, Lin-Zhuo and Lin, Zheng and Wang, Ziqin and Yang, Yong-Liang and Cheng, Ming-Ming},
  journal={IEEE Transactions on Image Processing},
  volume={30},
  pages={2313--2324},
  year={2021},
  publisher={IEEE}
}

@article{arbelaez2010contour,
  title={Contour detection and hierarchical image segmentation},
  author={Arbelaez, Pablo and Maire, Michael and Fowlkes, Charless and Malik, Jitendra},
  journal={IEEE transactions on pattern analysis and machine intelligence},
  volume={33},
  number={5},
  pages={898--916},
  year={2010},
  publisher={IEEE}
}

@inproceedings{long2015fully,
  title={Fully convolutional networks for semantic segmentation},
  author={Long, Jonathan and Shelhamer, Evan and Darrell, Trevor},
  booktitle={Proceedings of the IEEE conference on computer vision and pattern recognition},
  pages={3431--3440},
  year={2015}
}

@inproceedings{noh2015learning,
  title={Learning deconvolution network for semantic segmentation},
  author={Noh, Hyeonwoo and Hong, Seunghoon and Han, Bohyung},
  booktitle={Proceedings of the IEEE international conference on computer vision},
  pages={1520--1528},
  year={2015}
}

@misc{quan2016fusionnet,
  title={Fusionnet: A deep fully residual convolutional neural network for image segmentation in connectomics},
  author={Quan, Tran Minh and Hildebrand, David GC and Jeong, Won-Ki},
  year={2016}
}

@article{li2018h,
  title={H-DenseUNet: hybrid densely connected UNet for liver and tumor segmentation from CT volumes},
  author={Li, Xiaomeng and Chen, Hao and Qi, Xiaojuan and Dou, Qi and Fu, Chi-Wing and Heng, Pheng-Ann},
  journal={IEEE transactions on medical imaging},
  volume={37},
  number={12},
  pages={2663--2674},
  year={2018},
  publisher={IEEE}
}

@article{shah2018stacked,
  title={Stacked u-nets: a no-frills approach to natural image segmentation},
  author={Shah, Sohil and Ghosh, Pallabi and Davis, Larry S and Goldstein, Tom},
  journal={arXiv preprint arXiv:1804.10343},
  year={2018}
}

@inproceedings{hariharan2014simultaneous,
  title={Simultaneous detection and segmentation},
  author={Hariharan, Bharath and Arbel{\'a}ez, Pablo and Girshick, Ross and Malik, Jitendra},
  booktitle={European conference on computer vision},
  pages={297--312},
  year={2014},
  organization={Springer}
}

@inproceedings{ronneberger2015u,
  title={U-net: Convolutional networks for biomedical image segmentation},
  author={Ronneberger, Olaf and Fischer, Philipp and Brox, Thomas},
  booktitle={International Conference on Medical image computing and computer-assisted intervention},
  pages={234--241},
  year={2015},
  organization={Springer}
}

@article{badrinarayanan2017segnet,
  title={Segnet: A deep convolutional encoder-decoder architecture for image segmentation},
  author={Badrinarayanan, Vijay and Kendall, Alex and Cipolla, Roberto},
  journal={IEEE transactions on pattern analysis and machine intelligence},
  volume={39},
  number={12},
  pages={2481--2495},
  year={2017},
  publisher={IEEE}
}

@article{kendall2015bayesian,
author = {Kendall, Alex and Badrinarayanan, Vijay and Cipolla, Roberto},
year = {2017},
%% month = {01},
pages = {},
title = {Bayesian SegNet: Model Uncertainty in Deep Convolutional Encoder-Decoder Architectures for Scene Understanding},
doi = {10.5244/C.31.57}
}

@article{fu2019stacked,
  title={Stacked deconvolutional network for semantic segmentation},
  author={Fu, Jun and Liu, Jing and Wang, Yuhang and Zhou, Jin and Wang, Changyong and Lu, Hanqing},
  journal={IEEE Transactions on Image Processing},
  year={2019},
  publisher={IEEE}
}

@inproceedings{cheng2017locality,
  title={Locality-sensitive deconvolution networks with gated fusion for rgb-d indoor semantic segmentation},
  author={Cheng, Yanhua and Cai, Rui and Li, Zhiwei and Zhao, Xin and Huang, Kaiqi},
  booktitle={Proceedings of the IEEE conference on computer vision and pattern recognition},
  pages={3029--3037},
  year={2017}
}

@article{mohan2021efficientps,
  title={Efficientps: Efficient panoptic segmentation},
  author={Mohan, Rohit and Valada, Abhinav},
  journal={International Journal of Computer Vision},
  volume={129},
  number={5},
  pages={1551--1579},
  year={2021},
  publisher={Springer}
}

@InProceedings{Zhou_2021_CVPR,
    author    = {Zhou, Zixiang and Zhang, Yang and Foroosh, Hassan},
    title     = {Panoptic-PolarNet: Proposal-Free LiDAR Point Cloud Panoptic Segmentation},
    booktitle = {Proceedings of the IEEE/CVF Conference on Computer Vision and Pattern Recognition (CVPR)},
%    % month     = {June},
    year      = {2021},
    pages     = {13194-13203}
}

@InProceedings{Zhang_2020_CVPR,
author = {Zhang, Yang and Zhou, Zixiang and David, Philip and Yue, Xiangyu and Xi, Zerong and Gong, Boqing and Foroosh, Hassan},
title = {PolarNet: An Improved Grid Representation for Online LiDAR Point Clouds Semantic Segmentation},
booktitle = {Proceedings of the IEEE/CVF Conference on Computer Vision and Pattern Recognition (CVPR)},
% month = {June},
year = {2020}
}

@inproceedings{hong2021lidar,
  title={Lidar-based panoptic segmentation via dynamic shifting network},
  author={Hong, Fangzhou and Zhou, Hui and Zhu, Xinge and Li, Hongsheng and Liu, Ziwei},
  booktitle={Proceedings of the IEEE/CVF Conference on Computer Vision and Pattern Recognition},
  pages={13090--13099},
  year={2021}
}

@inproceedings{weber2020single,
  title={Single-shot panoptic segmentation},
  author={Weber, Mark and Luiten, Jonathon and Leibe, Bastian},
  booktitle={2020 IEEE/RSJ International Conference on Intelligent Robots and Systems (IROS)},
  pages={8476--8483},
  year={2020},
  organization={IEEE}
}

@article{chen2020scaling,
  title={Scaling Wide Residual Networks for Panoptic Segmentation},
  author={Chen, Liang-Chieh and Wang, Huiyu and Qiao, Siyuan},
  journal={arXiv preprint arXiv:2011.11675},
  year={2020}
}

@article{jayasumana2019bipartite,
  title={Bipartite conditional random fields for panoptic segmentation},
  author={Jayasumana, Sadeep and Ranasinghe, Kanchana and Jayawardhana, Mayuka and Liyanaarachchi, Sahan and Ranasinghe, Harsha},
  journal={arXiv preprint arXiv:1912.05307},
  year={2019}
}

@InProceedings{Wang_2021_CVPR,
    author    = {Wang, Huiyu and Zhu, Yukun and Adam, Hartwig and Yuille, Alan and Chen, Liang-Chieh},
    title     = {MaX-DeepLab: End-to-End Panoptic Segmentation With Mask Transformers},
    booktitle = {Proceedings of the IEEE/CVF Conference on Computer Vision and Pattern Recognition (CVPR)},
%    % month     = {June},
    year      = {2021},
    pages     = {5463-5474}
}

@InProceedings{Zhang_2021_ICCV,
    author    = {Zhang, Jiaming and Yang, Kailun and Constantinescu, Angela and Peng, Kunyu and M\"uller, Karin and Stiefelhagen, Rainer},
    title     = {Trans4Trans: Efficient Transformer for Transparent Object Segmentation To Help Visually Impaired People Navigate in the Real World},
    booktitle = {Proceedings of the IEEE/CVF International Conference on Computer Vision (ICCV) Workshops},
    % month     = {October},
    year      = {2021},
    pages     = {1760-1770}
}

@InProceedings{Zheng_2021_CVPR,
    author    = {Zheng, Sixiao and Lu, Jiachen and Zhao, Hengshuang and Zhu, Xiatian and Luo, Zekun and Wang, Yabiao and Fu, Yanwei and Feng, Jianfeng and Xiang, Tao and Torr, Philip H.S. and Zhang, Li},
    title     = {Rethinking Semantic Segmentation From a Sequence-to-Sequence Perspective With Transformers},
    booktitle = {Proceedings of the IEEE/CVF Conference on Computer Vision and Pattern Recognition (CVPR)},
    % month     = {June},
    year      = {2021},
    pages     = {6881-6890}
}

@inproceedings{carion2020end,
  title={End-to-end object detection with transformers},
  author={Carion, Nicolas and Massa, Francisco and Synnaeve, Gabriel and Usunier, Nicolas and Kirillov, Alexander and Zagoruyko, Sergey},
  booktitle={European Conference on Computer Vision},
  pages={213--229},
  year={2020},
  organization={Springer}
}

@article{chen2020unifying,
  title={Unifying Instance and Panoptic Segmentation with Dynamic Rank-1 Convolutions},
  author={Chen, Hao and Shen, Chunhua and Tian, Zhi},
  journal={arXiv preprint arXiv:2011.09796},
  year={2020}
}

@article{zhang2021k,
  title={K-Net: Towards Unified Image Segmentation},
  author={Zhang, Wenwei and Pang, Jiangmiao and Chen, Kai and Loy, Chen Change},
  journal={arXiv preprint arXiv:2106.14855},
  year={2021}
}

@InProceedings{Li_2021_CVPR,
    author    = {Li, Yanwei and Zhao, Hengshuang and Qi, Xiaojuan and Wang, Liwei and Li, Zeming and Sun, Jian and Jia, Jiaya},
    title     = {Fully Convolutional Networks for Panoptic Segmentation},
    booktitle = {Proceedings of the IEEE/CVF Conference on Computer Vision and Pattern Recognition (CVPR)},
    % month     = {June},
    year      = {2021},
    pages     = {214-223}
}

@article{chen2020panonet,
  title={PanoNet: Real-time Panoptic Segmentation through Position-Sensitive Feature Embedding},
  author={Chen, Xia and Wang, Jianren and Hebert, Martial},
  journal={arXiv preprint arXiv:2008.00192},
  year={2020}
}

@InProceedings{Wang_2020_CVPR,
author = {Wang, Haochen and Luo, Ruotian and Maire, Michael and Shakhnarovich, Greg},
title = {Pixel Consensus Voting for Panoptic Segmentation},
booktitle = {Proceedings of the IEEE/CVF Conference on Computer Vision and Pattern Recognition (CVPR)},
% month = {June},
year = {2020}
}

@inproceedings{wang2020axial,
  title={Axial-deeplab: Stand-alone axial-attention for panoptic segmentation},
  author={Wang, Huiyu and Zhu, Yukun and Green, Bradley and Adam, Hartwig and Yuille, Alan and Chen, Liang-Chieh},
  booktitle={European Conference on Computer Vision},
  pages={108--126},
  year={2020},
  organization={Springer}
}

@inproceedings{cheng2020panoptic,
  title={Panoptic-deeplab: A simple, strong, and fast baseline for bottom-up panoptic segmentation},
  author={Cheng, Bowen and Collins, Maxwell D and Zhu, Yukun and Liu, Ting and Huang, Thomas S and Adam, Hartwig and Chen, Liang-Chieh},
  booktitle={Proceedings of the IEEE/CVF conference on computer vision and pattern recognition},
  pages={12475--12485},
  year={2020}
}

@article{yang2019deeperlab,
  title={Deeperlab: Single-shot image parser},
  author={Yang, Tien-Ju and Collins, Maxwell D and Zhu, Yukun and Hwang, Jyh-Jing and Liu, Ting and Zhang, Xiao and Sze, Vivienne and Papandreou, George and Chen, Liang-Chieh},
  journal={arXiv preprint arXiv:1902.05093},
  year={2019}
}

@ARTICLE{9466324,
  author={Gao, Naiyu and Shan, Yanhu and Zhao, Xin and Huang, Kaiqi},
  journal={IEEE Transactions on Image Processing}, 
  title={Learning Category- and Instance-Aware Pixel Embedding for Fast Panoptic Segmentation}, 
  year={2021},
  volume={30},
  number={},
  pages={6013-6023},
  doi={10.1109/TIP.2021.3090522}}

@InProceedings{Chang_2020_ACCV,
    author    = {Chang, Chia-Yuan and Chang, Shuo-En and Hsiao, Pei-Yung and Fu, Li-Chen},
    title     = {EPSNet: Efficient Panoptic Segmentation Network with Cross-layer Attention Fusion},
    booktitle = {Proceedings of the Asian Conference on Computer Vision (ACCV)},
    % month     = {November},
    year      = {2020}
}

@article{chen2020spatialflow,
  title={Spatialflow: Bridging all tasks for panoptic segmentation},
  author={Chen, Qiang and Cheng, Anda and He, Xiangyu and Wang, Peisong and Cheng, Jian},
  journal={IEEE Transactions on Circuits and Systems for Video Technology},
  volume={31},
  number={6},
  pages={2288--2300},
  year={2020},
  publisher={IEEE}
}

@article{gasperini2021panoster,
  title={Panoster: End-to-end panoptic segmentation of lidar point clouds},
  author={Gasperini, Stefano and Mahani, Mohammad-Ali Nikouei and Marcos-Ramiro, Alvaro and Navab, Nassir and Tombari, Federico},
  journal={IEEE Robotics and Automation Letters},
  volume={6},
  number={2},
  pages={3216--3223},
  year={2021},
  publisher={IEEE}
}

@inproceedings{lin2017focal,
  title={Focal loss for dense object detection},
  author={Lin, Tsung-Yi and Goyal, Priya and Girshick, Ross and He, Kaiming and Doll{\'a}r, Piotr},
  booktitle={Proceedings of the IEEE international conference on computer vision},
  pages={2980--2988},
  year={2017}
}

@InProceedings{Hou_2020_CVPR,
author = {Hou, Rui and Li, Jie and Bhargava, Arjun and Raventos, Allan and Guizilini, Vitor and Fang, Chao and Lynch, Jerome and Gaidon, Adrien},
title = {Real-Time Panoptic Segmentation From Dense Detections},
booktitle = {Proceedings of the IEEE/CVF Conference on Computer Vision and Pattern Recognition (CVPR)},
% month = {June},
year = {2020}
}

@article{de2020fast,
  title={Fast panoptic segmentation network},
  author={de Geus, Daan and Meletis, Panagiotis and Dubbelman, Gijs},
  journal={IEEE Robotics and Automation Letters},
  volume={5},
  number={2},
  pages={1742--1749},
  year={2020},
  publisher={IEEE}
}

@INPROCEEDINGS{9340837,
  author={Milioto, Andres and Behley, Jens and McCool, Chris and Stachniss, Cyrill},
  booktitle={2020 IEEE/RSJ International Conference on Intelligent Robots and Systems (IROS)}, 
  title={LiDAR Panoptic Segmentation for Autonomous Driving}, 
  year={2020},
  volume={},
  number={},
  pages={8505-8512},
  doi={10.1109/IROS45743.2020.9340837}}

@inproceedings{graber2021panoptic,
  title={Panoptic segmentation forecasting},
  author={Graber, Colin and Tsai, Grace and Firman, Michael and Brostow, Gabriel and Schwing, Alexander G},
  booktitle={Proceedings of the IEEE/CVF Conference on Computer Vision and Pattern Recognition},
  pages={12517--12526},
  year={2021}
}

@inproceedings{saxena2005learning,
  title={Learning depth from single monocular images},
  author={Saxena, Ashutosh and Chung, Sung H and Ng, Andrew Y and others},
  booktitle={NIPS},
  volume={18},
  pages={1--8},
  year={2005}
}

@InProceedings{Qiao_2021_CVPR,
    author    = {Qiao, Siyuan and Zhu, Yukun and Adam, Hartwig and Yuille, Alan and Chen, Liang-Chieh},
    title     = {VIP-DeepLab: Learning Visual Perception With Depth-Aware Video Panoptic Segmentation},
    booktitle = {Proceedings of the IEEE/CVF Conference on Computer Vision and Pattern Recognition (CVPR)},
    % month     = {June},
    year      = {2021},
    pages     = {3997-4008}
}

@inproceedings{kim2020video,
  title={Video panoptic segmentation},
  author={Kim, Dahun and Woo, Sanghyun and Lee, Joon-Young and Kweon, In So},
  booktitle={Proceedings of the IEEE/CVF Conference on Computer Vision and Pattern Recognition},
  pages={9859--9868},
  year={2020}
}

@inproceedings{li2019attention,
  title={Attention-guided unified network for panoptic segmentation},
  author={Li, Yanwei and Chen, Xinze and Zhu, Zheng and Xie, Lingxi and Huang, Guan and Du, Dalong and Wang, Xingang},
  booktitle={Proceedings of the IEEE/CVF Conference on Computer Vision and Pattern Recognition},
  pages={7026--7035},
  year={2019}
}

@article{zhang2020ada,
  title={Ada-Segment: Automated Multi-loss Adaptation for Panoptic Segmentation},
  author={Zhang, Gengwei and Gao, Yiming and Xu, Hang and Zhang, Hao and Li, Zhenguo and Liang, Xiaodan},
  journal={arXiv preprint arXiv:2012.03603},
  year={2020}
}

@inproceedings{porzi2021improving,
  title={Improving Panoptic Segmentation at All Scales},
  author={Porzi, Lorenzo and Bulo, Samuel Rota and Kontschieder, Peter},
  booktitle={Proceedings of the IEEE/CVF Conference on Computer Vision and Pattern Recognition},
  pages={7302--7311},
  year={2021}
}

@inproceedings{li2020unifying,
  title={Unifying training and inference for panoptic segmentation},
  author={Li, Qizhu and Qi, Xiaojuan and Torr, Philip HS},
  booktitle={Proceedings of the IEEE/CVF Conference on Computer Vision and Pattern Recognition},
  pages={13320--13328},
  year={2020}
}

@InProceedings{Xiong_2019_CVPR,
author = {Xiong, Yuwen and Liao, Renjie and Zhao, Hengshuang and Hu, Rui and Bai, Min and Yumer, Ersin and Urtasun, Raquel},
title = {UPSNet: A Unified Panoptic Segmentation Network},
booktitle = {Proceedings of the IEEE/CVF Conference on Computer Vision and Pattern Recognition (CVPR)},
% month = {June},
year = {2019}
}

@InProceedings{Huang_2021_CVPR,
    author    = {Huang, Jiaxing and Guan, Dayan and Xiao, Aoran and Lu, Shijian},
    title     = {Cross-View Regularization for Domain Adaptive Panoptic Segmentation},
    booktitle = {Proceedings of the IEEE/CVF Conference on Computer Vision and Pattern Recognition (CVPR)},
    % month     = {June},
    year      = {2021},
    pages     = {10133-10144}
}

@inproceedings{arbelaez2014multiscale,
  title={Multiscale combinatorial grouping},
  author={Arbel{\'a}ez, Pablo and Pont-Tuset, Jordi and Barron, Jonathan T and Marques, Ferran and Malik, Jitendra},
  booktitle={Proceedings of the IEEE conference on computer vision and pattern recognition},
  pages={328--335},
  year={2014}
}

@article{rother2004grabcut,
  title={" GrabCut" interactive foreground extraction using iterated graph cuts},
  author={Rother, Carsten and Kolmogorov, Vladimir and Blake, Andrew},
  journal={ACM transactions on graphics (TOG)},
  volume={23},
  number={3},
  pages={309--314},
  year={2004},
  publisher={ACM New York, NY, USA}
}

@InProceedings{Li_2018_ECCV,
author = {Li, Qizhu and Arnab, Anurag and Torr, Philip H.S.},
title = {Weakly- and Semi-Supervised Panoptic Segmentation},
booktitle = {Proceedings of the European Conference on Computer Vision (ECCV)},
% month = {September},
year = {2018}
}

@inproceedings{wu2020bidirectional,
  title={Bidirectional graph reasoning network for panoptic segmentation},
  author={Wu, Yangxin and Zhang, Gengwei and Gao, Yiming and Deng, Xiajun and Gong, Ke and Liang, Xiaodan and Lin, Liang},
  booktitle={Proceedings of the IEEE/CVF Conference on Computer Vision and Pattern Recognition},
  pages={9080--9089},
  year={2020}
}

@inproceedings{yang2020sognet,
  title={Sognet: Scene overlap graph network for panoptic segmentation},
  author={Yang, Yibo and Li, Hongyang and Li, Xia and Zhao, Qijie and Wu, Jianlong and Lin, Zhouchen},
  booktitle={Proceedings of the AAAI Conference on Artificial Intelligence},
  volume={34},
  number={07},
  pages={12637--12644},
  year={2020}
}

@inproceedings{lazarow2020learning,
  title={Learning instance occlusion for panoptic segmentation},
  author={Lazarow, Justin and Lee, Kwonjoon and Shi, Kunyu and Tu, Zhuowen},
  booktitle={Proceedings of the IEEE/CVF Conference on Computer Vision and Pattern Recognition},
  pages={10720--10729},
  year={2020}
}

@inproceedings{liu2019end,
  title={An end-to-end network for panoptic segmentation},
  author={Liu, Huanyu and Peng, Chao and Yu, Changqian and Wang, Jingbo and Liu, Xu and Yu, Gang and Jiang, Wei},
  booktitle={Proceedings of the IEEE/CVF Conference on Computer Vision and Pattern Recognition},
  pages={6172--6181},
  year={2019}
}

@article{wu2020auto,
  title={Auto-Panoptic: Cooperative Multi-Component Architecture Search for Panoptic Segmentation},
  author={Wu, Yangxin and Zhang, Gengwei and Xu, Hang and Liang, Xiaodan and Lin, Liang},
  journal={arXiv preprint arXiv:2010.16119},
  year={2020}
}

@inproceedings{lin2017feature,
  title={Feature pyramid networks for object detection},
  author={Lin, Tsung-Yi and Doll{\'a}r, Piotr and Girshick, Ross and He, Kaiming and Hariharan, Bharath and Belongie, Serge},
  booktitle={Proceedings of the IEEE conference on computer vision and pattern recognition},
  pages={2117--2125},
  year={2017}
}

@inproceedings{zhao2017pyramid,
  title={Pyramid scene parsing network},
  author={Zhao, Hengshuang and Shi, Jianping and Qi, Xiaojuan and Wang, Xiaogang and Jia, Jiaya},
  booktitle={Proceedings of the IEEE conference on computer vision and pattern recognition},
  pages={2881--2890},
  year={2017}
}

@inproceedings{he2019dynamic,
  title={Dynamic multi-scale filters for semantic segmentation},
  author={He, Junjun and Deng, Zhongying and Qiao, Yu},
  booktitle={Proceedings of the IEEE/CVF International Conference on Computer Vision},
  pages={3562--3572},
  year={2019}
}

@inproceedings{ding2018context,
  title={Context contrasted feature and gated multi-scale aggregation for scene segmentation},
  author={Ding, Henghui and Jiang, Xudong and Shuai, Bing and Liu, Ai Qun and Wang, Gang},
  booktitle={Proceedings of the IEEE Conference on Computer Vision and Pattern Recognition},
  pages={2393--2402},
  year={2018}
}

@inproceedings{zhang2018exfuse,
  title={Exfuse: Enhancing feature fusion for semantic segmentation},
  author={Zhang, Zhenli and Zhang, Xiangyu and Peng, Chao and Xue, Xiangyang and Sun, Jian},
  booktitle={Proceedings of the European Conference on Computer Vision (ECCV)},
  pages={269--284},
  year={2018}
}

@inproceedings{xie2017aggregated,
  title={Aggregated residual transformations for deep neural networks},
  author={Xie, Saining and Girshick, Ross and Doll{\'a}r, Piotr and Tu, Zhuowen and He, Kaiming},
  booktitle={Proceedings of the IEEE conference on computer vision and pattern recognition},
  pages={1492--1500},
  year={2017}
}

@inproceedings{he2019adaptive,
  title={Adaptive pyramid context network for semantic segmentation},
  author={He, Junjun and Deng, Zhongying and Zhou, Lei and Wang, Yali and Qiao, Yu},
  booktitle={Proceedings of the IEEE/CVF Conference on Computer Vision and Pattern Recognition},
  pages={7519--7528},
  year={2019}
}

@inproceedings{lin2018multi,
  title={Multi-scale context intertwining for semantic segmentation},
  author={Lin, Di and Ji, Yuanfeng and Lischinski, Dani and Cohen-Or, Daniel and Huang, Hui},
  booktitle={Proceedings of the European Conference on Computer Vision (ECCV)},
  pages={603--619},
  year={2018}
}

@inproceedings{liu2015parsenet,
author = {Liu, Wei and Rabinovich, Andrew and Berg, Alexander},
year = {2016},
% month = {06},
pages = {},
title = {ParseNet: Looking Wider to See Better}
}

@inproceedings{lin2017refinenet,
  title={Refinenet: Multi-path refinement networks for high-resolution semantic segmentation},
  author={Lin, Guosheng and Milan, Anton and Shen, Chunhua and Reid, Ian},
  booktitle={Proceedings of the IEEE conference on computer vision and pattern recognition},
  pages={1925--1934},
  year={2017}
}

@inproceedings{huang2021fapn,
  title={FaPN: Feature-Aligned Pyramid Network for Dense Image Prediction},
  author={Huang, Shihua and Lu, Zhichao and Cheng, Ran and He, Cheng},
  booktitle={Proceedings of the IEEE/CVF International Conference on Computer Vision},
  pages={864--873},
  year={2021}
}

@inproceedings{he2016deep,
  title={Deep residual learning for image recognition},
  author={He, Kaiming and Zhang, Xiangyu and Ren, Shaoqing and Sun, Jian},
  booktitle={Proceedings of the IEEE conference on computer vision and pattern recognition},
  pages={770--778},
  year={2016}
}

@inproceedings{dai2017deformable,
  title={Deformable convolutional networks},
  author={Dai, Jifeng and Qi, Haozhi and Xiong, Yuwen and Li, Yi and Zhang, Guodong and Hu, Han and Wei, Yichen},
  booktitle={Proceedings of the IEEE international conference on computer vision},
  pages={764--773},
  year={2017}
}

@article{chen2017deeplab,
  title={Deeplab: Semantic image segmentation with deep convolutional nets, atrous convolution, and fully connected crfs},
  author={Chen, Liang-Chieh and Papandreou, George and Kokkinos, Iasonas and Murphy, Kevin and Yuille, Alan L},
  journal={IEEE transactions on pattern analysis and machine intelligence},
  volume={40},
  number={4},
  pages={834--848},
  year={2017},
  publisher={IEEE}
}

@inproceedings{chen2018encoder,
  title={Encoder-decoder with atrous separable convolution for semantic image segmentation},
  author={Chen, Liang-Chieh and Zhu, Yukun and Papandreou, George and Schroff, Florian and Adam, Hartwig},
  booktitle={Proceedings of the European conference on computer vision (ECCV)},
  pages={801--818},
  year={2018}
}

@article{chen2017rethinking,
  title={Rethinking atrous convolution for semantic image segmentation},
  author={Chen, Liang-Chieh and Papandreou, George and Schroff, Florian and Adam, Hartwig},
  journal={arXiv preprint arXiv:1706.05587},
  year={2017}
}

@article{yu2015multi,
  title={Multi-scale context aggregation by dilated convolutions},
  author={Yu, Fisher and Koltun, Vladlen},
  journal={arXiv preprint arXiv:1511.07122},
  year={2015}
}

@inproceedings{wang2018understanding,
  title={Understanding convolution for semantic segmentation},
  author={Wang, Panqu and Chen, Pengfei and Yuan, Ye and Liu, Ding and Huang, Zehua and Hou, Xiaodi and Cottrell, Garrison},
  booktitle={2018 IEEE winter conference on applications of computer vision (WACV)},
  pages={1451--1460},
  year={2018},
  organization={IEEE}
}

@inproceedings{yang2018denseaspp,
  title={Denseaspp for semantic segmentation in street scenes},
  author={Yang, Maoke and Yu, Kun and Zhang, Chi and Li, Zhiwei and Yang, Kuiyuan},
  booktitle={Proceedings of the IEEE conference on computer vision and pattern recognition},
  pages={3684--3692},
  year={2018}
}

@article{li2019global,
  title={Global aggregation then local distribution in fully convolutional networks},
  author={Li, Xiangtai and Zhang, Li and You, Ansheng and Yang, Maoke and Yang, Kuiyuan and Tong, Yunhai},
  journal={arXiv preprint arXiv:1909.07229},
  year={2019}
}

@inproceedings{yuan2020multi,
  title={Multi Receptive Field Network for Semantic Segmentation},
  author={Yuan, Jianlong and Deng, Zelu and Wang, Shu and Luo, Zhenbo},
  booktitle={2020 IEEE Winter Conference on Applications of Computer Vision (WACV)},
  pages={1883--1892},
  year={2020},
  organization={IEEE}
}

@inproceedings{fu2019dual,
  title={Dual attention network for scene segmentation},
  author={Fu, Jun and Liu, Jing and Tian, Haijie and Li, Yong and Bao, Yongjun and Fang, Zhiwei and Lu, Hanqing},
  booktitle={Proceedings of the IEEE/CVF Conference on Computer Vision and Pattern Recognition},
  pages={3146--3154},
  year={2019}
}

@inproceedings{zhao2018psanet,
  title={Psanet: Point-wise spatial attention network for scene parsing},
  author={Zhao, Hengshuang and Zhang, Yi and Liu, Shu and Shi, Jianping and Loy, Chen Change and Lin, Dahua and Jia, Jiaya},
  booktitle={Proceedings of the European Conference on Computer Vision (ECCV)},
  pages={267--283},
  year={2018}
}

@inproceedings{yuan2020object,
  title={Object-contextual representations for semantic segmentation},
  author={Yuan, Yuhui and Chen, Xilin and Wang, Jingdong},
  booktitle={Computer Vision--ECCV 2020: 16th European Conference, Glasgow, UK, August 23--28, 2020, Proceedings, Part VI 16},
  pages={173--190},
  year={2020},
  organization={Springer}
}

@inproceedings{yuan2021segmentation,
  title={Segmentation transformer: Object-contextual representations for semantic segmentation},
  author={Yuan, Yuhui and Chen, Xiaokang and Chen, Xilin and Wang, Jingdong},
  booktitle={European Conference on Computer Vision (ECCV)},
  volume={1},
  year={2021}
}

@article{xie2021segformer,
  title={SegFormer: Simple and efficient design for semantic segmentation with transformers},
  author={Xie, Enze and Wang, Wenhai and Yu, Zhiding and Anandkumar, Anima and Alvarez, Jose M and Luo, Ping},
  journal={Advances in Neural Information Processing Systems},
  volume={34},
  pages={12077--12090},
  year={2021}
}

@article{dosovitskiy2020image,
  title={An image is worth 16x16 words: Transformers for image recognition at scale},
  author={Dosovitskiy, Alexey and Beyer, Lucas and Kolesnikov, Alexander and Weissenborn, Dirk and Zhai, Xiaohua and Unterthiner, Thomas and Dehghani, Mostafa and Minderer, Matthias and Heigold, Georg and Gelly, Sylvain and others},
  journal={arXiv preprint arXiv:2010.11929},
  year={2020}
}

@inproceedings{li2019expectation,
  title={Expecta\-tion-maximization attention networks for semantic segmentation},
  author={Li, Xia and Zhong, Zhisheng and Wu, Jianlong and Yang, Yibo and Lin, Zhouchen and Liu, Hong},
  booktitle={Proceedings of the IEEE/CVF International Conference on Computer Vision},
  pages={9167--9176},
  year={2019}
}

@inproceedings{yu2018learning,
  title={Learning a discriminative feature network for semantic segmentation},
  author={Yu, Changqian and Wang, Jingbo and Peng, Chao and Gao, Changxin and Yu, Gang and Sang, Nong},
  booktitle={Proceedings of the IEEE conference on computer vision and pattern recognition},
  pages={1857--1866},
  year={2018}
}

@inproceedings{zhong2020squeeze,
  title={Squeeze-and-attention networks for semantic segmentation},
  author={Zhong, Zilong and Lin, Zhong Qiu and Bidart, Rene and Hu, Xiaodan and Daya, Ibrahim Ben and Li, Zhifeng and Zheng, Wei-Shi and Li, Jonathan and Wong, Alexander},
  booktitle={Proceedings of the IEEE/CVF Conference on Computer Vision and Pattern Recognition},
  pages={13065--13074},
  year={2020}
}

@inproceedings{hu2018squeeze,
  title={Squeeze-and-excitation networks},
  author={Hu, Jie and Shen, Li and Sun, Gang},
  booktitle={Proceedings of the IEEE conference on computer vision and pattern recognition},
  pages={7132--7141},
  year={2018}
}

@inproceedings{liu2021swin,
author = {Liu, Ze and Lin, Yutong and Cao, Yue and Hu, Han and Wei, Yixuan and Zhang, Zheng and Lin, Stephen and Guo, Baining},
year = {2021},
% month = {10},
pages = {9992-10002},
title = {Swin Transformer: Hierarchical Vision Transformer using Shifted Windows},
doi = {10.1109/ICCV48922.2021.00986}
}

@inproceedings{huang2019ccnet,
  title={Ccnet: Criss-cross attention for semantic segmentation},
  author={Huang, Zilong and Wang, Xinggang and Huang, Lichao and Huang, Chang and Wei, Yunchao and Liu, Wenyu},
  booktitle={Proceedings of the IEEE/CVF International Conference on Computer Vision},
  pages={603--612},
  year={2019}
}

@inproceedings{chen2016attention,
  title={Attention to scale: Scale-aware semantic image segmentation},
  author={Chen, Liang-Chieh and Yang, Yi and Wang, Jiang and Xu, Wei and Yuille, Alan L},
  booktitle={Proceedings of the IEEE conference on computer vision and pattern recognition},
  pages={3640--3649},
  year={2016}
}

@article{tao2020hierarchical,
  title={Hierarchical multi-scale attention for semantic segmentation},
  author={Tao, Andrew and Sapra, Karan and Catanzaro, Bryan},
  journal={arXiv preprint arXiv:2005.10821},
  year={2020}
}

@article{cheng2021per,
  title={Per-pixel classification is not all you need for semantic segmentation},
  author={Cheng, Bowen and Schwing, Alex and Kirillov, Alexander},
  journal={Advances in Neural Information Processing Systems},
  volume={34},
  pages={17864--17875},
  year={2021}
}

@inproceedings{zheng2015conditional,
  title={Conditional random fields as recurrent neural networks},
  author={Zheng, Shuai and Jayasumana, Sadeep and Romera-Paredes, Bernardino and Vineet, Vibhav and Su, Zhizhong and Du, Dalong and Huang, Chang and Torr, Philip HS},
  booktitle={Proceedings of the IEEE international conference on computer vision},
  pages={1529--1537},
  year={2015}
}

@inproceedings{chen2016semantic,
  title={Semantic image segmentation with task-specific edge detection using cnns and a discriminatively trained domain transform},
  author={Chen, Liang-Chieh and Barron, Jonathan T and Papandreou, George and Murphy, Kevin and Yuille, Alan L},
  booktitle={Proceedings of the IEEE conference on computer vision and pattern recognition},
  pages={4545--4554},
  year={2016}
}

@inproceedings{hu2019acnet,
  title={Acnet: Attention based network to exploit complementary features for rgbd semantic segmentation},
  author={Hu, Xinxin and Yang, Kailun and Fei, Lei and Wang, Kaiwei},
  booktitle={2019 IEEE International Conference on Image Processing (ICIP)},
  pages={1440--1444},
  year={2019},
  organization={IEEE}
}

@article{paszke2016enet,
  title={Enet: A deep neural network architecture for real-time semantic segmentation},
  author={Paszke, Adam and Chaurasia, Abhishek and Kim, Sangpil and Culurciello, Eugenio},
  journal={arXiv preprint arXiv:1606.02147},
  year={2016}
}

@article{raj2015multi,
  title={Multi-scale convolutional architecture for semantic segmentation},
  author={Raj, Aman and Maturana, Daniel and Scherer, Sebastian},
  journal={Robotics Institute, Carnegie Mellon University, Tech. Rep. CMU-RITR-15-21},
  year={2015}
}

@inproceedings{roy2016multi,
  title={A multi-scale cnn for affordance segmentation in rgb images},
  author={Roy, Anirban and Todorovic, Sinisa},
  booktitle={European conference on computer vision},
  pages={186--201},
  year={2016},
  organization={Springer}
}

@inproceedings{bian2016multiscale,
  title={Multiscale fully convolutional network with application to industrial inspection},
  author={Bian, Xiao and Lim, Ser Nam and Zhou, Ning},
  booktitle={2016 IEEE winter conference on applications of computer vision (WACV)},
  pages={1--8},
  year={2016},
  organization={IEEE}
}

@inproceedings{pinheiro2014recurrent,
  title={Recurrent convolutional neural networks for scene labeling},
  author={Pinheiro, Pedro and Collobert, Ronan},
  booktitle={International conference on machine learning},
  pages={82--90},
  year={2014},
  organization={PMLR}
}

@article{krahenbuhl2011efficient,
  title={Efficient inference in fully connected crfs with gaussian edge potentials},
  author={Kr{\"a}henb{\"u}hl, Philipp and Koltun, Vladlen},
  journal={Advances in neural information processing systems},
  volume={24},
  pages={109--117},
  year={2011}
}

@inproceedings{lin2016efficient,
  title={Efficient piecewise training of deep structured models for semantic segmentation},
  author={Lin, Guosheng and Shen, Chunhua and Van Den Hengel, Anton and Reid, Ian},
  booktitle={Proceedings of the IEEE conference on computer vision and pattern recognition},
  pages={3194--3203},
  year={2016}
}

@article{mozer1989focused,
  title={A focused back-propagation algorithm for temporal pattern recognition},
  author={Mozer, Michael C},
  journal={Complex systems},
  volume={3},
  number={4},
  pages={349--381},
  year={1989}
}

@inproceedings{liu2015semantic,
  title={Semantic image segmentation via deep parsing network},
  author={Liu, Ziwei and Li, Xiaoxiao and Luo, Ping and Loy, Chen-Change and Tang, Xiaoou},
  booktitle={Proceedings of the IEEE international conference on computer vision},
  pages={1377--1385},
  year={2015}
}

@inproceedings{vemulapalli2016gaussian,
  title={Gaussian conditional random field network for semantic segmentation},
  author={Vemulapalli, Raviteja and Tuzel, Oncel and Liu, Ming-Yu and Chellapa, Rama},
  booktitle={Proceedings of the IEEE conference on computer vision and pattern recognition},
  pages={3224--3233},
  year={2016}
}

@inproceedings{bell2015material,
  title={Material recognition in the wild with the materials in context database},
  author={Bell, Sean and Upchurch, Paul and Snavely, Noah and Bala, Kavita},
  booktitle={Proceedings of the IEEE conference on computer vision and pattern recognition},
  pages={3479--3487},
  year={2015}
}

@inproceedings{fu2016retinal,
  title={Retinal vessel segmentation via deep learning network and fully-connected conditional random fields},
  author={Fu, Huazhu and Xu, Yanwu and Wong, Damon Wing Kee and Liu, Jiang},
  booktitle={2016 IEEE 13th international symposium on biomedical imaging (ISBI)},
  pages={698--701},
  year={2016},
  organization={IEEE}
}

@inproceedings{visin2016reseg,
  title={Reseg: A recurrent neural network-based model for semantic segmentation},
  author={Visin, Francesco and Ciccone, Marco and Romero, Adriana and Kastner, Kyle and Cho, Kyunghyun and Bengio, Yoshua and Matteucci, Matteo and Courville, Aaron},
  booktitle={Proceedings of the IEEE Conference on Computer Vision and Pattern Recognition Workshops},
  pages={41--48},
  year={2016}
}

@article{visin2015recurrent,
  title={A recurrent neural network based alternative to convolutional networks},
  author={Visin, F and Kastner, K and Cho, K and Matteucci, M and Courville, A and Bengio, Y},
  journal={arXiv preprint arXiv:1505.00393},
  year={2015}
}

@inproceedings{shuai2016dag,
  title={Dag-recurrent neural networks for scene labeling},
  author={Shuai, Bing and Zuo, Zhen and Wang, Bing and Wang, Gang},
  booktitle={Proceedings of the IEEE conference on computer vision and pattern recognition},
  pages={3620--3629},
  year={2016}
}

@inproceedings{byeon2015scene,
  title={Scene labeling with lstm recurrent neural networks},
  author={Byeon, Wonmin and Breuel, Thomas M and Raue, Federico and Liwicki, Marcus},
  booktitle={Proceedings of the IEEE Conference on Computer Vision and Pattern Recognition},
  pages={3547--3555},
  year={2015}
}

@inproceedings{li2016lstm,
  title={Lstm-cf: Unifying context modeling and fusion with lstms for rgb-d scene labeling},
  author={Li, Zhen and Gan, Yukang and Liang, Xiaodan and Yu, Yizhou and Cheng, Hui and Lin, Liang},
  booktitle={European conference on computer vision},
  pages={541--557},
  year={2016},
  organization={Springer}
}

@inproceedings{liang2016semantic,
  title={Semantic object parsing with graph lstm},
  author={Liang, Xiaodan and Shen, Xiaohui and Feng, Jiashi and Lin, Liang and Yan, Shuicheng},
  booktitle={European Conference on Computer Vision},
  pages={125--143},
  year={2016},
  organization={Springer}
}

@techreport{achanta2010slic,
  title={Slic superpixels},
  author={Achanta, Radhakrishna and Shaji, Appu and Smith, Kevin and Lucchi, Aurelien and Fua, Pascal and S{\"u}sstrunk, Sabine},
  year={2010}
}

@inproceedings{hu2016segmentation,
  title={Segmentation from natural language expressions},
  author={Hu, Ronghang and Rohrbach, Marcus and Darrell, Trevor},
  booktitle={European Conference on Computer Vision},
  pages={108--124},
  year={2016},
  organization={Springer}
}

@article{yuan2018ocnet,
author = {Yuan, Yuhui and Huang, Lang and Guo, Jianyuan and Zhang, Chao and Chen, Xilin and Wang, Jingdong},
year = {2021},
pages = {},
title = {OCNet: Object Context for Semantic Segmentation},
volume = {129},
journal = {International Journal of Computer Vision},
doi = {10.1007/s11263-021-01465-9}
}

@inproceedings{takikawa2019gated,
  title={Gated-scnn: Gated shape cnns for semantic segmentation},
  author={Takikawa, Towaki and Acuna, David and Jampani, Varun and Fidler, Sanja},
  booktitle={Proceedings of the IEEE/CVF International Conference on Computer Vision},
  pages={5229--5238},
  year={2019}
}

@inproceedings{zhang2018context,
  title={Context encoding for semantic segmentation},
  author={Zhang, Hang and Dana, Kristin and Shi, Jianping and Zhang, Zhongyue and Wang, Xiaogang and Tyagi, Ambrish and Agrawal, Amit},
  booktitle={Proceedings of the IEEE conference on Computer Vision and Pattern Recognition},
  pages={7151--7160},
  year={2018}
}

@article{li2018pyramid,
  title={Pyramid attention network for semantic segmentation},
  author={Li, Hanchao and Xiong, Pengfei and An, Jie and Wang, Lingxue},
  journal={arXiv preprint arXiv:1805.10180},
  year={2018}
}

@article{luc2016semantic,
  title={Semantic segmentation using adversarial networks},
  author={Luc, Pauline and Couprie, Camille and Chintala, Soumith and Verbeek, Jakob},
  journal={arXiv preprint arXiv:1611. 08408},
  year={2016}
}

@inproceedings{souly2017semi,
  title={Semi supervised semantic segmentation using generative adversarial network},
  author={Souly, Nasim and Spampinato, Concetto and Shah, Mubarak},
  booktitle={Proceedings of the IEEE international conference on computer vision},
  pages={5688--5696},
  year={2017}
}

@article{hung2018adversarial,
  title={Adversarial learning for semi-supervised semantic segmentation},
  author={Hung, Wei-Chih and Tsai, Yi-Hsuan and Liou, Yan-Ting and Lin, Yen-Yu and Yang, Ming-Hsuan},
  journal={arXiv preprint arXiv:1802.07934},
  year={2018}
}

@article{xue2018segan,
  title={Segan: Adversarial network with multi-scale l 1 loss for medical image segmentation},
  author={Xue, Yuan and Xu, Tao and Zhang, Han and Long, L Rodney and Huang, Xiaolei},
  journal={Neuroinformatics},
  volume={16},
  number={3},
  pages={383--392},
  year={2018},
  publisher={Springer}
}

@inproceedings{chartsias2017adversarial,
  title={Adversarial image synthesis for unpaired multi-modal cardiac data},
  author={Chartsias, Agisilaos and Joyce, Thomas and Dharmakumar, Rohan and Tsaftaris, Sotirios A},
  booktitle={International workshop on simulation and synthesis in medical imaging},
  pages={3--13},
  year={2017},
  organization={Springer}
}

@inproceedings{zhang2018translating,
  title={Translating and segmenting multimodal medical volumes with cycle-and shape-consistency generative adversarial network},
  author={Zhang, Zizhao and Yang, Lin and Zheng, Yefeng},
  booktitle={Proceedings of the IEEE conference on computer vision and pattern Recognition},
  pages={9242--9251},
  year={2018}
}

@inproceedings{shin2018medical,
  title={Medical image synthesis for data augmentation and anonymization using generative adversarial networks},
  author={Shin, Hoo-Chang and Tenenholtz, Neil A and Rogers, Jameson K and Schwarz, Christopher G and Senjem, Matthew L and Gunter, Jeffrey L and Andriole, Katherine P and Michalski, Mark},
  booktitle={International workshop on simulation and synthesis in medical imaging},
  pages={1--11},
  year={2018},
  organization={Springer}
}

@article{changhee2019learning,
  title={Learning more with less: GAN-based medical image augmentation},
  author={Changhee, H and Kohei, M and Shin'ichi, S and Hideki, N},
  journal={Med. Imaging Technol},
  volume={6},
  year={2019}
}

@article{yang2018mri,
author = {Yang, Qianye and Li, Nannan and Zhao, Zixu and Fan, Xingyu and Chang, Eric and Xu, Yan},
year = {2020},
% month = {02},
pages = {3753},
title = {MRI Cross-Modality Image-to-Image Translation},
volume = {10},
journal = {Scientific Reports},
doi = {10.1038/s41598-020-60520-6}
}

@inproceedings{yu20183d,
author = {Yu, Biting and Zhou, Luping and Wang, Lei and Fripp, Jurgen and Bourgeat, Pierrick},
year = {2018},
% month = {04},
pages = {626-630},
title = {3D cGAN based cross-modality MR image synthesis for brain tumor segmentation},
doi = {10.1109/ISBI.2018.8363653}
}

@inproceedings{riegler2017octnet,
  title={Octnet: Learning deep 3d representations at high resolutions},
  author={Riegler, Gernot and Osman Ulusoy, Ali and Geiger, Andreas},
  booktitle={Proceedings of the IEEE conference on computer vision and pattern recognition},
  pages={3577--3586},
  year={2017}
}

@inproceedings{graham20183d,
  title={3d semantic segmentation with submanifold sparse convolutional networks},
  author={Graham, Benjamin and Engelcke, Martin and Van Der Maaten, Laurens},
  booktitle={Proceedings of the IEEE conference on computer vision and pattern recognition},
  pages={9224--9232},
  year={2018}
}

@inproceedings{abhishek2019mask2lesion,
  title={Mask2lesion: Mask-constrained adversarial skin lesion image synthesis},
  author={Abhishek, Kumar and Hamarneh, Ghassan},
  booktitle={International Workshop on Simulation and Synthesis in Medical Imaging},
  pages={71--80},
  year={2019},
  organization={Springer}
}

@inproceedings{mohajerani2019cloudmaskgan,
  title={Cloudmaskgan: A content-aware unpaired image-to-image translation algorithm for remote sensing imagery},
  author={Mohajerani, Sorour and Asad, Reza and Abhishek, Kumar and Sharma, Neha and van Duynhoven, Alysha and Saeedi, Parvaneh},
  booktitle={2019 IEEE International Conference on Image Processing (ICIP)},
  pages={1965--1969},
  year={2019},
  organization={IEEE}
}

@inproceedings{zhang2021dcnas,
  title={Dcnas: Densely connected neural architecture search for semantic image segmentation},
  author={Zhang, Xiong and Xu, Hongmin and Mo, Hong and Tan, Jianchao and Yang, Cheng and Wang, Lei and Ren, Wenqi},
  booktitle={Proceedings of the IEEE/CVF Conference on Computer Vision and Pattern Recognition},
  pages={13956--13967},
  year={2021}
}

@article{chen2018searching,
  title={Searching for efficient multi-scale architectures for dense image prediction},
  author={Chen, Liang-Chieh and Collins, Maxwell and Zhu, Yukun and Papandreou, George and Zoph, Barret and Schroff, Florian and Adam, Hartwig and Shlens, Jon},
  journal={Advances in neural information processing systems},
  volume={31},
  year={2018}
}

@inproceedings{golovin2017google,
  title={Google vizier: A service for black-box optimization},
  author={Golovin, Daniel and Solnik, Benjamin and Moitra, Subhodeep and Kochanski, Greg and Karro, John and Sculley, David},
  booktitle={Proceedings of the 23rd ACM SIGKDD international conference on knowledge discovery and data mining},
  pages={1487--1495},
  year={2017}
}

@inproceedings{fragkiadaki2015learning,
  title={Learning to segment moving objects in videos},
  author={Fragkiadaki, Katerina and Arbelaez, Pablo and Felsen, Panna and Malik, Jitendra},
  booktitle={Proceedings of the IEEE Conference on Computer Vision and Pattern Recognition},
  pages={4083--4090},
  year={2015}
}

@inproceedings{tsai2016video,
  title={Video segmentation via object flow},
  author={Tsai, Yi-Hsuan and Yang, Ming-Hsuan and Black, Michael J},
  booktitle={Proceedings of the IEEE conference on computer vision and pattern recognition},
  pages={3899--3908},
  year={2016}
}

@inproceedings{li2017primary,
  title={Primary video object segmentation via complementary CNNs and neighborhood reversible flow},
  author={Li, Jia and Zheng, Anlin and Chen, Xiaowu and Zhou, Bin},
  booktitle={Proceedings of the IEEE international conference on computer vision},
  pages={1417--1425},
  year={2017}
}

@inproceedings{li2018instance,
  title={Instance embedding transfer to unsupervised video object segmentation},
  author={Li, Siyang and Seybold, Bryan and Vorobyov, Alexey and Fathi, Alireza and Huang, Qin and Kuo, C-C Jay},
  booktitle={Proceedings of the IEEE conference on computer vision and pattern recognition},
  pages={6526--6535},
  year={2018}
}

@inproceedings{li2018unsupervised,
  title={Unsupervised video object segmentation with motion-based bilateral networks},
  author={Li, Siyang and Seybold, Bryan and Vorobyov, Alexey and Lei, Xuejing and Kuo, C-C Jay},
  booktitle={Proceedings of the European conference on computer vision (ECCV)},
  pages={207--223},
  year={2018}
}

@INPROCEEDINGS{6758588,
  author={Nguyen, Anh and Le, Bac},
  booktitle={2013 6th IEEE Conference on Robotics, Automation and Mechatronics (RAM)}, 
  title={3D point cloud segmentation: A survey}, 
  year={2013},
  volume={},
  number={},
  pages={225-230},
  doi={10.1109/RAM.2013.6758588}}

@inproceedings{song2018pyramid,
  title={Pyramid dilated deeper convlstm for video salient object detection},
  author={Song, Hongmei and Wang, Wenguan and Zhao, Sanyuan and Shen, Jianbing and Lam, Kin-Man},
  booktitle={Proceedings of the European conference on computer vision (ECCV)},
  pages={715--731},
  year={2018}
}

@inproceedings{jain2017fusionseg,
  title={Fusionseg: Learning to combine motion and appearance for fully automatic segmentation of generic objects in videos},
  author={Jain, Suyog Dutt and Xiong, Bo and Grauman, Kristen},
  booktitle={2017 IEEE conference on computer vision and pattern recognition (CVPR)},
  pages={2117--2126},
  year={2017},
  organization={IEEE}
}

@inproceedings{cheng2017segflow,
  title={Segflow: Joint learning for video object segmentation and optical flow},
  author={Cheng, Jingchun and Tsai, Yi-Hsuan and Wang, Shengjin and Yang, Ming-Hsuan},
  booktitle={Proceedings of the IEEE international conference on computer vision},
  pages={686--695},
  year={2017}
}

@inproceedings{tokmakov2017learning,
  title={Learning video object segmentation with visual memory},
  author={Tokmakov, Pavel and Alahari, Karteek and Schmid, Cordelia},
  booktitle={Proceedings of the IEEE International Conference on Computer Vision},
  pages={4481--4490},
  year={2017}
}

@inproceedings{li2019motion,
  title={Motion guided attention for video salient object detection},
  author={Li, Haofeng and Chen, Guanqi and Li, Guanbin and Yu, Yizhou},
  booktitle={Proceedings of the IEEE/CVF International Conference on Computer Vision},
  pages={7274--7283},
  year={2019}
}

@inproceedings{zhou2020motion,
  title={Motion-attentive transition for zero-shot video object segmentation},
  author={Zhou, Tianfei and Wang, Shunzhou and Zhou, Yi and Yao, Yazhou and Li, Jianwu and Shao, Ling},
  booktitle={Proceedings of the AAAI Conference on Artificial Intelligence},
  volume={34},
  number={07},
  pages={13066--13073},
  year={2020}
}

@article{lu2020zero,
  title={Zero-shot video object segmentation with co-attention siamese networks},
  author={Lu, Xiankai and Wang, Wenguan and Shen, Jianbing and Crandall, David and Luo, Jiebo},
  journal={IEEE transactions on pattern analysis and machine intelligence},
  year={2020},
  publisher={IEEE}
}

@inproceedings{ren2021reciprocal,
  title={Reciprocal Transformations for Unsupervised Video Object Segmentation},
  author={Ren, Sucheng and Liu, Wenxi and Liu, Yongtuo and Chen, Haoxin and Han, Guoqiang and He, Shengfeng},
  booktitle={Proceedings of the IEEE/CVF Conference on Computer Vision and Pattern Recognition},
  pages={15455--15464},
  year={2021}
}

@inproceedings{zhang2020unsupervised,
  title={Unsupervised video object segmentation with joint hotspot tracking},
  author={Zhang, Lu and Zhang, Jianming and Lin, Zhe and M{\v{e}}ch, Radom{\'\i}r and Lu, Huchuan and He, You},
  booktitle={Computer Vision--ECCV 2020: 16th European Conference, Glasgow, UK, August 23--28, 2020, Proceedings, Part XIV 16},
  pages={490--506},
  year={2020},
  organization={Springer}
}

@inproceedings{wang2019zero,
  title={Zero-shot video object segmentation via attentive graph neural networks},
  author={Wang, Wenguan and Lu, Xiankai and Shen, Jianbing and Crandall, David J and Shao, Ling},
  booktitle={Proceedings of the IEEE/CVF International Conference on Computer Vision},
  pages={9236--9245},
  year={2019}
}

@inproceedings{jin2017video,
  title={Video scene parsing with predictive feature learning},
  author={Jin, Xiaojie and Li, Xin and Xiao, Huaxin and Shen, Xiaohui and Lin, Zhe and Yang, Jimei and Chen, Yunpeng and Dong, Jian and Liu, Luoqi and Jie, Zequn and others},
  booktitle={Proceedings of the IEEE International Conference on Computer Vision},
  pages={5580--5588},
  year={2017}
}

@inproceedings{gadde2017semantic,
  title={Semantic video cnns through representation warping},
  author={Gadde, Raghudeep and Jampani, Varun and Gehler, Peter V},
  booktitle={Proceedings of the IEEE International Conference on Computer Vision},
  pages={4453--4462},
  year={2017}
}

@inproceedings{huang2018efficient,
  title={Efficient uncertainty estimation for semantic segmentation in videos},
  author={Huang, Po-Yu and Hsu, Wan-Ting and Chiu, Chun-Yueh and Wu, Ting-Fan and Sun, Min},
  booktitle={Proceedings of the European Conference on Computer Vision (ECCV)},
  pages={520--535},
  year={2018}
}

@inproceedings{nilsson2018semantic,
  title={Semantic video segmentation by gated recurrent flow propagation},
  author={Nilsson, David and Sminchisescu, Cristian},
  booktitle={Proceedings of the IEEE conference on computer vision and pattern recognition},
  pages={6819--6828},
  year={2018}
}

@inproceedings{kundu2016feature,
  title={Feature space optimization for semantic video segmentation},
  author={Kundu, Abhijit and Vineet, Vibhav and Koltun, Vladlen},
  booktitle={Proceedings of the IEEE conference on computer vision and pattern recognition},
  pages={3168--3175},
  year={2016}
}

@inproceedings{hur2016joint,
  title={Joint optical flow and temporally consistent semantic segmentation},
  author={Hur, Junhwa and Roth, Stefan},
  booktitle={European Conference on Computer Vision},
  pages={163--177},
  year={2016},
  organization={Springer}
}

@inproceedings{shelhamer2016clockwork,
  title={Clockwork convnets for video semantic segmentation},
  author={Shelhamer, Evan and Rakelly, Kate and Hoffman, Judy and Darrell, Trevor},
  booktitle={European Conference on Computer Vision},
  pages={852--868},
  year={2016},
  organization={Springer}
}

@inproceedings{zhu2017deep,
  title={Deep feature flow for video recognition},
  author={Zhu, Xizhou and Xiong, Yuwen and Dai, Jifeng and Yuan, Lu and Wei, Yichen},
  booktitle={Proceedings of the IEEE conference on computer vision and pattern recognition},
  pages={2349--2358},
  year={2017}
}

@inproceedings{mahasseni2017budget,
  title={Budget-aware deep semantic video segmentation},
  author={Mahasseni, Behrooz and Todorovic, Sinisa and Fern, Alan},
  booktitle={Proceedings of the IEEE Conference on Computer Vision and Pattern Recognition},
  pages={1029--1038},
  year={2017}
}

@inproceedings{li2018low,
  title={Low-latency video semantic segmentation},
  author={Li, Yule and Shi, Jianping and Lin, Dahua},
  booktitle={Proceedings of the IEEE Conference on Computer Vision and Pattern Recognition},
  pages={5997--6005},
  year={2018}
}

@inproceedings{xu2018dynamic,
  title={Dynamic video segmentation network},
  author={Xu, Yu-Syuan and Fu, Tsu-Jui and Yang, Hsuan-Kung and Lee, Chun-Yi},
  booktitle={Proceedings of the IEEE conference on computer vision and pattern recognition},
  pages={6556--6565},
  year={2018}
}

@inproceedings{liu2020efficient,
  title={Efficient semantic video segmentation with per-frame inference},
  author={Liu, Yifan and Shen, Chunhua and Yu, Changqian and Wang, Jingdong},
  booktitle={European Conference on Computer Vision},
  pages={352--368},
  year={2020},
  organization={Springer}
}

\end{document}